%% file: acl_latex.tex
\documentclass[11pt]{article}

\usepackage[final]{acl}
\usepackage[utf8]{inputenc}
\usepackage{times}
\usepackage{latexsym}
\usepackage{xspace}
\usepackage{booktabs}
\usepackage{amsmath}
\usepackage{etoc}
\usepackage{graphicx}
\usepackage{wrapfig}
\usepackage{graphicx}
\usepackage{xcolor}
\usepackage{soul}
\usepackage{subcaption}

\usepackage[T1]{fontenc}
\usepackage[utf8]{inputenc}
\usepackage{enumitem}
\usepackage[T5]{fontenc}       % hỗ trợ bảng mã tiếng Việt
\usepackage[utf8]{inputenc}    % sử dụng mã hóa UTF-8 (nếu bạn gõ tiếng Việt trực tiếp)
\usepackage[vietnamese,english]{babel}  % ngôn ngữ tiếng Việt
\usepackage{fancyvrb}
\usepackage{microtype}

\usepackage{inconsolata}
\usepackage[dvipsnames]{xcolor}

\usepackage{pifont}
\usepackage{tcolorbox}
\usepackage{xcolor}       
\usepackage{caption}     
\usepackage{amssymb}      
\usepackage{pifont}       
\usepackage{float}         
\usepackage{array}         
\usepackage{pifont}
\usepackage[table]{xcolor}

\newcommand{\cmark}{\ding{51}} % ✓
\newcommand{\xmark}{\ding{55}} % ✗
\usepackage{fontawesome5}
\usepackage{hyperref}

\usepackage{hyperref} 
\usepackage[nameinlink,capitalise,noabbrev]{cleveref} %
\usepackage{xspace}  

\newcommand{\vmmu}{\texttt{VMMU}\xspace}
\newcommand{\subsec}[1]{\noindent\textbf{#1}~~}

\newcommand{\eg}{e.g.\xspace}
\newcommand{\ie}{i.e.\xspace}

\crefname{figure}{Fig.}{Figs.}
\Crefname{figure}{Fig.}{Figs.}

\crefname{table}{Tab.}{Tabs.}
\Crefname{table}{Tab.}{Tabs.}

\crefname{section}{Sec.}{Secs.}
\Crefname{section}{Sec.}{Secs.}

\PassOptionsToPackage{table,usenames,dvipsnames}{xcolor}

\newcommand{\greentick}{\textcolor{ForestGreen}{\cmark}\xspace}
\newcommand{\redxmark}{\textcolor{red}{\xmark}\xspace}

\newcommand{\ayasmalllogo}{{\includegraphics[scale=0.03]{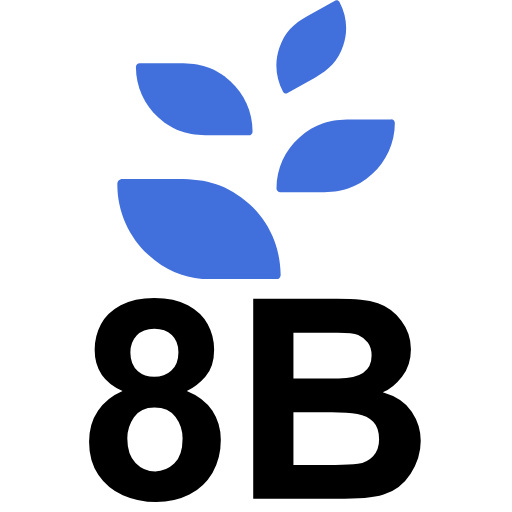}}\xspace}
\newcommand{\ayalargelogo}{{\includegraphics[scale=0.03]{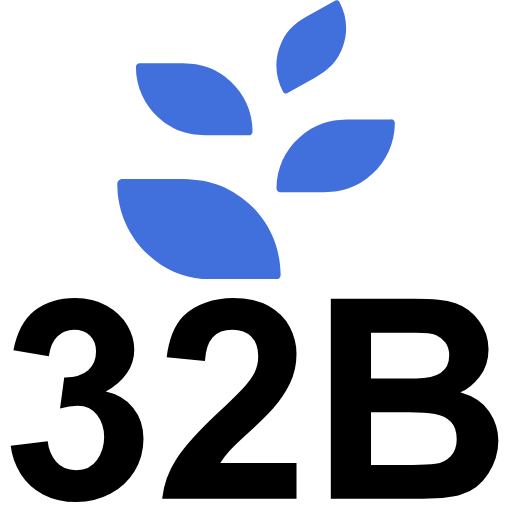}}\xspace}

\newcommand{\gemmasmalllogo}{{\includegraphics[scale=0.03]{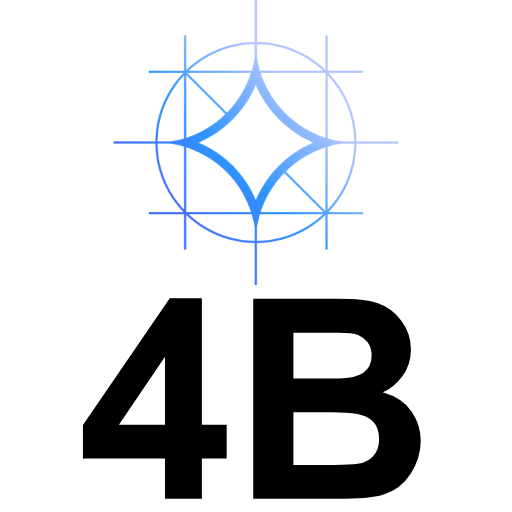}}\xspace}
\newcommand{\gemmalargelogo}{{\includegraphics[scale=0.03]{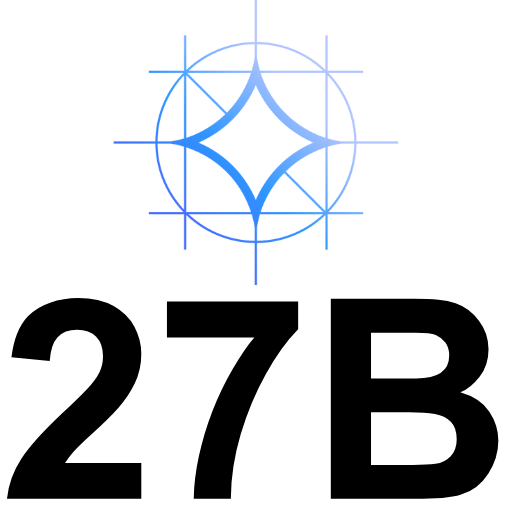}}\xspace}

\newcommand{\llamavericklogo}{{\includegraphics[scale=0.03]{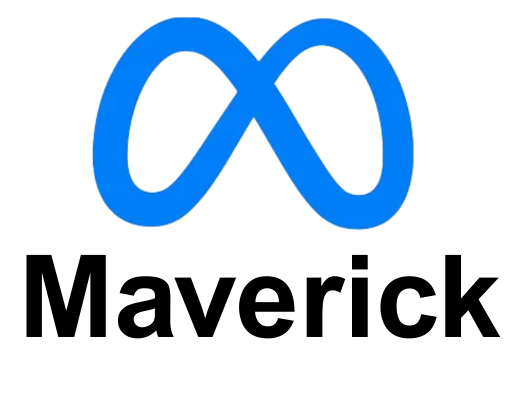}}\xspace}
\newcommand{\llamascoutlogo}{{\includegraphics[scale=0.03]{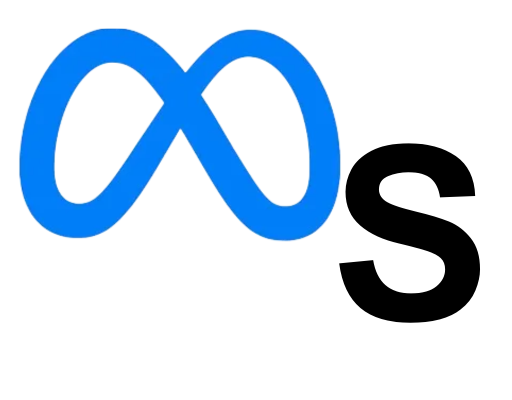}}\xspace}

\newcommand{\mistralsmalllogo}{{\includegraphics[scale=0.03]{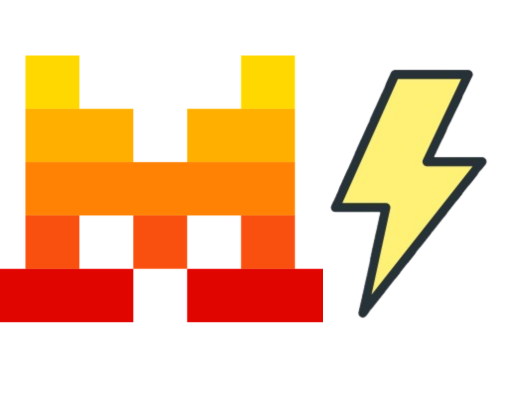}}\xspace}
\newcommand{\mistralmediumlogo}{{\includegraphics[scale=0.03]{images/Mistral.png}}\xspace}

\newcommand{\qwensmalllogo}{{\includegraphics[scale=0.0345]{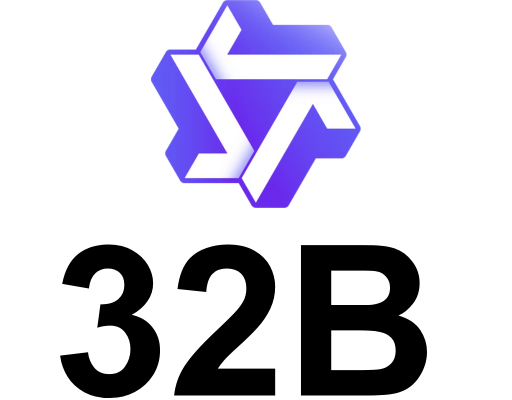}}\xspace}
\newcommand{\qwenlargelogo}{{\includegraphics[scale=0.0215]{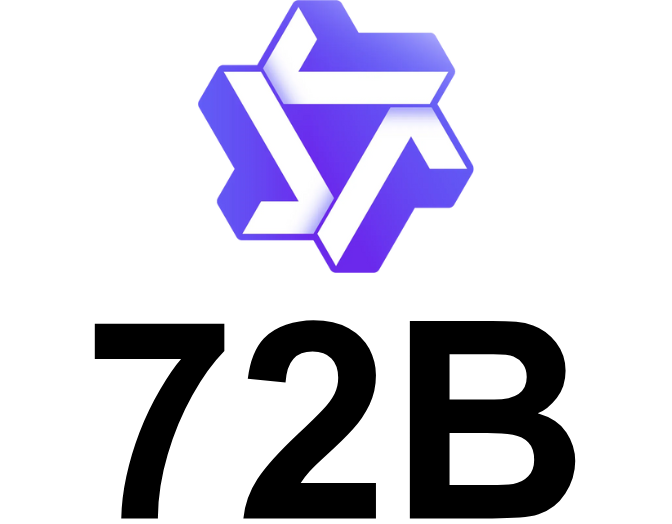}}\xspace}

\newcommand{\geminiflashlogo}{{\includegraphics[scale=0.028]{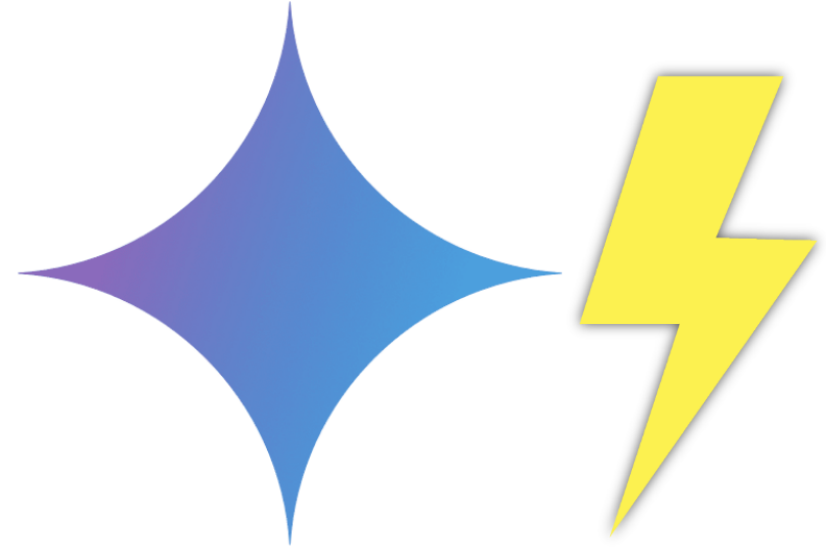}}\xspace}
\newcommand{\sonnetfourlogo}{{\includegraphics[scale=0.0085]{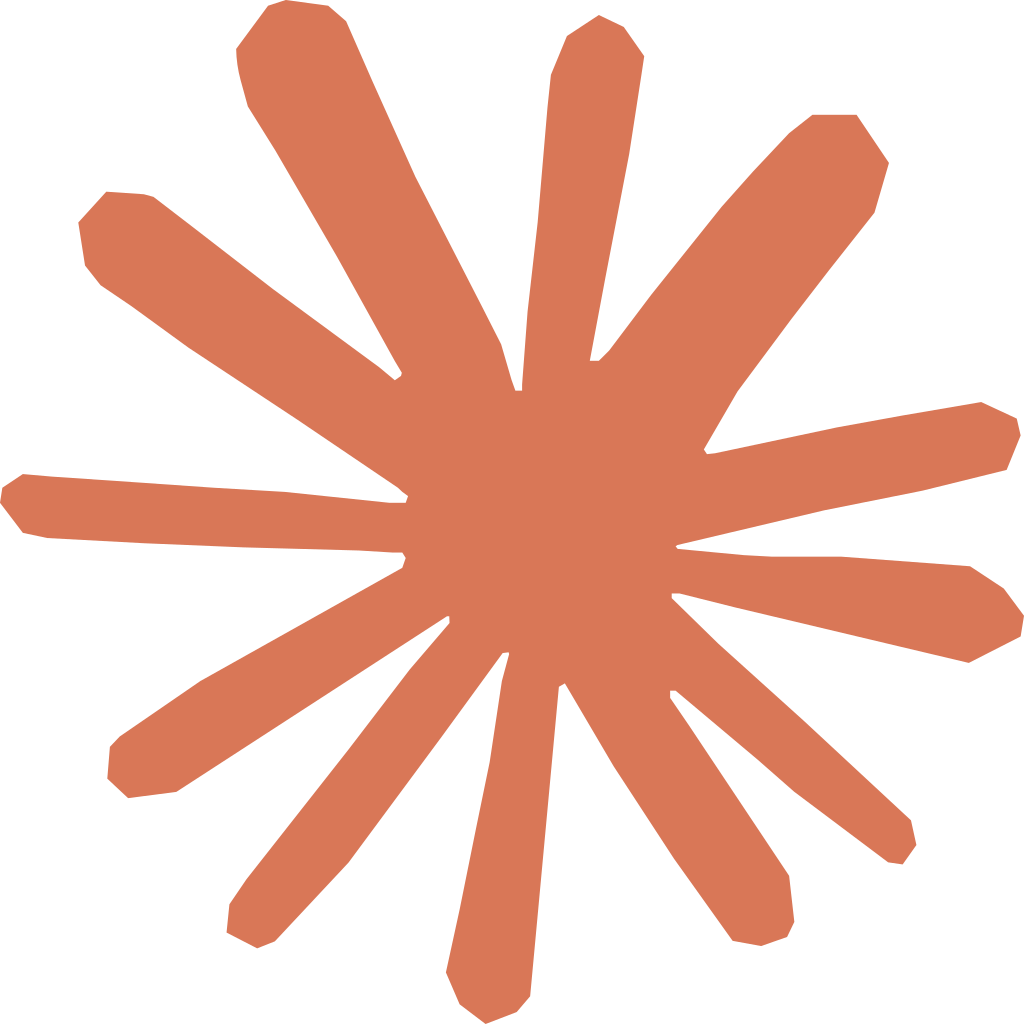}}\xspace}

\newcommand{\gptfourlogo}{{\includegraphics[scale=0.020]{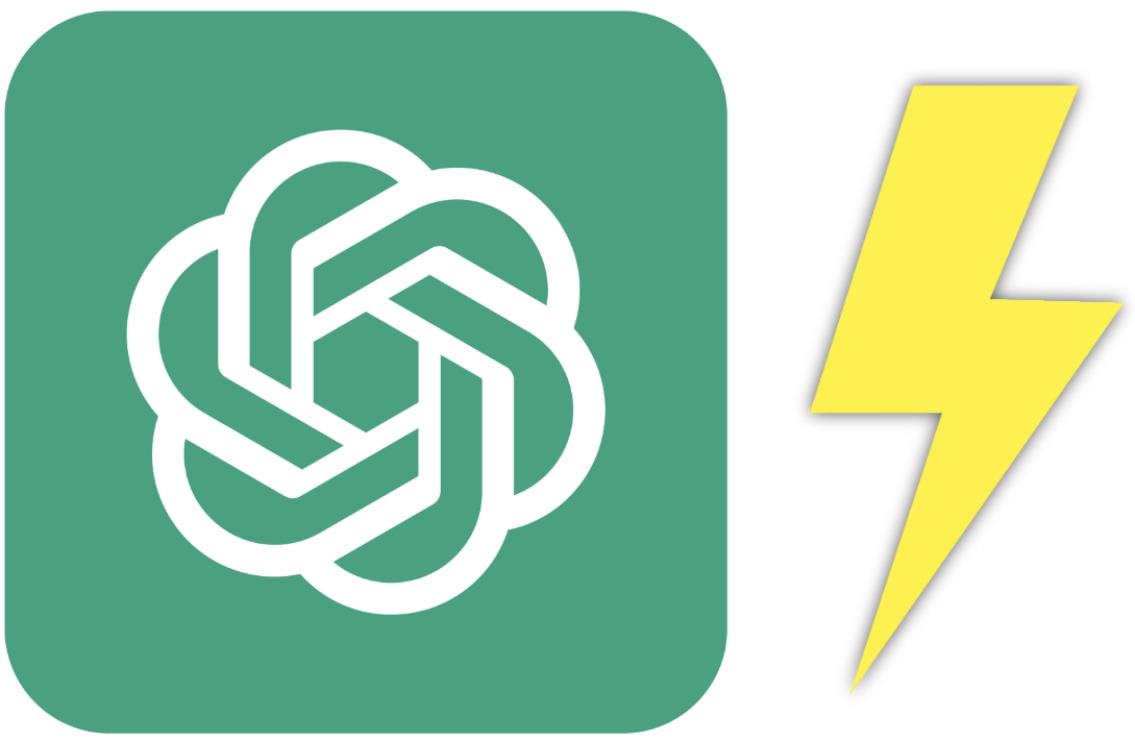}}\xspace}

\newcommand{\gptthinklogo}{{\includegraphics[scale=0.0185]{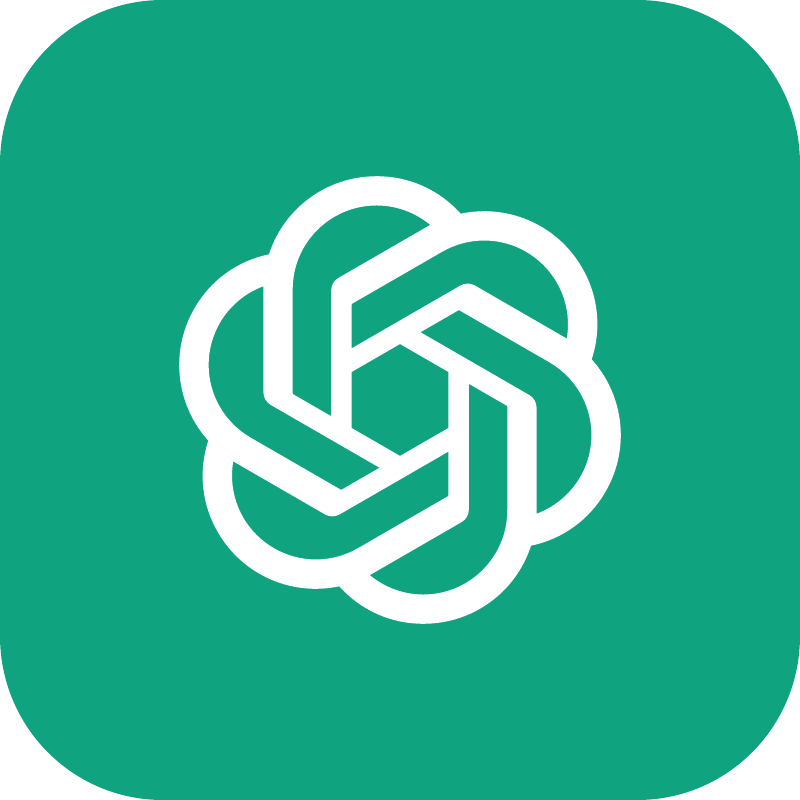}}\xspace}

\newcommand{\geminiprologo}{{\includegraphics[scale=0.027]
{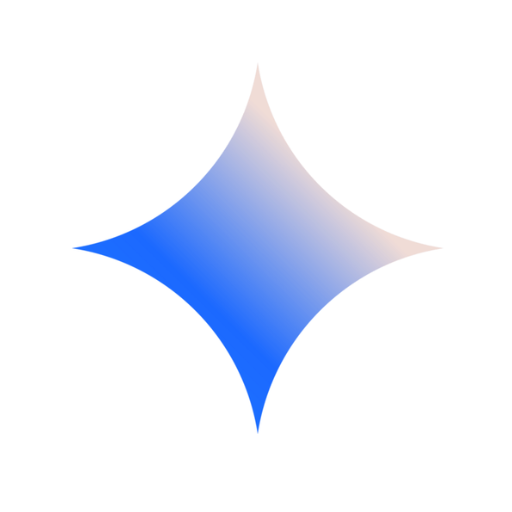}}\xspace}

\newcommand{\gptfivelogo}{{\includegraphics[scale=0.038]{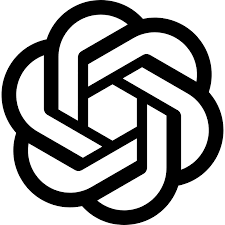}}\xspace}

\newcommand{\vmmulogo}{{\includegraphics[scale=0.020]{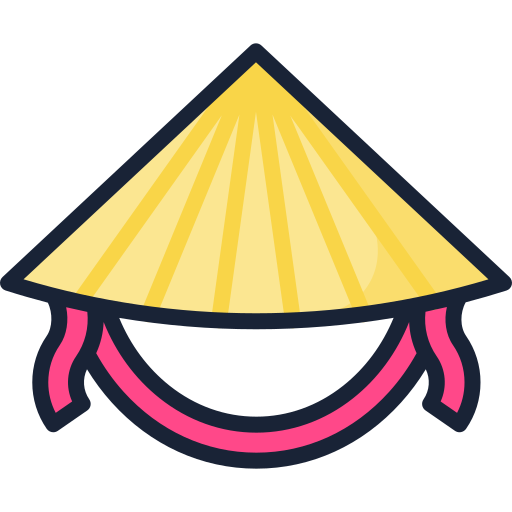}}\xspace}

\newcommand{\mathlogo}{{\includegraphics[scale=0.02]{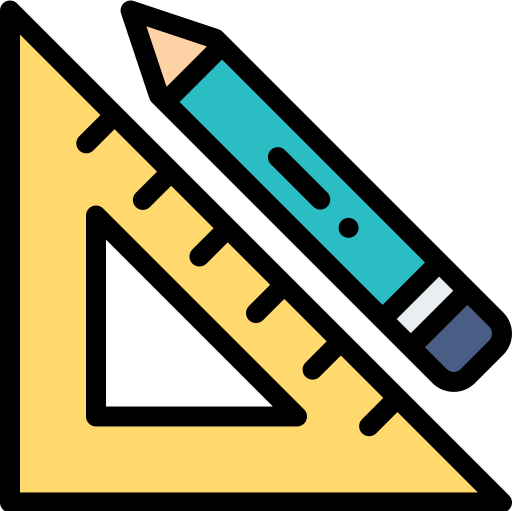}}\xspace}
\newcommand{\physiclogo}{{\includegraphics[scale=0.023]{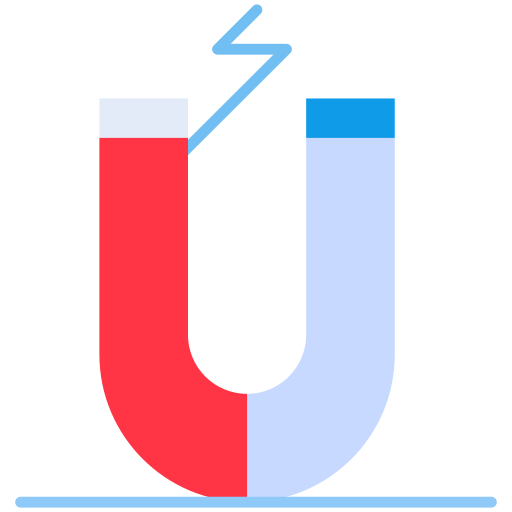}}\xspace}
\newcommand{\chemistrylogo}{{%
  \includegraphics[scale=0.17]{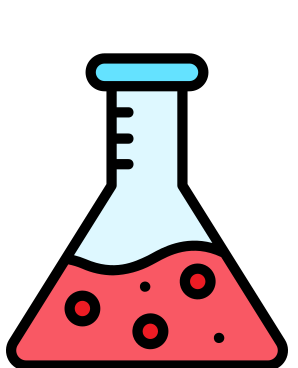}%
}\xspace}
\newcommand{\biologylogo}{{\includegraphics[scale=0.016]{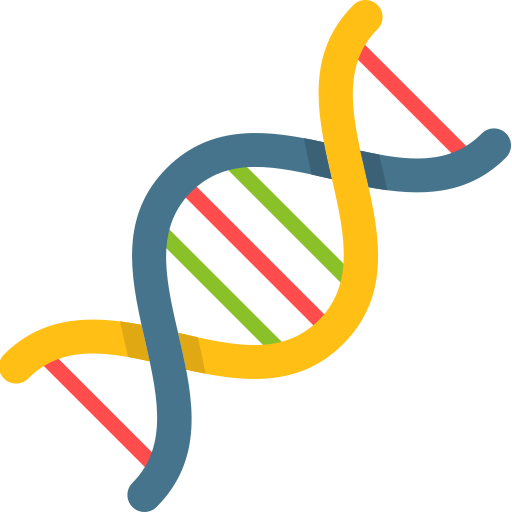}}\xspace}
\newcommand{\geographylogo}{{\includegraphics[scale=0.02]{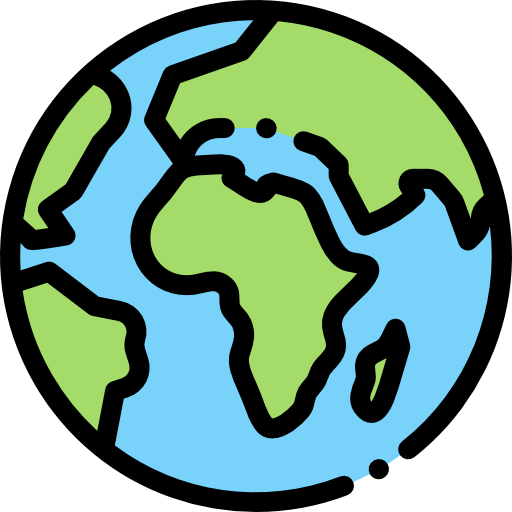}}\xspace}
\newcommand{\drivinglogo}{{\includegraphics[scale=0.02]{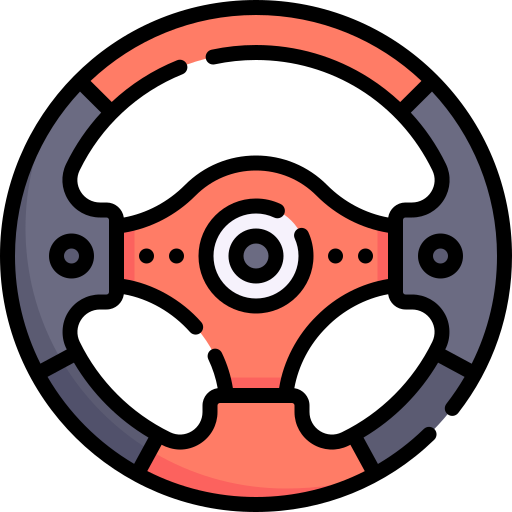}}\xspace}
\newcommand{\iqlogo}{{\includegraphics[scale=0.018]{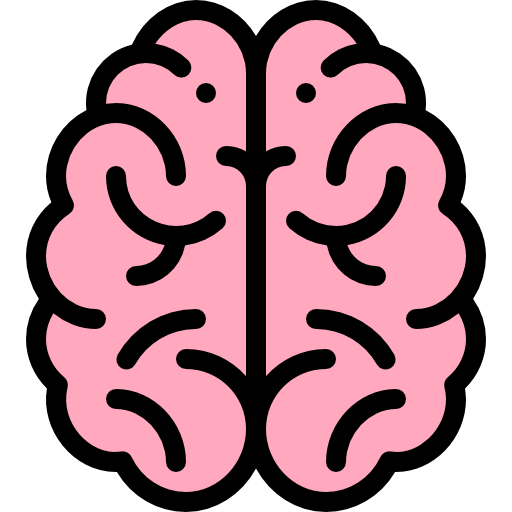}}\xspace}

\definecolor{aya_blue}{RGB}{59,118,227}
\definecolor{gemma_pink}{RGB}{236,64,122}  
\definecolor{llama_blue}{RGB}{0,119,255}
\definecolor{mistral_orange}{RGB}{255,128,0}
\definecolor{qwen_purple}{RGB}{108,54,255}
\definecolor{gpt_green}{RGB}{26,127,100}
\definecolor{sonnet_orange}{RGB}{255,159,28}
\definecolor{gemini_blue}{RGB}{66,133,244}
\definecolor{o3_purple}{RGB}{103,80,164}

\newcommand{\modelname}[4]{%
  \texttt{%
    \textcolor{black}{#1}%
    \textcolor{#2}{#3}%
    \textcolor{black}{#4}%
  }%
}

\newcommand{\geminipro}{\modelname{Gemini-}{gemini_blue}{3}{-Pro}\xspace}

\newcommand{\geminiflash}{\modelname{Gemini-}{gemini_blue}{2.5}{-Flash}\xspace}

\newcommand{\sonnetfour}{\modelname{Sonnet-}{sonnet_orange}{4.0}{}\xspace}

\newcommand{\opusfour}{\modelname{Opus-}{sonnet_orange}{4.6}{}\xspace}

\newcommand{\gptfour}{\modelname{GPT-}{gpt_green}{4.1}{}\xspace}
\newcommand{\gptfive}{\modelname{GPT-}{gpt_green}{5}{}\xspace}

\newcommand{\othree}{\textcolor{o3_purple}{o3}\xspace}
\newcommand{\gptthink}{\othree} % đã có \xspace trong \othree

\newcommand{\llamascout}{\modelname{Llama-}{black}{4}{-Scout}\xspace}
\newcommand{\llamamaverick}{\modelname{Llama-}{black}{4}{-Maverick}\xspace}

\newcommand{\gemmasmall}{\modelname{Gemma-}{black}{3}{-4B}\xspace}
\newcommand{\gemmalarge}{\modelname{Gemma-}{black}{3}{-27B}\xspace}

\newcommand{\qwensmall}{\modelname{Qwen-}{black}{2.5}{-VL-32B}\xspace}
\newcommand{\qwenlarge}{\modelname{Qwen-}{black}{2.5}{-VL-72B}\xspace}

\newcommand{\mistralsmall}{\modelname{Mistral-Small-}{black}{3.2}{-24B}\xspace}
\newcommand{\mistralmedium}{\modelname{Mistral-Medium-}{black}{3}{}\xspace}

\newcommand{\ayasmall}{\modelname{Aya-Vision-}{black}{8B}{}\xspace}
\newcommand{\ayalarge}{\modelname{Aya-Vision-}{black}{32B}{}\xspace}

\usepackage{tcolorbox}
\global
\tcbset{
  aibox/.style={
    width=\textwidth,
    top=10pt,
    colback=white,
    colframe=black,
    colbacktitle=black,
    center,
  }
}

\tcbset{
  aiboxcol/.style={
    width=\columnwidth,
    top=10pt,
    colback=white,
    colframe=black,
    colbacktitle=black,
    center,
    fontupper=\small,  % Controls content font size
    fonttitle=\small,  % Controls title font size
  }
}

\newtcolorbox{AIbox}[2][]{aibox,title=#2,#1}
\tcbset{
  aiboxsmall/.style={
    width=0.62\textwidth,
    top=10pt,
    colback=white,
    colframe=black,
    colbacktitle=black,
    enhanced,
    top,
    attach boxed title to top left={yshift=-0.1in,xshift=0.15in},
    boxed title style={boxrule=0pt,colframe=white,},
  }
}

\newtcolorbox{AIboxcol}[2][]{aiboxcol,title=#2,#1}
\tcbset{
  aiboxsmall/.style={
    width=0.62\columnwidth,
    top=10pt,
    colback=white,
    colframe=black,
    colbacktitle=black,
    enhanced,
    top,
    attach boxed title to top left={yshift=-0.1in,xshift=0.15in},
    boxed title style={boxrule=0pt,colframe=white,},
  }
}

\newtcolorbox{AIboxSmall}[2][]{aiboxsmall,title=#2,#1}
\definecolor{aigold}{RGB}{244,210, 1} 
\definecolor{aired}{RGB}{255,180,181}
\newlength\savewidth

\definecolor{defaultcolor}{gray}{0.9}

\usepackage{multirow} 
\definecolor{darkblue}{RGB}{31, 78, 121}   % Dark blue for the user
\definecolor{darkgreen}{RGB}{78, 121, 31}  % Dark green for GPT-4V
\definecolor{darkred}{RGB}{139,0,0}

\newcommand{\quest}{\textcolor{darkblue}{\textbf{Question:}} }
\newcommand{\groundtruth}{\textcolor{darkgreen}{\textbf{Ground truth:}} }
\newcommand{\prompt}{\textcolor{darkred}{\textbf{Prompt:}} }

\newcommand{\increase}[1]{(\textcolor{ForestGreen}{+#1})}
\newcommand{\increasenoparent}[1]{\textcolor{ForestGreen}{+#1}}
\newcommand{\decrease}[1]{(\textcolor{red}{-#1})}
\newcommand{\decreasenoparent}[1]{\textcolor{red}{-#1}}

\definecolor{hallucinateyellow}{RGB}{255, 245, 234}

\definecolor{lightblue}{RGB}{210, 235, 255}
\definecolor{lightgreen}{RGB}{220, 255, 220}
\newcommand{\lightbluepanel}[1]{\sethlcolor{lightblue}\hl{#1}}

\newcommand{\papertitle}[0]{\includegraphics[scale=0.028]{images/conical.png}\xspace VMMU: A Vietnamese Multitask Multimodal\\Understanding and Reasoning Benchmark}

\title{\papertitle}

\newcommand{\equalcontrib}{\textsuperscript{*}}

\makeatletter
\newcommand{\symbolthanks}[2]{%
  \footnotemark[#1]%
  \protected@xdef\@thanks{%
    \@thanks
    \protect\footnotetext[#1]{#2}%
  }%
}

\author{
  \textbf{Vy Tuong Dang\equalcontrib\textsuperscript{1}},
  \textbf{An Vo\equalcontrib\symbolthanks{3}{Work done while AV was at MBZUAI and KAIST.}\textsuperscript{3}},
  \textbf{Emilio Villa-Cueva\textsuperscript{2}},
  \\
  \textbf{Quang Tau\textsuperscript{1}},
  \textbf{Duc Dm\textsuperscript{1}},
  \textbf{Thamar Solorio\textsuperscript{2}},
  \textbf{Daeyoung Kim\textsuperscript{1}}
\\
  \textsuperscript{1}KAIST,
  \textsuperscript{2}MBZUAI,
  \textsuperscript{3}University of Michigan
\\
\small{
  \textsuperscript{*}\textit{Equal contribution}
}
\\
\small{
  \href{mailto:vydang@kaist.ac.kr,quangtau,ducdm200158@kaist.ac.kr,kimd@kaist.ac.kr}
    {\texttt{\{vydang,ducdm200158,kimd\}@kaist.ac.kr}}
}
\\
\small{
  \href{mailto:voan@umich.edu}
    {\texttt{voan@umich.edu}}
  \quad
  \href{mailto:emilio.villa@anmbzuai.ac.ae,thamar.solorio@mbzuai.ac.ae}
    {\texttt{\{emilio.villa,thamar.solorio\}@mbzuai.ac.ae}}
}
}

\begin{document}
\maketitle

% \twocolumn[{
% \maketitle
% \vspace{-30pt}
% \begin{center}
%     \includegraphics[width=0.8\textwidth]{images/vmmu_teaser_figure.pdf}
%     \captionof{figure}{Test}
%     % \vspace{1mm}
%     \label{fig:teaser}
% \end{center}
% }]

\begin{abstract}
Vision language models are predominantly evaluated on English-centric multimodal benchmarks, leaving their behavior in other languages and cultures insufficiently understood.
We introduce \vmmu, a Vietnamese Multitask Multimodal Understanding and Reasoning benchmark of 2{,}548 multiple-choice questions across seven domains: Mathematics, Physics, Chemistry, Biology, Geography, Driving Test, and IQ Test. Every question requires jointly reasoning over Vietnamese text and non-text visual evidence (charts, diagrams, tables, traffic scenes), and is distributed as a single rendered page image. We evaluate 10 open-source and 5 closed-source VLMs on \vmmu. Test-time compute improves performance from 57\% to 78\% accuracy, but it remains inferior to an expert baseline (99\% accuracy). Extensive error analyses show that closed-source VLMs already achieve strong Vietnamese OCR performance, yet still struggle on \vmmu. This suggests that the primary bottleneck is multimodal grounding and reasoning rather than text recognition or Vietnamese language understanding. 
Code and data are available at: \href{https://vmmu-bench.github.io/}{vmmu-bench.github.io}.

% We further conduct fine-grained analyses of model behavior, examining common failure patterns such as option-selection bias, miscalibration, cross-lingual prompting effects, and robustness to image perturbations. 
% Our results highlight persistent challenges in multimodal reasoning under non-English conditions and position \vmmu as a focused benchmark for diagnosing VLM behavior beyond English-centric evaluation settings.

\end{abstract}
\addtocontents{toc}{\protect\setcounter{tocdepth}{-1}}

\section{Introduction}

\begin{figure}[t!]
    \centering
    \includegraphics[width=\linewidth]{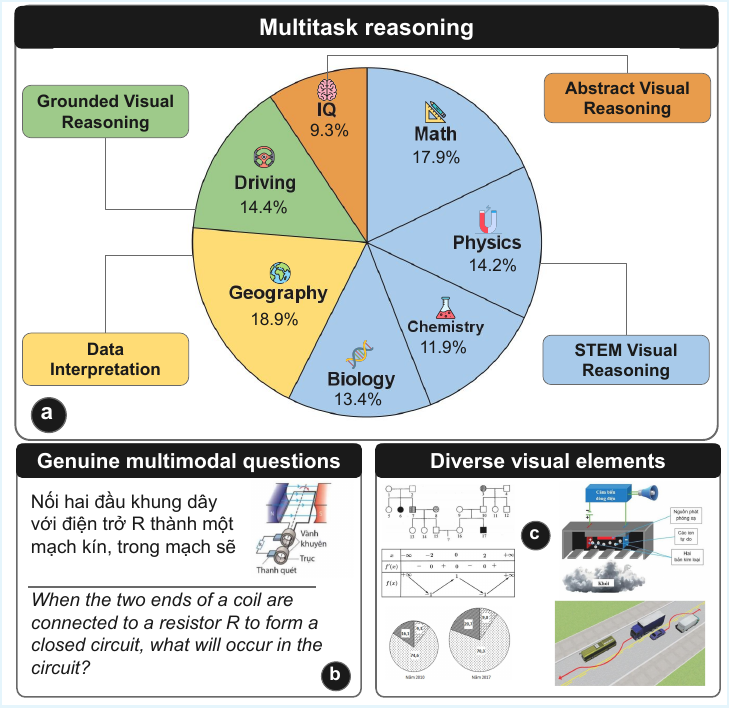}
    \caption{(a) \vmmu covers seven domains and supports multitask multimodal reasoning. (b) Many items require joint interpretation of Vietnamese text and the accompanying image. (c) \vmmu includes diverse visual evidence (\eg, diagrams, charts, tables, and traffic scenes).
    % The benchmark also provides settings to separate OCR errors from multimodal grounding, language understanding, bias, and reasoning failures.
    }
    \label{fig:teaser}
    % \vspace{-5mm}
\end{figure}

Vision language models (VLMs) have made rapid progress on multimodal benchmarks, demonstrating strong performance on complex visual and textual reasoning tasks in English \cite{mmmu,mmmu_pro,mmbench,scienceqa,mathvista}. However, it remains unclear how well these models perform in low-resource languages when language understanding, visual grounding, and reasoning are all required to complete a task. In this work, we study this question in the context of Vietnamese, a widely spoken language with nearly 100 million speakers that remains significantly underrepresented in multimodal evaluation.

This underrepresentation matters because many tasks require understanding image-based documents: people frequently rely on screenshots or photos that include instructions with structured visual content. For example, learning materials often consist of textbook pages or PDFs that include questions alongside visual information (\cref{fig:teaser}b). Similarly, people routinely encounter administrative and government documents (\eg, application forms, notices, and records) that combine dense text with visually structured elements such as tables, stamps, and form layouts.
VLMs therefore have the potential to support Vietnamese users across a range of these settings, from helping parents explain homework solutions at home to enabling citizens to interpret administrative documents. To accomplish this, models must process image-based questions that integrate \textit{Vietnamese text} with \textit{structured visual content} (\eg, charts, diagrams, tables; \cref{fig:teaser}c), which requires reliable optical character recognition (OCR) of Vietnamese, adequate grounding between textual instructions and visual evidence, and robust multimodal reasoning. Frontier VLMs have recently been reported to reach gold-medal performance on English mathematics and coding olympiads \cite{alphageometry2,gemini_imo_gold_2025,openai_competitive_programming_2025}, suggesting they may be sufficiently capable to reliably support such applications. Whether this capability actually transfers to Vietnamese multimodal tasks, however, remains untested.

%Vietnamese is one of the world’s most widely spoken languages, with around 100 million speakers, yet it remains underrepresented in multimodal benchmarks. 

%Many real-world tasks (and Vietnamese tasks are no exception) require models to jointly interpret linguistically rich prompts and \emph{structured visual content} (\eg, charts, diagrams, tables, symbolic illustrations; \cref{fig:teaser}c). Successfully solving such problems demands not only accurate language understanding, but also robust optical character recognition (OCR), effective multimodal integration, and reliable answer selection on multiple-choice questions where positional biases may occur.

% ===============================
\begin{figure*}[ht!]

    \centering
    \includegraphics[width=\textwidth]{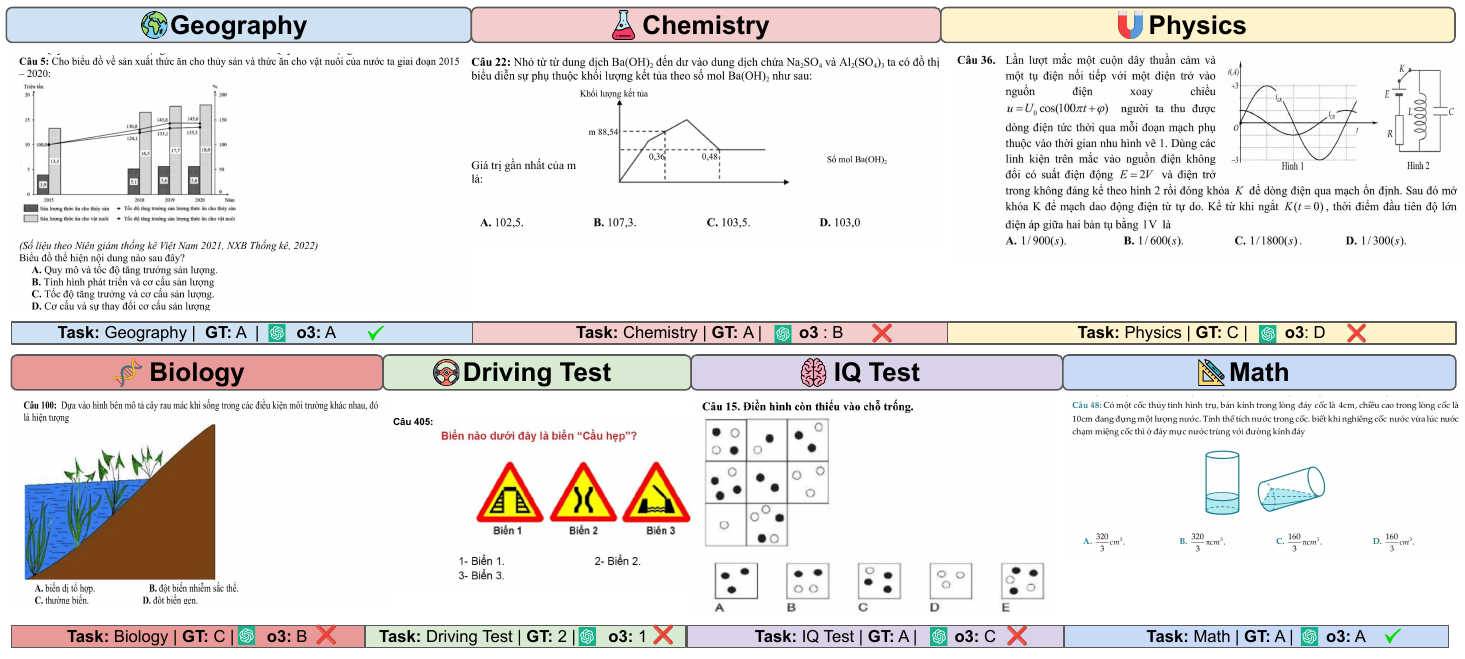} 
    \caption{
        Sample questions from \vmmulogo~\vmmu spanning seven domains. English translations are shown in \cref{fig:eng_vmmu_examples}.
    }
    \label{fig:vmmu_examples}
    % \vspace{-2mm}
\end{figure*}

% ===============================

Several LLM/VLM benchmarks include partial or full Vietnamese coverage~\cite{global_mmlu,vmlu,mmlu_pro_x}, but they are either text-only or contain only text embedded in images (rather than true visual evidence such as graphs, tables, or charts). As a result, they do not require joint visual grounding and language reasoning. Consequently, prior benchmarks offer limited insight into VLM performance on \emph{genuinely multimodal} Vietnamese questions, where non-text visual evidence is essential and must be integrated with the text to reach the correct answer. Importantly, English-only multimodal evaluation \cite{mmmu,mmmu_pro} is not sufficient to assess multilingual generalization.
%Some datasets discard visual evidence during preprocessing, converting multimodal questions into text-only problems \cite{vnhsge}. Others include Vietnamese questions paired with images, but many examples are primarily screenshots of rendered text, where the image adds little beyond providing the question and testing OCR \cite{m3exam}. 
% As a result, prior benchmarks offer limited insight into VLM performance on \emph{genuinely multimodal} Vietnamese questions, where non-text visual evidence is essential and must be integrated with the text to reach the correct answer. Importantly, English-only multimodal evaluation \cite{mmmu,mmmu_pro} is not sufficient to assess multilingual generalization.
%Even if the dominant failures arise from multimodal grounding and reasoning, model behavior is still conditioned on the question language, and robustness in English does not guarantee robustness under Vietnamese instructions and constraints. Moreover, 
Vietnamese tasks exhibit differences in rendered text, labels, domain distributions, and conventions, limiting the usefulness of English benchmarks for diagnosing whether VLMs reliably ground and reason in Vietnamese settings. 

%These considerations motivate a Vietnamese-native multimodal benchmark that directly measures cross-lingual robustness rather than assuming English performance transfers. 
%In this work
To address this gap, we introduce \vmmu: a Vietnamese Multitask Multimodal Understanding and Reasoning evaluation suite (\cref{fig:teaser}). We define \emph{multimodal questions} as those in which non-text visual evidence is necessary to solve the problem, so the model must integrate the question with visual elements. We filter out all samples where the image is not needed to answer the question, only retaining items that require understanding non-text visual evidence (\eg, diagrams, charts, tables).
% All questions in \vmmu require models to simultaneously process rendered Vietnamese text, interpret visual content, and reason based on this evidence (\cref{fig:teaser}b). We primarily evaluate VLMs in an \emph{Integrated-MM} setting, where the question text, options, and visual evidence are rendered within a single page image and additionally provide a \emph{Split-MM} variant in which the prompt is presented as text alongside a separate visual crop.

\vmmu consists of 2{,}548 carefully curated questions spanning seven domains (\cref{fig:teaser}a): \mathlogo~Mathematics, \physiclogo~Physics, \chemistrylogo~Chemistry, \biologylogo~Biology, \geographylogo~Geography, \drivinglogo~Driving Test, and \iqlogo~IQ Test. The benchmark is constructed using a semi-automated pipeline with human validation to ensure multimodal integrity and question quality. Unlike prior datasets that simplify or discard visual information~\cite{vnhsge,m3exam}, \vmmu preserves visual content in its original form, enabling direct evaluation of multimodal reasoning. Our key findings are:

\begin{enumerate}[itemsep=2pt,topsep=0pt,leftmargin=*]

     \item Open-source models average only 37\% accuracy and non-thinking closed-source models 57\% on \vmmu, both below the average high-school student (66\%). Test-time compute via thinking models improves to 78\% but remains well below the expert reference ($\sim$99\%), so no group is reliable enough for tutoring or automatic checking (\S\ref{sec:results_main_integrated}).

    \item OCR is a major bottleneck for open-source VLMs but not for closed-source ones. Closed-source models achieve strong Vietnamese embedded-text recognition (mean BLEU 89\%), while open-source models score substantially lower (mean BLEU 58\%). Thus, open-source failures are partly attributable to OCR, whereas proprietary failures point more strongly to multimodal grounding and reasoning (\S\ref{sec:results_ocr_main}).

    \item Separating text from visual evidence improves reliability. Providing question and options as text alongside a visual crop improves every evaluated VLM by \increasenoparent{7} points on average. This finding suggests that separating textual content from the image encourages models to rely more on visual reasoning while reducing distraction from text embedded in the image (\S\ref{sec:results_split_mm}).
    
    \item Translating Vietnamese inputs into English consistently hurts performance for most models by \decreasenoparent{2} points on average, which shows that native-language multimodal evaluation cannot be replaced by translation, even for models trained predominantly on English (\S\ref{sec:results_translation}).

\end{enumerate}

\section{Related work}
\subsec{Multimodal Evaluation} Recent multimodal benchmarks evaluate VLMs on increasingly diverse and structured visual media, including text-in-image, documents, and charts \cite{textvqa,stvqa,docvqa,chartqa}, and broad multi-discipline reasoning that requires jointly processing text and visual evidence \cite{mmmu,mmmu_pro,emma}. These benchmarks typically provide the textual question and the visual evidence as separate inputs. \textbf{In contrast, \texttt{VMMU}'s primary setting renders the question, answer options, and visual evidence together within a single image, matching how Vietnamese exam materials commonly appear (\cref{fig:vmmu_examples})}. Beyond this unified-image setting, \vmmu also contributes a genuine multimodal filter that removes text-only screenshots, and a suite of diagnostic variants (see \cref{sec:analyses_discussions}) that isolate distinct failure modes rather than reporting only a single aggregate score. Previous benchmarks are also largely English-centric, and directly translating them does not capture Vietnamese-native layouts, in-image labels, or curriculum conventions, leaving limited understanding of how OCR, language effects, and multimodal reasoning interact in low-resource language settings such as Vietnamese.
\begin{table}[h!]
\centering
\resizebox{\columnwidth}{!}{
\begin{tabular}{lrrc}
\toprule
\textbf{Benchmark} & \textbf{\#VN Q} & \textbf{\#VN MM Q} & \textbf{Genuine MM} \\
\midrule
EXAMS-V \cite{exam-v}          & 0       & 0         & \redxmark   \\
SeaExam \cite{seaexam}         & 1{,}745 & 0         & \redxmark  \\
VNHSGE \cite{vnhsge}           & 19{,}300& 0 & \redxmark  \\
VMLU \cite{vmlu}      & 10{,}880& 0     & \redxmark    \\
M3Exam \cite{m3exam}           & 1{,}817 & 116       & \textit{partial}    \\
\midrule
\textbf{\vmmulogo~\vmmu~(Ours)}& \textbf{2{,}548} & \textbf{2{,}548} & \greentick \\
\bottomrule
\end{tabular}}
% \vspace{-2mm}
\caption{Vietnamese coverage in representative benchmarks.
\#VN MM Q counts Vietnamese questions intended to be answered with an \emph{image input}.
\textbf{Genuine MM} marks questions requiring integration of Vietnamese text with \emph{non-text visual evidence} (\eg, charts/diagrams/maps), excluding text-only screenshots.}
\label{tab:vn_coverage}
\vspace{-2mm}
\end{table}

\subsec{Multilingual and Vietnamese Benchmarks}
Vietnamese text-only language models and evaluation benchmarks have advanced substantially \cite{phobert,vmlu}, paralleling language-specific evaluation efforts for other non-English languages such as Korean \cite{kmmlu} and Kazakh \cite{kazmmlu}. Non-English \emph{multimodal} benchmarks, however, remain concentrated on a handful of languages such as Chinese \cite{cmmmu} and Japanese \cite{jmmmu,jmmmmu_pro}. 
Meanwhile, Vietnamese multimodal coverage in particular is sparse. Most existing Vietnamese benchmarks (\cref{tab:vn_coverage}) either contain no multimodal questions or rely on weakly multimodal formats (\eg, text-only screenshots), which prevents systematic evaluation of multimodal integration and related failure modes.
For example, although VNHSGE \cite{vnhsge} contains Vietnamese exam questions, it converts all images to text-only format during preprocessing.
In contrast, \vmmu provides Vietnamese questions where visual content is integral and is designed to support diagnostic analysis of OCR, multimodal reasoning, language effects, and answer-option biases.

\begin{figure*}[ht!]

    \centering
    \includegraphics[width=\textwidth]{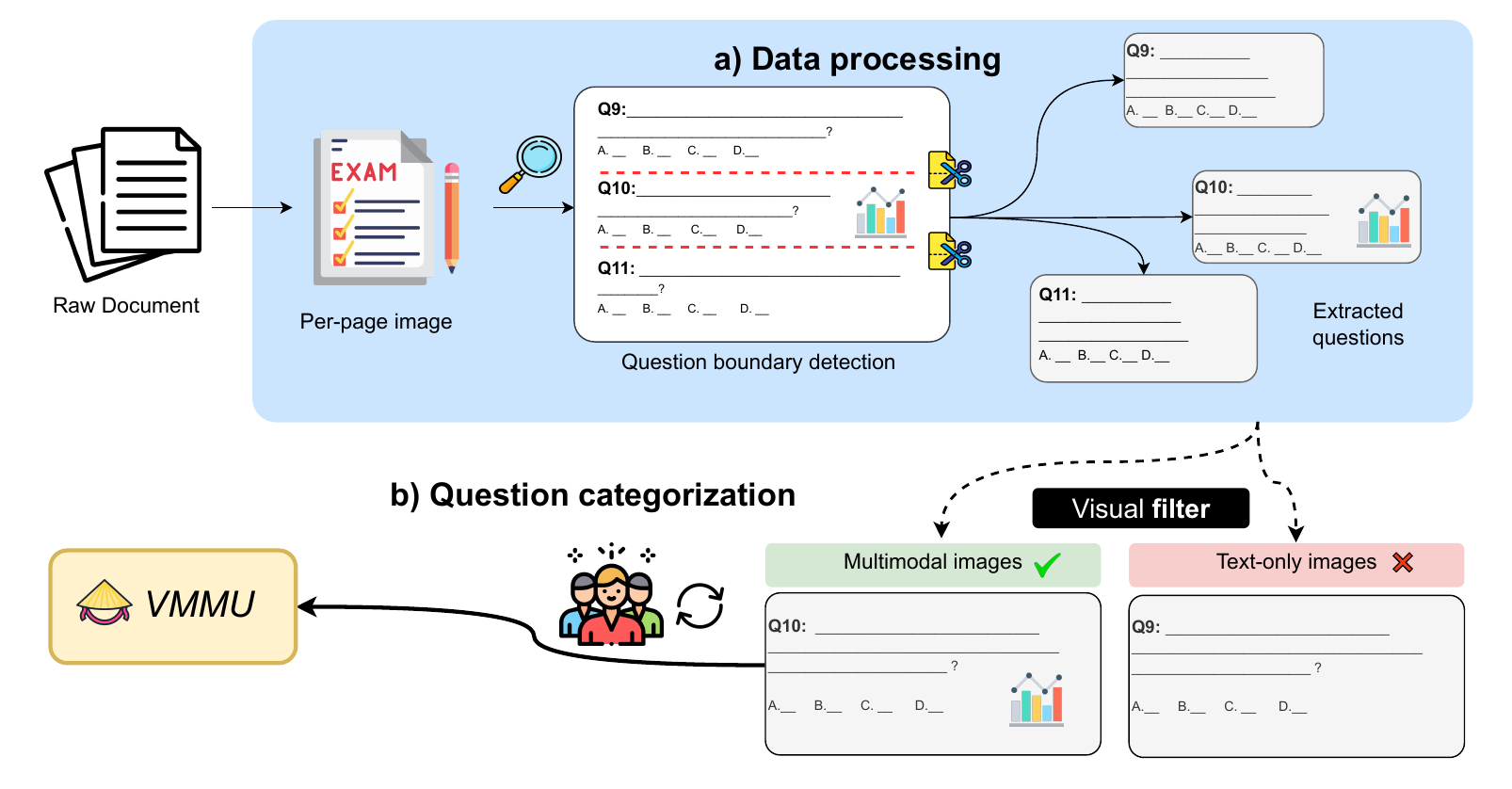} 
        \caption{\vmmu data curation pipeline. (a) Data processing renders PDFs as images and detects question boundaries. (b) Items are filtered for multimodal content and verified by native speakers.}

    \label{fig:dataset_collection}
\end{figure*}

\section{The \vmmu Benchmark}\label{sec:vmmu_benchmark}

\begin{figure}[h!]
    \centering
    \includegraphics[width=\linewidth]{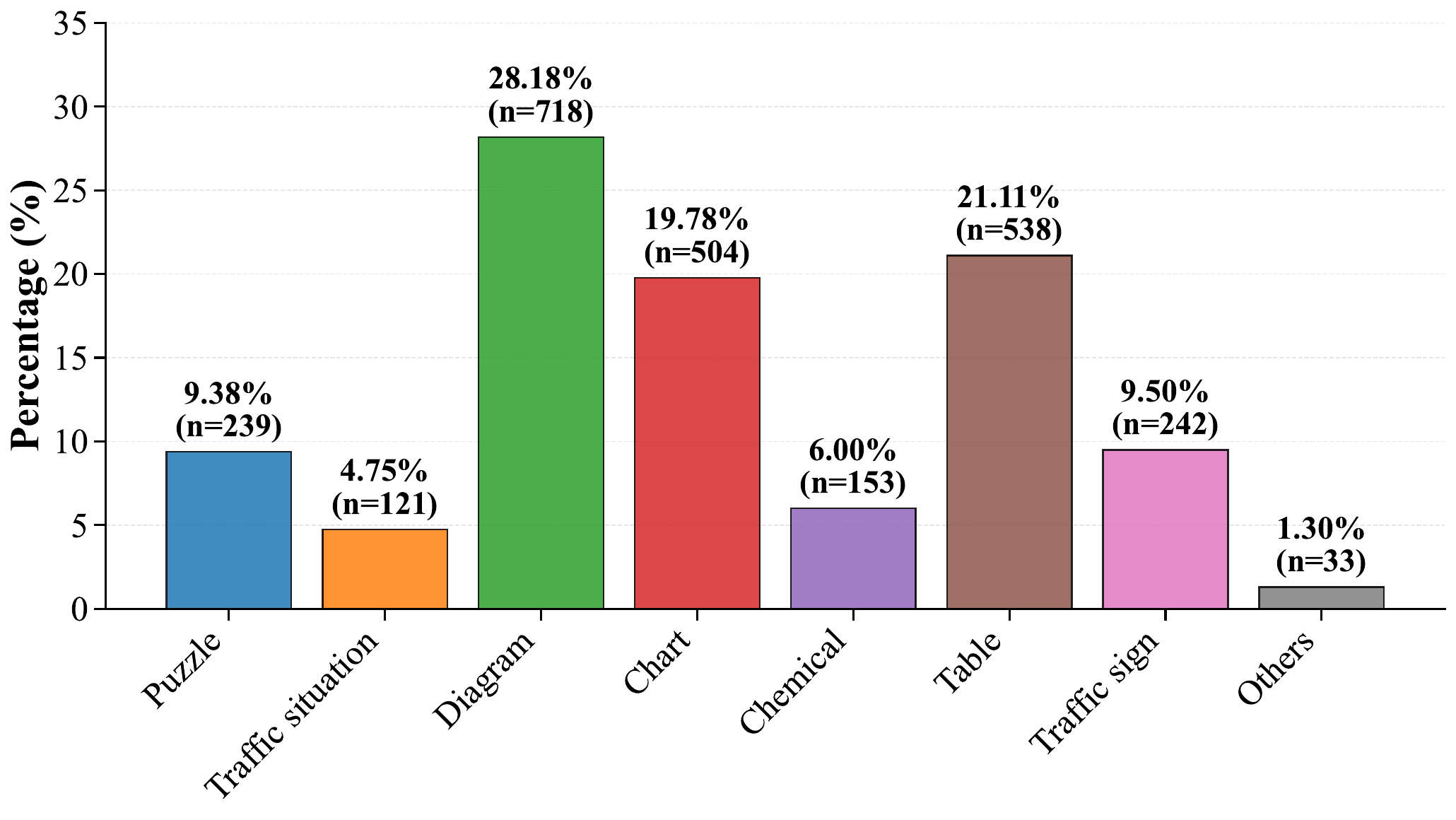}
    \caption{Distribution of visual evidence categories in \vmmu.}
    \label{fig:category_distribution}
\end{figure}

\vmmu targets scenarios where Vietnamese language content is embedded within images and must be interpreted jointly with structured visual evidence (\eg, diagrams, charts, maps, tables; see~\cref{fig:category_distribution}). 

\subsection{Benchmark scope}\label{sec:vmmu_taxonomy}
\vmmu contains 2{,}548 questions spanning seven domains: \mathlogo~Mathematics, \physiclogo~Physics, \chemistrylogo~Chemistry, \biologylogo~Biology, \geographylogo~Geography, \drivinglogo~Driving Test, and \iqlogo~IQ Test. Examples are illustrated in \cref{fig:vmmu_examples}.
These seven domains collectively probe three multimodal reasoning skills:
(\textit{i}) STEM and data interpretation (equations, diagrams, tables, plots): Math, Physics, Chemistry, Biology, Geography;
(\textit{ii}) rule-governed grounded reasoning (traffic signs and situations): Driving Test;
and (\textit{iii}) abstract visual reasoning (pattern completion and spatial logic): IQ Test.
All questions are multiple-choice. A very small fraction ($\approx$1\%) requires selecting multiple answers because the image contains more than one question.
As the first large-scale Vietnamese multimodal benchmark of this kind, we intentionally prioritize curriculum-grounded sources (official exams and government materials) so that ground-truth answers, licensing, and question boundaries are unambiguous, which is a prerequisite for clean diagnostic analysis. Extending \vmmu to in-the-wild multimodal scenarios (\eg, UI screenshots, Vietnamese cultural scenes) is a natural next step that requires different collection and verification pipelines, and we leave it to future work.

\subsec{Tasks 1--5: STEM and data interpretation (Math, Physics, Chemistry, Biology, Geography)}
Questions for these five domains are collected from Vietnam's \emph{High School Graduation Examination}, a nationwide standardized test used for graduation eligibility and university/college admission. Items span equations, scientific diagrams, Vietnamese-specific geographical charts, and maps. Samples are in \cref{sec:math_qualitative_results,sec:physics_qualitative_results,sec:chemistry_qualitative_results,sec:biology_qualitative_results,sec:geography_qualitative_results}.

\subsec{Task 6: Rule-governed grounded reasoning (Driving Test)}
Vietnamese driving license examinations are standardized safety-critical assessments. These questions test traffic-rule knowledge and situational judgment from scenario-based images that require visual understanding and application of Vietnamese traffic regulations. Samples are in \cref{sec:driving_test_qualitative_results}.

\subsec{Task 7: Abstract visual reasoning (IQ Test)}
IQ Test items are drawn from standardized pattern-recognition and spatial-logic puzzles, testing non-verbal reasoning independent of subject-matter knowledge. Samples are in \cref{sec:IQ_test_qualitative_results}.

\subsection{Data sources and curation pipeline}\label{sec:vmmu_curation}

% To ensure question quality, we collect question-answer pairs from official government sources (\eg, the Ministry of Education and Training, the Ministry of Transportation), official national academic tests, and curated public repositories (\cref{tab:vmmu_sources}). All questions are collected under Article 25 of Vietnam’s IP Law, which permits non-commercial research use (see \nameref{sec:ethics}).
% We also implement a systematic collection and filtering pipeline. We collect tests in PDF format and render each page into PNG images. 

% The key challenge is separating image-rich questions from text-only items. We use the Tesseract OCR library~\cite{tesseract} to locate question boundaries by detecting Vietnamese markers (\eg ``Câu''; Question). We then automatically identify image-containing regions using three criteria: (1) contour area analysis to capture non-textual shapes exceeding minimum area thresholds, (2) geometric filtering to distinguish text-like rectangles from image elements based on aspect ratios and dimensions, and (3) morphological operations to detect complex visual structures. Finally, all questions that pass the automated filters undergo manual review in a web-based interface, where three Vietnamese native speakers (\ie, co-authors) perform triple verification with binary accept/reject decisions to ensure content quality. Details of data collection and curation are provided in \cref{appsec:dataset_construction_method}.
We collect question-answer pairs from official government sources and standardized national academic tests (\cref{tab:vmmu_sources}), used under Article 25 of Vietnam's IP Law (non-commercial research; see \nameref{sec:ethics}). Our semi-automated pipeline (see \cref{fig:dataset_collection}) renders exam PDFs to page images, detects question boundaries with Tesseract OCR~\cite{tesseract}, filters out text-only screenshots via geometric analysis of non-text visual structures, and finalizes every item through triple verification by three Vietnamese native speakers (co-authors). Full details are in \cref{appsec:dataset_construction_method}.

% \subsection{Diagnostic variants for failure analysis}\label{sec:vmmu_diagnostic_variants}
% A key goal of \vmmu is to attribute errors to \emph{where} they arise. Starting from Integrated-MM items, we create controlled variants that hold semantics constant while changing the presentation:

% \begin{itemize}[leftmargin=1.2em]
%     \item \textbf{Integrated-MM (original):} full question image with Vietnamese text + visual evidence interleaved.
%     \item \textbf{Split-MM:} OCR text prompt/options + cropped visual evidence region (two-input representation).
%     \item \textbf{Text-only control:} OCR text prompt/options without the visual crop, to estimate how much can be solved from text priors alone.
% \end{itemize}

\section{Main evaluation: Thinking models achieve the strongest performance}\label{sec:results_main_integrated}

% \subsection{}

% Having established that proprietary OCR is reliable (\cref{sec:results_ocr_main}), we next measure end-to-end accuracy on \vmmu's main Integrated-MM setting, where question text, options, and visual evidence are rendered together in a single image (\cref{sec:vmmu_formats,fig:vmmu_formats}).

We evaluate VLMs on \vmmu, where question text, options, and visual evidence are rendered together in a single image. Our goal is to characterize how current VLMs perform on Vietnamese multimodal questions, and what level of reliability they reach relative to humans.

% \begin{figure}[h]
%     \centering

%     \begin{subfigure}{\linewidth}
%         \centering
%         \includegraphics[width=\linewidth]{images/category_distribution.pdf}
%         \caption{}
%         \label{fig:category_distribution}
%     \end{subfigure}
%     \hfill
%     \begin{subfigure}{\linewidth}
%         \centering
%         \includegraphics[width=\linewidth]{images/model_accuracy_by_category.pdf}
%         \caption{}
%         \label{fig:model_accuracy_by_category}
%     \end{subfigure}

%     \caption{Although most questions in \vmmu are Diagram (28.2\%), Table (9.4\%), Chart (19.8\%) (a), VLMs consistently perform worst on Puzzle (IQ Test) and Traffic situation questions, while Table and Traffic sign are comparatively easier (b).}
    
%     \label{fig:categories_performance}
%     \vspace{-3mm}
% \end{figure}

% \subsec{Experiments} We evaluate 5 SOTA proprietary VLMs as in \cref{sec:results_ocr_main}, and additionally include 2 recent thinking models (\gptfive and \opusfour) to strengthen our results, on the full \vmmu benchmark using the original Integrated-MM images (see~\cref{sec:vmmu_formats,fig:vmmu_formats}) with the prompt shown in \cref{fig:prompt_question,fig:prompt_question2}. Details of the experimental settings are provided in \cref{section:detail_experiment}.

\subsection{Experiments} We evaluate 3 groups of models on the full \vmmu benchmark with the prompt shown in \cref{fig:prompt_question,fig:prompt_question2}: (1) 10 open-source VLMs spanning multiple families and sizes, (2) 3 non-thinking closed-source VLMs, and (3) 4 thinking variants of closed-source VLMs (including 2 recent thinking models, \gptfive and \opusfour). Full experimental details are in \cref{section:detail_experiment}.

\begin{table}[h]
    \centering
    \resizebox{\columnwidth}{!}{%
    \begin{tabular}{lcccccccc}
    \toprule
    Model &
    Math &
    Phys &
    Chem &
    Bio &
    Geo &
    Drv &
    IQ &
    Mean \\
    \midrule
\multicolumn{9}{c}{Human \& Random baseline} \\ 
High schoolers (Average)
& 64.50 & 66.70 & 66.80 & 62.80 & 71.90 & -- & -- & 66.54 \\
Human expert (Best)
& 98.00 & 100.0 & 100.0 & 100.0 & 100.0 & -- & -- & 99.60 \\
Random baseline
& 25.00 & 24.66 & 24.38 & 24.67 & 25.00 & 33.24 & 24.46 & 25.92 \\

\midrule

\multicolumn{9}{c}{Open-source VLMs} \\

\ayasmalllogo~~\ayasmall
& 24.34 & 26.04 & 27.81 & 24.93 & 26.20 & 29.16 & 20.42 & 25.56 \\
\ayalargelogo~~\ayalarge
& 23.90 & 24.10 & 19.54 & 20.53 & 14.55 & 34.60 & 20.00 & 22.46 \\
\gemmasmalllogo~~\gemmasmall
& 26.32 & 17.45 & 22.19 & 22.29 & 27.65 & 40.60 & 21.67 & 25.45 \\
\gemmalargelogo~~\gemmalarge
& 47.37 & 30.47 & 38.74 & 30.79 & 47.40 & 43.87 & 35.00 & 39.09 \\
\qwensmalllogo~~\qwensmall
& 61.18 & 49.17 & 44.70 & 42.82 & 71.31 & 54.50 & 41.25 & 52.13 \\
\qwenlargelogo~~\qwenlarge
& 58.33 & 49.86 & 48.34 & 40.47 & 64.66 & 53.95 & 46.67 & 51.75 \\
\llamascoutlogo~~\llamascout
& 59.65 & 44.88 & 45.03 & 38.71 & 69.02 & 49.05 & 37.08 & 49.06 \\
\llamavericklogo~~\llamamaverick
& 42.54 & 29.36 & 32.78 & 27.57 & 53.22 & 53.95 & 29.17 & 38.37 \\
\mistralsmalllogo~~\mistralsmall
& 34.65 & 30.28 & 28.48 & 28.15 & 30.98 & 39.78 & 30.83 & 31.88 \\
\mistralmediumlogo~~\mistralmedium
& 44.74 & 36.57 & 37.75 & 29.33 & 36.59 & 45.78 & 33.75 & 37.78 \\
\midrule
\rowcolor{gray!20}  Model Mean & 42.30 & 33.82 & 34.54 & 30.56 & 44.16 & 44.52 & 31.58 & \lightbluepanel{37.35} \\

    \midrule

 \multicolumn{9}{c}{Non-thinking closed-source VLMs} \\ 
    \geminiflashlogo~~\textcolor{gemini_blue}{\geminiflash}
& 82.46 & 67.04 & 78.15 & 63.05 & 85.24 & 71.39 & 52.08 & 71.34 \\

    \sonnetfourlogo~~\textcolor{sonnet_orange}{\sonnetfour}
        & 64.25 & 41.08 & 53.31 & 44.87 & 48.44 & 58.04 & 44.17 & 50.58 \\
    \gptfourlogo~~\textcolor{gpt_green}{\gptfour}
        & 46.27 & 43.21 & 44.37 & 44.87 & 69.85 & 66.21 & 46.25 & 51.58 \\

    \midrule
    \rowcolor{gray!20} Model Mean & 64.33 & 50.44 & 58.61 & 50.93 & 67.84 & 65.21 & 47.50 & \lightbluepanel{57.83} \\
    
    \midrule
    \multicolumn{9}{c}{Thinking closed-source VLMs} \\
    \gptfivelogo~~\gptfive & 91.23 & 76.45 & 85.10 & 72.73 & 89.40 & 75.20 & 58.75 & 78.41 \\

    \opusfour & 83.33 & 67.31 & 79.47 & 71.26 & 90.85 & 73.84 & 56.25 & 74.62 \\

    \gptthinklogo~~\textcolor{o3_purple}{\gptthink}
        & 84.87 & 68.98 & 82.78 & 67.16 & 88.98 & 74.66 & 50.42 & 73.98 \\
    \geminiprologo~~\textcolor{gemini_blue}{\geminipro}
        & \textbf{92.54} & \textbf{81.16} & \textbf{91.72} & \textbf{80.94} & \textbf{94.59} & \textbf{91.28} & \textbf{72.08} & \textbf{86.33} \\
\midrule
\rowcolor{gray!20} Model Mean & 87.99 & 73.48 & 84.77 & 73.02 & 90.96 & 78.75 & 59.38 & \lightbluepanel{78.34} \\

    \bottomrule
    \end{tabular}%
}
    \caption{Thinking closed-source VLMs are substantially stronger than non-thinking closed-source VLMs across all seven domains of \vmmu (78.34\% vs.\ 57.83\% mean), while open-source VLMs fall further behind (37.35\%).
    % \geminiprologo~\geminipro leads overall with 86.33\% mean accuracy. Overall accuracy across VLMs is \lightbluepanel{66.76\%}, and performance varies widely by subject, with \iqlogo~IQ being the most challenging category for VLMs. 
    % Humans achieve 66.54\% on average, while the best-performing test-takers reach 99.60\%.
Human numbers are estimated from the official score distributions of the Vietnamese national high school graduation exam \cite{vnexpress_vnhgs_2024}.}
    \label{tab:overall_performance}
\end{table}

% \subsec{Results} We observe a clear gap between non-thinking proprietary and thinking proprietary VLMs (57.83\% vs.\ 78.34\%). Even in the thinking group, mean accuracy remains well below the best-test-taker reference (99.60\%, \cref{tab:overall_performance}), far from the reliability level that would be needed for downstream educational deployment of frontier VLMs whose English-language capabilities are already at Olympiad-gold level \cite{alphageometry2,gemini_imo_gold_2025,openai_competitive_programming_2025}. Such a gap is expected given that \vmmu questions are mainly STEM, driving, and IQ, which require multi-step deliberation rather than surface-level perception. Our conclusions are stable across runs: repeating the evaluation 5 times preserves model ranking and qualitative trends (\cref{sec:appendix_runs_consistency}).
% Moreover, the performance gap is not an artifact of answer parsing: SOTA VLMs follow the required output format reliably with near-perfect answer-extraction rates (\cref{sec:appendix_extraction_rate}).
% We also find that non-thinking VLMs are substantially more overconfident and less well-calibrated than thinking VLMs (\cref{sec:appendix_confidence_calibration}). Across the 5 SOTA proprietary VLMs, performance varies substantially by category: Puzzle and Traffic situation are consistently hardest, while Table and Traffic sign are comparatively easier (\cref{fig:categories_performance}).

\subsection{Results}

\paragraph{Results on open-source models} Open-source VLMs achieve only 37.35\% mean accuracy on \vmmu (\cref{tab:overall_performance}), drastically worse than the average Vietnamese high-school student (66.54\%). Even the strongest open-source models (\qwensmalllogo~\qwensmall at 52.13\% and \qwenlargelogo~\qwenlarge at 51.75\%) remain well below average-student level on every task. Within some families, larger variants improve accuracy (Gemma, Mistral). However, scaling to larger models is not uniformly beneficial (\eg, Aya, Qwen and Llama). \textbf{Overall, open-source VLMs perform far below the average student}. We diagnose the main cause of this failure in \cref{sec:results_ocr_main}.

\paragraph{Results on non-thinking closed-source models} Non-thinking closed-source VLMs do better, but still fall short of the average student (57.83\% mean vs.\ 66.54\%; \cref{tab:overall_performance}). Only \geminiflash exceeds the student-average reference (71.34\%), and both \sonnetfour and \gptfour remain near 50\%. \textbf{Non-thinking closed-source VLMs also underperform the average student}, so using a frontier proprietary VLM without thinking capabilities is not sufficient for \vmmu.

\paragraph{Results on thinking closed-source models} Enabling thinking in closed-source VLMs yields the largest single jump in our evaluation (78.34\% mean vs.\ 57.83\% for non-thinking proprietary; \cref{tab:overall_performance}). Thinking models are the only group that exceeds the average-student threshold, with the strongest model (\geminipro) reaching 86.33\%. However, mean accuracy remains well below the best-test-taker reference (99.60\%, \cref{tab:overall_performance}), far from the reliability level that would be needed to deploy these models as tutors or for automated checking, even though frontier VLMs already perform at Olympiad-gold level on English-language tasks \cite{alphageometry2,gemini_imo_gold_2025,openai_competitive_programming_2025}. Such a gap is expected given that \vmmu questions are mainly STEM, driving, and IQ, which require multi-step reasoning rather than surface-level perception. \textbf{Thinking VLMs outperform the average-student baseline, but remain far from the expert level.}

\paragraph{Discussions} Our conclusions are stable across runs: repeating the evaluation 5 times preserves model ranking and qualitative trends (\cref{sec:appendix_runs_consistency}). The performance gap is not an artifact of answer parsing either: SOTA VLMs follow the required output format reliably with near-perfect answer-extraction rates (\cref{sec:appendix_extraction_rate}). We also find that non-thinking VLMs are substantially more overconfident and less well-calibrated than thinking VLMs (\cref{sec:appendix_confidence_calibration}). Shuffling multiple-choice options yields only small accuracy changes ($\Delta=-0.33\%$) and preserves model rankings, which shows that option-position bias does not drive our findings (\cref{sec:results_shuffle}). 
Across models, performance varies substantially by domains: IQ is consistently hardest, while Geography and Math are comparatively easier (\cref{tab:overall_performance}). On the other hand, even the strongest thinking closed-source model (\geminipro at 86.33\%) is roughly 13 points below the best-test-taker reference. Thus, current VLMs are not yet ready for automatic checking or tutoring on Vietnamese multimodal questions. The remainder of this section analyzes \emph{why} models fail, in order to identify directions for improvement.

\section{Analyses and discussions}\label{sec:analyses_discussions}
\subsection{Open-source models fail at OCR while closed-source models do not}\label{sec:results_ocr_main}

To understand the performance gap between open-source and closed-source models observed in \cref{sec:results_main_integrated}, we analyze OCR capability, which is a prerequisite for any downstream multimodal reasoning over text-in-image inputs: a model that cannot reliably read the embedded Vietnamese text cannot be expected to answer correctly regardless of its reasoning ability.

% Here, we first verify that VLMs can effectively read Vietnamese text, establishing that poor performance on \vmmu stems from multimodal reasoning challenges rather than basic text recognition failures. Prior work shows SOTA VLMs excel at OCR tasks~\cite{reference_ocr}, so we test whether failures on multimodal Vietnamese exam questions result from inability to process Vietnamese text or from more complex multimodal understanding requirements.

\subsec{Experiments} We construct OCR ground truth via human-AI collaboration on a 210-question subset. We first ask \gptthink to transcribe all embedded text, including the prompt and answer options (see~\cref{fig:prompt_ocr,fig:prompt_ocr2}), then a human annotator verifies and edits the transcription to correct residual errors. We evaluate the same 10 open-source VLMs and 5 closed-source VLMs from \cref{sec:results_main_integrated} (see \cref{section:detail_experiment} for more details).

\begin{table}[h!]
\centering
\resizebox{\columnwidth}{!}{%
\begin{tabular}{lcccc}
\toprule
Model & BLEU (\%)$\uparrow$ & F1 $\uparrow$ & CER (\%)$\downarrow$ & WER (\%)$\downarrow$ \\
\midrule
\multicolumn{5}{c}{Open-source VLMs} \\
\ayasmalllogo~~\ayasmall
    & 2.72  & 0.20 & 229.30 & 253.09 \\
\ayalargelogo~~\ayalarge
    & 10.48 & 0.33 & 150.73 & 180.28 \\
\gemmasmalllogo~~\gemmasmall
    & 65.12 & 0.80 & 24.79  & 37.23 \\
\gemmalargelogo~~\gemmalarge
    & 78.60 & 0.87 & 16.18  & 23.89 \\
\qwensmalllogo~~\qwensmall
    & 77.50 & 0.88 & 19.05  & 24.17 \\
\qwenlargelogo~~\qwenlarge
    & 82.30 & 0.89 & 10.73  & 18.49 \\
\llamavericklogo~~\llamamaverick
    & 70.59 & 0.81 & 48.77  & 53.91 \\
\llamascoutlogo~~\llamascout
    & 38.95 & 0.54 & 217.92 & 211.96 \\
\mistralmediumlogo~~\mistralmedium
    & 81.23 & 0.89 & 9.08   & 16.11 \\
\mistralsmalllogo~~\mistralsmall
    & 72.28 & 0.84 & 19.39  & 28.41 \\
\midrule
\rowcolor{gray!20} Model Mean & 57.98 & 0.70 & 74.59 & 84.75 \\
\midrule
\multicolumn{5}{c}{SOTA closed-source VLMs} \\
\geminiflashlogo~~\geminiflash
    & 82.58 & 0.90 & 14.46 & 17.57 \\
\sonnetfourlogo~~\sonnetfour
    & 88.77 & 0.95 & 4.24  & 7.52 \\
\gptfourlogo~~\gptfour
    & 90.28 & 0.95 & 4.10  & 7.04 \\
\gptthinklogo~~\gptthink
    & 94.28 & 0.97 & 4.09 & 5.25 \\
\geminiprologo~~\geminipro
    & 89.13 & 0.95 & 6.04  & 9.27 \\
\midrule
\rowcolor{gray!20}  Model Mean & 89.01 & 0.94 & 6.59 & 9.33 \\
\bottomrule
\end{tabular}
}
\caption{OCR evaluation results across VLMs show a clear gap between open-source and SOTA closed-source models. This indicates that embedded-text OCR is a major bottleneck for open-source VLM performance.}
\label{tab:ocr_performance}
\end{table}

\subsec{Results} Across closed-source VLMs, OCR is consistently strong (mean BLEU 89.01\%, F1 0.94\%, CER 6.59\%, WER 9.33\%; \cref{tab:ocr_performance}). Reading Vietnamese text embedded in images is thus largely reliable for closed-source VLMs, so their downstream errors must instead come from multimodal grounding, \ie, locating the relevant visual evidence and reasoning over it, rather than from OCR failures. Therefore, we analyze these grounding/reasoning failures in the remaining subsections.

The picture is very different for open-source VLMs, which exhibit large and unstable OCR degradation. Several models collapse to unusable quality (\eg, \ayasmalllogo~\ayasmall and \llamascoutlogo~\llamascout both exceed 200\% CER) and some fail to follow the OCR instruction entirely, hallucinating image content. Even the strongest open-source VLMs scores below the weakest proprietary one on every metric (\cref{tab:ocr_performance}). These findings suggest that \textbf{open-source models cannot reliably read the embedded Vietnamese text in the first place, and improving OCR and instruction-following is therefore a prerequisite for competitive open-source performance on \vmmu.}

\subsection{Separating text from visual evidence improves multimodal performance}\label{sec:results_split_mm}
Integrated question images require VLMs to simultaneously (i) read dense rendered text and (ii) locate and interpret the relevant visual evidence (\eg, charts, tables, maps, diagrams) within the same canvas. This can introduce distraction from text-heavy regions and make it harder to attend to the correct evidence, even when models are capable of reading the text \cite{mmmu_pro}. We test whether separating the textual prompt from the visual evidence improves reliability by making both components easier to process.

\begin{figure}[h!]
    \centering
    \includegraphics[width=\linewidth]{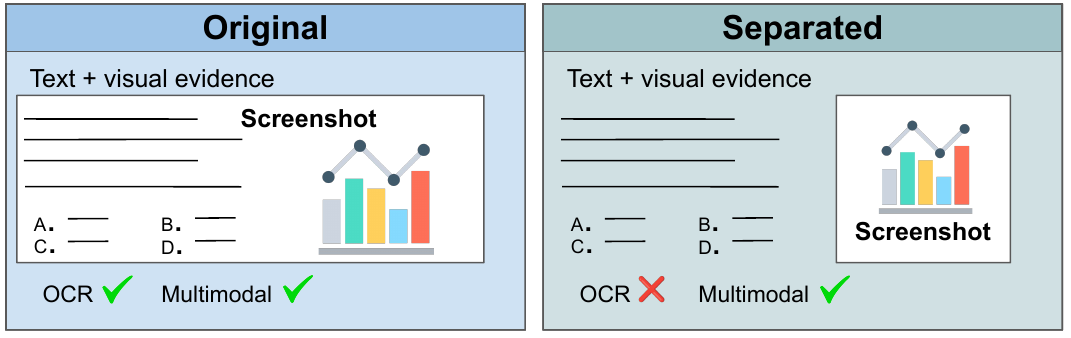}
    \caption{
Comparison between the original and separated evaluation settings in \vmmu.
In the Original setting, the question text, answer options, and visual evidence are rendered together in a single image.
In the separated setting, the question text and answer options are provided as plain text, while the visual evidence is provided as a separate image crop.
In our evaluation, \textbf{only the content shown on the white canvas is passed to VLMs as image input}, and any text outside the white canvas is provided as plain text.}
    \label{fig:vmmu_formats}
    % \vspace{-4mm}
\end{figure}

\subsec{Experiments}
We create the separated setting (see~\cref{fig:vmmu_formats}) for the \textit{full} \vmmu benchmark by separating the textual questions and answer options from the visual evidence. Since SOTA VLMs already perform strongly on OCR tasks (see \cref{sec:results_ocr_main}), we use \gptthink to automatically OCR-transcribe the textual elements for all 2{,}548 questions in \vmmu, and then manually crop the required visual evidence regions into separate images. Due to evaluation cost, we evaluate the separated setting on a representative subset of models rather than the full model suite. This subset covers the main model groups in our study: open-source VLMs, non-thinking closed-source VLMs, and thinking closed-source VLMs. We use the prompts shown in \cref{fig:prompt_split_mm}.

\begin{table}[h!]
    \centering
    \resizebox{\columnwidth}{!}{%
    \begin{tabular}{lcc}
    \toprule
    Model & Original & Separated ($\Delta$) \\
    \midrule
    \multicolumn{3}{c}{\textbf{Open-source VLMs}} \\
    \gemmalargelogo~~\gemmalarge
        & 39.09 & 47.98 \increase{8.89} \\
    \qwenlargelogo~~\qwenlarge
        & 51.75 & 52.30 \increase{0.55} \\
    \llamascoutlogo~~\llamascout
        & 49.06 & 52.55 \increase{3.49} \\
    \llamavericklogo~~\llamamaverick
        & 38.37 & 52.07 \increase{13.70} \\
    \mistralsmalllogo~~\mistralsmall
        & 31.88 & 45.82 \increase{13.94} \\
    \mistralmediumlogo~~\mistralmedium
        & 37.78 & 55.08 \increase{17.30} \\
    \midrule
    \rowcolor{gray!20}
    Model Mean & 41.32 & 50.97 \increase{9.65} \\
    \midrule

    \multicolumn{3}{c}{\textbf{Non-thinking closed-source VLMs}} \\
    \geminiflashlogo~~\geminiflash
        & 71.34 & 73.70 \increase{2.36} \\
    \sonnetfourlogo~~\sonnetfour
        & 50.58 & 64.44 \increase{13.86} \\
    \gptfourlogo~~\gptfour
        & 51.58 & 61.05 \increase{9.47} \\
    \midrule
    \rowcolor{gray!20}
    Model Mean & 57.83 & 66.40 \increase{8.56} \\
    \midrule

    \multicolumn{3}{c}{\textbf{Thinking closed-source VLMs}} \\
    \gptthinklogo~~\gptthink
        & 73.98 & 76.59 \increase{2.61} \\
    \geminiprologo~~\geminipro
        & 86.33 & 87.92 \increase{1.59} \\
    \midrule
    \rowcolor{gray!20}
    Model Mean & 80.16 & 82.26 \increase{2.10} \\
    \midrule

    \rowcolor{gray!30}
    Overall Mean & 52.89 & 60.86 \increase{7.97} \\
    \bottomrule
    \end{tabular}%
    }
    \caption{Separating textual content from visual evidence improves accuracy across all model groups. Open-source and non-thinking closed-source VLMs benefit the most, while thinking closed-source VLMs show smaller gains, suggesting that their performance is closer to saturation under the original image format.}
    \label{tab:original_vs_separated_all}
\end{table}

The separated setting improves the performance of every model (\increasenoparent{7.97} points on average; \cref{tab:original_vs_separated_all}). The gains are largest for open-source VLMs (\increasenoparent{9.65} points) and non-thinking closed-source VLMs (\increasenoparent{8.56} points), while thinking closed-source VLMs improve only modestly (\increasenoparent{2.10} points). This suggests that separating textual content from the image encourages models to rely more on visual reasoning while reducing distraction from text embedded in the image.

The benefit of the separated setting is a cleaner reasoning setup. Specifically, models no longer need to locate and transcribe text while simultaneously grounding the relevant visual evidence. Open-source VLMs benefit the most, likely because they are more affected by OCR and layout complexity, while non-thinking closed-source VLMs also gain substantially from reduced text-image interference. Thinking closed-source VLMs appear more robust to the original image format, so their smaller gains are consistent with partial saturation. Overall, \textbf{separating the textual questions and answer options from the visual evidence reduces text-in-image interference and makes grounded reasoning more reliable}, especially for open-source and non-thinking models.

% \begin{table}[h]
%     \centering
%     \resizebox{\columnwidth}{!}{%
%     \begin{tabular}{lcc}
%     \toprule
%     Model & Integrated-MM & Split-MM ($\Delta$) \\
%     \midrule
%     \multicolumn{3}{c}{Open-source VLMs} \\
%     \gemmalargelogo~~\gemmalarge
%         & 39.09 & 47.98 \increase{8.89} \\
%     \qwenlargelogo~~\qwenlarge
%         & 51.75 & 52.30 \increase{0.55} \\
%     \llamascoutlogo~~\llamascout
%         & 49.06 & 52.55 \increase{3.49} \\
%     \llamavericklogo~~\llamamaverick
%         & 38.37 & 52.07 \increase{13.70} \\
%     \mistralsmalllogo~~\mistralsmall
%         & 31.88 & 45.82 \increase{13.94} \\
%     \mistralmediumlogo~~\mistralmedium
%         & 37.78 & 55.08 \increase{17.30} \\
%     \midrule
%     \rowcolor{gray!20} Model Mean & 41.32 & 50.97 \increase{9.65} \\
%     \bottomrule
%     \end{tabular}%
%     }
%     \caption{Separating the question text from the visual evidence (Split-MM) consistently improves accuracy across open-source VLMs compared to the original Integrated-MM format, with a mean gain of \increasenoparent{9.65} points. The effect is broad-based: five of six models gain more than \increasenoparent{8} points, with the largest improvements on Mistral-Medium-3 (\increasenoparent{17.30}) and Mistral-Small-3.2-24B (\increasenoparent{13.94}). Qwen-2.5-VL-72B is the only outlier (\increasenoparent{0.55}), suggesting it already handles the Integrated-MM format competently.}
%     \label{tab:integrated_vs_split_opensource}
% \end{table}

\subsection{English translation does not improve performance}\label{sec:results_translation}
A natural hypothesis is that performance is limited by language understanding, and that translating the prompt and answer options into English could improve results for models trained predominantly on English data \cite{impact_propt_translation,lm_think_better_english,is_tranlsation_all_you_need}.

\begin{figure}
    \centering
    \includegraphics[width=\linewidth]{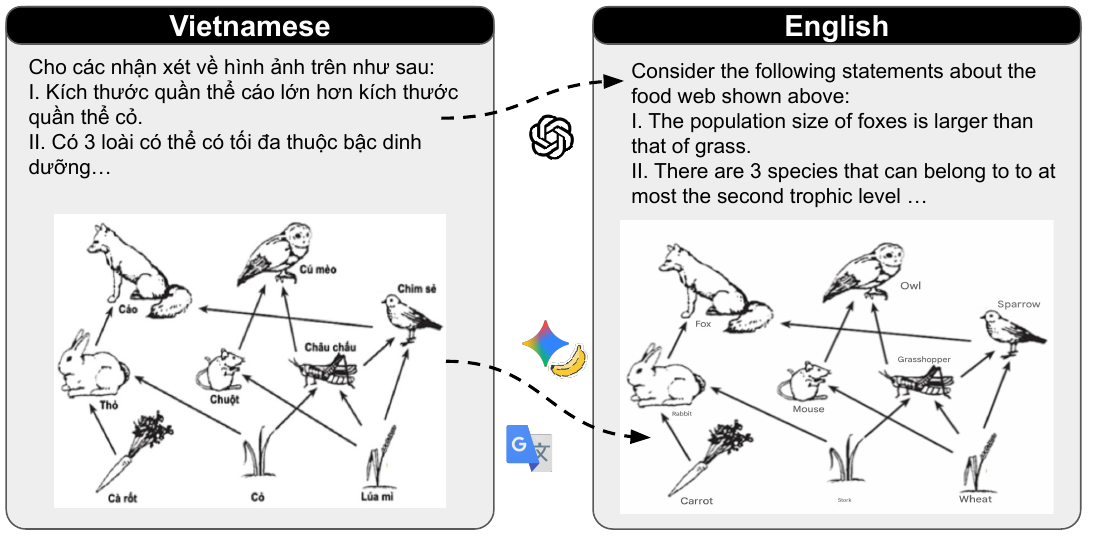}
    \caption{Our Vietnamese-to-English translation pipeline: \gptthink translates the question text, Google Translate (with Nano Banana 2 as fallback) translates the text in the image.}
    \label{fig:english_translation}
\end{figure}

\subsec{Experiments}
We use the same subset of 210 questions from \cref{sec:results_ocr_main}, evaluated in the separated setting from \cref{sec:results_split_mm}. For the text component (\ie, questions and answer options), we use \gptthink to translate the Vietnamese text into English (see \cref{fig:english_translation}). For the text embedded within the visual evidence crop, we first apply Google Translate's image translation feature \cite{googletranslate_image}. For cases where the translation is incorrect (6.19\%), as identified through human verification, we use Nano Banana 2 \cite{nanobanana2} to translate the image directly.

We evaluate this translation setting only on closed-source VLMs due to evaluation cost and because these models are the strongest models in our study. This provides a favorable test of the translation hypothesis: if translating Vietnamese inputs into English were beneficial, it should improve the performance of highly capable multilingual VLMs. We then pair the translated text with the translated visual evidence crop as input to each VLM to measure performance changes attributable to translation. Both the plain-text component and the text within the visual evidence crop are manually verified by native-speaker annotators to ensure translation quality. Only 6.19\% of visual-crop translations required correction, which we report as a reliability signal for the translated condition.

% \begin{figure}[h!]
%     \centering
%     % \vspace{-0.8cm}
%     \begin{tabular}{@{}c@{}}
%         \includegraphics[width=\linewidth]{images/vietnamese_vs_english_performance.pdf}\\[-0.6ex]
%         \makebox[\linewidth][l]{%
%             \makebox[0pt][c]{\hspace{2.7cm}\geminiflashlogo}%
%             \makebox[0pt][c]{\hspace{5cm}\sonnetfourlogo}%
%             \makebox[0pt][c]{\hspace{7.3cm}\gptfourlogo}%
%             \makebox[0pt][c]{\hspace{9.5cm}\gptthinklogo}%
%             \makebox[0pt][c]{\hspace{11.8cm}\geminiprologo}%
%             \makebox[0pt][c]{\hspace{13.8cm}Mean}%

%         }
%     \end{tabular}
%     \caption{Translating the Vietnamese questions into English reduces accuracy on average (\decreasenoparent{2.47} points). This shows that translation is not a reliable shortcut, even for VLMs heavily trained on English.}

%     \vspace{-0.3cm}
%     \label{tab:vietnamese_vs_english}
% \end{figure}

\begin{table}[h!]
    \centering
    \resizebox{\columnwidth}{!}{%
    \begin{tabular}{lcc}
    \toprule
    Model & Vietnamese & English ($\Delta$) \\
    \midrule
    \geminiflashlogo~~\geminiflash
        & 82.38 & 76.19 \decrease{6.19} \\
    \sonnetfourlogo~~\sonnetfour
        & 67.14 & 69.05 \increase{1.91} \\
    \gptfourlogo~~\gptfour
        & 67.14 & 66.19 \decrease{0.95} \\
    \gptthinklogo~~\gptthink
        & 81.43 & 78.10 \decrease{3.33} \\
    \geminiprologo~~\geminipro
        & 92.38 & 88.57 \decrease{3.81} \\
    \midrule
    \rowcolor{gray!20}
    Model Mean & 78.09 & 75.62 \decrease{2.47} \\
    \bottomrule
    \end{tabular}%
    }
    \caption{Translating Vietnamese questions into English reduces accuracy by \decreasenoparent{2.47} points on average. This suggests that English translation is not a reliable shortcut for \vmmu, even for VLMs heavily trained on English data.}
    \label{tab:vietnamese_vs_english}
\end{table}

\subsec{Results}
English translation consistently reduces the performance of most models (\decreasenoparent{2.47\%} accuracy; \cref{tab:vietnamese_vs_english,fig:eng_vmmu_examples}). 
% Since the visual evidence is identical in content and OCR is not required in Split-MM, these results indicate that switching to English does not resolve the remaining errors. 
We attribute this drop to two factors. First, translation introduces semantic drift, particularly for Vietnam-specific terminology in domains like traffic regulations, geography, and chemistry nomenclature, where direct English equivalents may not preserve the original meaning. Second, the consistent degradation across all models suggests that SOTA VLMs already possess sufficient Vietnamese understanding such that translation acts as a lossy intermediate step rather than a helpful shortcut. 

This finding goes against the common assumption that translating low-resource language input to English improves performance \cite{impact_propt_translation,lm_think_better_english,is_tranlsation_all_you_need}, and suggests that \textbf{high-performance Vietnamese multimodal reasoning requires native-language training rather than translation-based workarounds}.

\section{Conclusion}\label{sec:discussion_conclusion}

In this paper, we introduce \vmmu, a Vietnamese multitask multimodal benchmark consisting of 2{,}548 questions across seven domains. Our evaluation shows that current VLMs still struggle with Vietnamese multimodal reasoning. The dominant source of error is \emph{multimodal grounding and downstream reasoning}, rather than Vietnamese text recognition: closed-source models achieve reliable OCR performance, yet still make substantial errors on \vmmu (\cref{sec:results_ocr_main}). The benchmark remains challenging, although performance improves with test-time compute through thinking models and with the separated reformulation (\cref{sec:results_main_integrated,sec:results_split_mm}).

Our additional analyses show that removing visual evidence leaves accuracy above chance, but this residual performance reflects standard exam test-taking behavior, such as option elimination and domain knowledge, rather than benchmark contamination. We further find that reliance on such priors accounts for roughly 42\% of errors on average, where models commit to a prior answer despite the visual evidence being present (\cref{sec:results_knowledge_priors,sec:results_noimage_heuristics,sec:results_prior_overreliance}).

The remaining failures are also language-dependent. In particular, even when both the prompt and visual evidence are translated into English, accuracy consistently decreases (\cref{sec:results_translation}). This is because translation introduces semantic drift for Vietnam-specific terminology in traffic, geography, and chemistry, and adds a lossy intermediate step that strong VLMs do not need when they already handle Vietnamese reliably. These findings show that Vietnamese-native evaluation is necessary for measuring grounded multimodal reasoning in the settings users actually face. Future work should develop stronger joint text-vision reasoning under dense layouts and Vietnamese-centric instruction tuning that targets grounding and reasoning beyond what English-centric resources provide.

\section*{Limitations}\label{sec:limitations}
This work focuses on evaluation and failure analysis rather than proposing a new training or fine-tuning method, as \vmmu is the first benchmark of its kind for Vietnamese multimodal reasoning and large-scale Vietnamese multimodal training resources are not yet available. We hope \vmmu will encourage the community to build larger native Vietnamese multimodal datasets for training and adaptation of VLMs in low-resource languages.
% Our evaluation also reflects a fixed snapshot of current proprietary and open-source VLMs. Future frontier models may shift absolute accuracy, though we expect the qualitative trends we report (OCR vs.\ grounding vs.\ reasoning bottlenecks, thinking vs.\ non-thinking gap, reliance on text-only priors) to remain informative.
% In addition, many questions in \vmmu follow a standardized white-canvas style, which may underrepresent more realistic inputs such as natural photographs, mobile captures, cluttered documents, or handwritten content. Finally, while \vmmu spans seven domains, it does not yet cover other high-impact areas (\eg, medical or legal scenarios). Expanding domain coverage and realism is an important direction for future work.

\section*{Ethical Considerations}\label{sec:ethics}

% We adhere to the ACL Code of Ethics. 
Parts of this work were conducted via human-AI collaboration, where the human contributors are co-authors of this paper. This involvement posed minimal risk, and all participating authors agreed to contribute. The benchmark is curated from materials that are publicly published by governmental agencies and public educational institutions in Vietnam. All questions are used under Article 25 of Vietnam’s Intellectual Property Law, which permits non-commercial research use while copyright remains with the original creators. Prior Vietnamese benchmarking efforts have followed similar practices \cite{vmlu,vnhsge}. Furthermore, we will remove or replace any items upon request.
We used large language models to assist with annotating parts of the dataset, with all outputs subsequently verified and corrected when needed by human annotators. We also used LLM-based tools for light editing and coding assistance.

\section*{Acknowledgments}
This work was supported by the National Research Foundation of Korea (NRF) grant funded by the Korea government (MSIT) (RS-2025-00573160), and the ‘Advanced GPU Utilization Support Program’ funded by the Government of the Republic of Korea (Ministry of Science and ICT).

We also thank Rada Mihalcea (University of Michigan), Patrick Vibild, Hoang Minh Son (KAIST), and Duc-Vu Ngo for support, feedback and discussions of the earlier results. 
AV was supported by the Hyundai Motor Chung Mong-Koo Global Scholarship, and VTD was supported by Cohere Lab’s Research Grant.

\bigskip

\bibliography{custom}
\clearpage

\input{latex/appendixsec}

\end{document}

%% file: latex/appendixsec.tex
\clearpage
\appendix

\twocolumn[
  \begin{center}
    {\Large\bfseries Appendix for:\\\papertitle\par}
    \vspace{0.2in}
  \end{center}
]

\setcounter{tocdepth}{2}
\tableofcontents

\clearpage
% Your appendix sections start here:
% \section{Additional findings}
\twocolumn

\addtocontents{toc}{\protect\setcounter{tocdepth}{2}}

\section{Additional findings}
\subsection{Option-position bias does not drive our findings}\label{sec:results_shuffle}
Our earlier results report gaps between thinking and non-thinking VLMs (\cref{sec:results_main_integrated}) and between original and separated settings (\cref{sec:results_split_mm}). A natural concern is whether these differences are artifacts of multiple-choice positional biases, where models over-prefer certain option letters or benefit from consistent option ordering~\cite{order_bias,llm_not_robust_mc}. We test this by re-running the separated setting with shuffled answer options.

\subsec{Experiments} Starting from the separated setting in \cref{sec:results_split_mm}, we randomly shuffle the order of answer options for each question while keeping the prompt text and visual evidence crop unchanged.

\begin{figure}[h!]
    \centering
    \includegraphics[width=\linewidth]{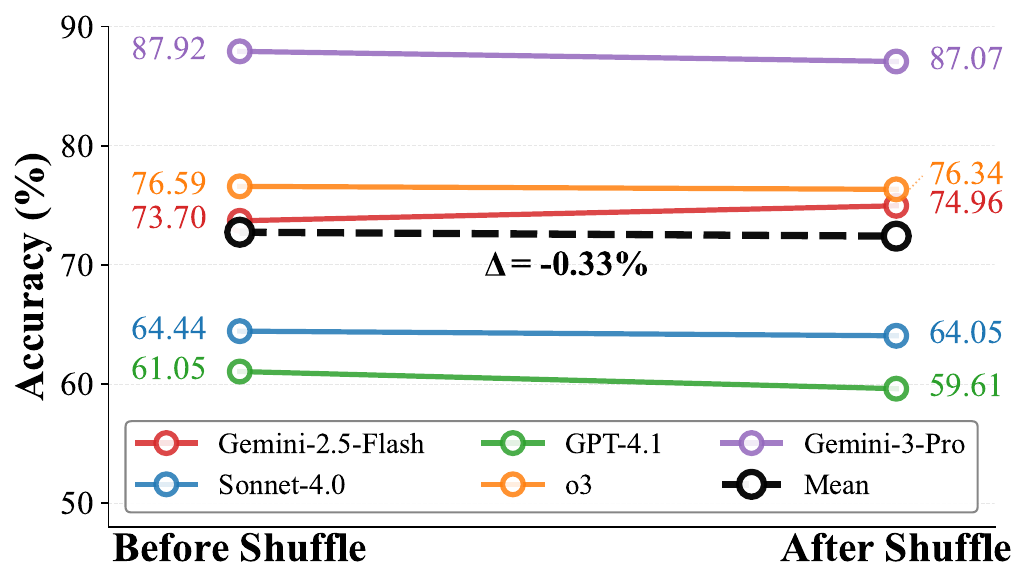}
    \caption{Accuracy before and after shuffling answer-option order. Changes are small ($\Delta=-0.33\%$) and model ranking is unchanged, showing limited sensitivity to option order.}

    \label{fig:before_after_shuffle}
    \vspace{-0.3cm}
\end{figure}

\subsec{Results} Randomly shuffling the multiple-choice answer options produces only small accuracy changes across VLMs ($\Delta=-0.33\%$; \cref{fig:before_after_shuffle}), and the relative ranking of models is unchanged. The gaps reported in \cref{sec:results_main_integrated,sec:results_split_mm} are therefore not explained by option-ordering heuristics and reflect genuine differences in multimodal reasoning.

\subsection{Image-required questions are harder than text-sufficient ones}\label{sec:results_visual_necessity}
\cref{sec:results_split_mm} shows that \emph{separated} setting improves accuracy, but this gain could be misattributed if a non-trivial portion of the benchmark is solvable from text alone. To confirm that \vmmu primarily evaluates multimodal reasoning, we explicitly measure how often the image is necessary for answering.

\subsec{Experiments} We label each question as \emph{image-required} or \emph{text-sufficient} based on whether the correct answer can be determined from the textual prompt and options alone. We use \gptthink as a judge to make an initial decision for all 2,548 questions, then apply human verification to ensure labeling quality. This process yields 2,363 image-required questions (92.74\%) and 185 text-sufficient questions (7.26\%). 
% We then report Original and separated accuracy separately on these two subsets using the same evaluation protocol as \cref{sec:results_main_integrated,sec:results_split_mm}.

%-=============================
\begin{table}[h!]
    \centering
    
    \scriptsize
    \resizebox{\columnwidth}{!}{%
    \begin{tabular}{lcccc}
    \toprule
    Model &
    \multicolumn{2}{c}{Image required} &
    \multicolumn{2}{c}{Text sufficient} \\
    \cmidrule(lr){2-3}\cmidrule(lr){4-5}
    & Original & Separated & Original & Separated \\
    \midrule
    \geminiflashlogo~~\textcolor{gemini_blue}{\geminiflash}
        & 69.92 & 72.75 \increase{2.83} & 92.47 & 91.78 \decrease{0.69} \\
    \sonnetfourlogo~~\textcolor{sonnet_orange}{\sonnetfour}
        & 49.82 & 63.22 \increase{13.40} & 71.38 & 88.06 \increase{16.68} \\
    \gptfourlogo~~\textcolor{gpt_green}{\gptfour}
        & 51.31 & 60.30 \increase{8.99} & 71.47 & 82.06 \increase{10.59} \\
    \gptthinklogo~~\textcolor{o3_purple}{\gptthink}
        & 72.49 & 75.48 \increase{2.99} & 95.26 & 93.70 \decrease{1.56} \\
    \geminiprologo~\geminipro
        & \textbf{85.62} & \textbf{87.39} \increase{1.77} & \textbf{97.57} & \textbf{96.89} \decrease{0.68} \\
    \midrule
    Mean
        & 65.83 & 71.83 \increase{6.00} & 85.63 & 90.50 \increase{4.42} \\
    \bottomrule
    \end{tabular}%
    }
    \caption{Separating the question text from the visual evidence (separated setting) improves performance on both image-required and text-sufficient questions (\increasenoparent{4-6} points). Gains are largest for non-thinking VLMs (\geminiflashlogo, \sonnetfourlogo, \gptfourlogo), especially on image-required questions where separated setting yields sizable improvements over original setting. Thinking VLMs (\gptthinklogo, \geminiprologo) benefit less and can slightly drop on text-sufficient questions, suggesting they already perform near saturation.}
    \label{tab:image_required_text_sufficient}
    \vspace{-2mm}
\end{table}

% ================================

\subsec{Results} In \cref{tab:image_required_text_sufficient}, image-required questions are substantially harder than text-sufficient ones in both original (65.83\% vs.\ 85.63\%) and separated settings (71.83\% vs.\ 90.50\%). The separated setting yields comparable gains on both image-required and text-sufficient subsets (\increasenoparent{4-6} percentage points). Despite this gain, accuracy on image-required questions on separated setting remains far from saturated (71.83\%; \cref{tab:image_required_text_sufficient}). Grounded multimodal reasoning, therefore, remains the primary bottleneck.

\subsection{Multiple runs do not change model ranking or our main conclusions}\label{sec:appendix_runs_consistency}

Since VLM decoding can be stochastic, we verify that our conclusions are not sensitive to a single evaluation run.

\subsec{Experiments}
We sample 210 questions from 7 domains (30 questions per domain) and repeat evaluation for each proprietary VLM across 5 independent runs.

\begin{table}[h!]
\centering
\resizebox{0.85\columnwidth}{!}{
\begin{tabular}{lc}
\toprule
Model & Accuracy ($\uparrow$) on 5 runs \\
\midrule
\geminiflashlogo~~\geminiflash & 75.70 $\pm$ 3.64 \\
\sonnetfourlogo~~\sonnetfour & 54.67 $\pm$ 4.16 \\
\gptfourlogo~~\gptfour & 60.19 $\pm$ 4.03 \\
\gptthinklogo~~\gptthink & 79.62 $\pm$ 4.03 \\
\geminiprologo~~\geminipro & 89.43 $\pm$ 1.73 \\ 
\midrule
\textbf{Mean} & 71.92 $\pm$ 3.52 \\
\bottomrule
\end{tabular}
}
\caption{Accuracy is stable across 5 independent runs (mean $71.92 \pm 3.52$), and model ranking is unchanged.}
\label{apptab:summary_runs_accuracy}
\end{table}

\subsec{Results}
Accuracy is stable across runs (mean $71.92 \pm 3.52$), and the relative ordering remains unchanged:
\geminiprologo~\geminipro is best (89.43$\pm$1.73), followed by \gptthinklogo~\gptthink (79.62$\pm$4.03) and
\geminiflashlogo~\geminiflash (75.70$\pm$3.64),
with \gptfourlogo~\gptfour (60.19$\pm$4.03) and \sonnetfourlogo~\sonnetfour (54.67$\pm$4.16) lower
(\cref{apptab:summary_runs_accuracy}). Thus, run-to-run variance does not affect our main conclusions.

\subsection{Non-thinking VLMs are overconfident}\label{sec:appendix_confidence_calibration}

Beyond accuracy, reliability depends on whether a model’s expressed confidence matches its correctness. Overconfident errors are particularly problematic when VLMs are used for decision-making. We therefore measure self-reported confidence and calibration on \vmmu.

% =================
\begin{table}[h!]
\centering
\resizebox{\columnwidth}{!}{
\begin{tabular}{lcc}
\toprule
Model & Confidence score (\%) $\uparrow$ & ECE score (\%) $\downarrow$ \\
\midrule
\geminiflashlogo~~\geminiflash & 93.22 & 33.76 \\
\sonnetfourlogo~~\sonnetfour & 76.08 & 28.46 \\
\gptfourlogo~~\gptfour & \textbf{96.76} & 48.20 \\
\gptthinklogo~~\gptthink & 56.72 & \textbf{17.66} \\
\midrule
\textbf{Mean} & 80.69 & 32.02 \\
\bottomrule
\end{tabular}
}
\caption{Non-thinking VLMs (\geminiflashlogo, \sonnetfourlogo, \gptfourlogo) report high self-confidence yet show substantially worse calibration (higher ECE), whereas the thinking VLM (\gptthinklogo) is less confident but best calibrated.}
\label{apptab:summary_conf_ece}
\end{table}

% =================
\subsec{Experiments}
We compute confidence and calibration on a single \vmmu evaluation run (the same outputs in \cref{sec:results_main_integrated}). Following \citet{b_score}, after the VLM selects an answer option, we ask in a second turn for a \emph{verbalized confidence} score $p_i \in [0,1]$ for its chosen option. We report the mean of these self-reported scores as the \textbf{confidence score}. We compute expected calibration error (ECE) directly from these per-example confidence scores by binning predictions into $M$ confidence bins $\{B_m\}_{m=1}^M$:
\begin{equation}
\text{ECE} = \sum_{m=1}^{M} \frac{|B_m|}{N}\, \left| \text{acc}(B_m) - \text{conf}(B_m) \right|,
\end{equation}
where $\text{acc}(B_m) = \frac{1}{|B_m|} \sum_{i \in B_m} \mathbf{1}(\hat{y}_i = y_i)$ and
$\text{conf}(B_m) = \frac{1}{|B_m|} \sum_{i \in B_m} p_i$. Lower ECE indicates better calibration.

\subsec{Results}
Non-thinking VLMs (\geminiflashlogo,\sonnetfourlogo,\gptfourlogo) report very high confidence (\eg, \gptfourlogo~\gptfour: 96.76\%; \geminiflashlogo~\geminiflash: 93.22\%), while the thinking model \gptthinklogo~\gptthink reports much lower confidence (56.72) despite achieving the highest accuracy (\cref{apptab:summary_runs_accuracy}). This mismatch is reflected in calibration: \gptfourlogo~\gptfour has the highest confidence but also the worst ECE (48.20), whereas \gptthinklogo~\gptthink has the best ECE (17.66). Overall, these results show that non-thinking VLMs tend to be overconfident,
while thinking VLMs are more conservative and better calibrated.

\subsection{Proprietary and open-source VLMs follow the answer format reliably}\label{sec:appendix_extraction_rate}

End-to-end accuracy can be underestimated if models frequently produce outputs that are not parseable into a valid multiple-choice option. We therefore measure whether evaluation failures could be explained by answer-formatting issues rather than reasoning.

\subsec{Experiments}
Using the same settings as the main evaluation (\cref{sec:results_main_integrated}), we compute the TRUE extraction rate: the percentage of model responses from which a valid answer option can be extracted under our post-processing rules. We report this rate for all 5 proprietary VLMs and all 10 open-source VLMs.

\begin{table}[h!]
\centering
\resizebox{0.9\columnwidth}{!}{
\begin{tabular}{lc}
\toprule
Model & Extraction rate (\%) $\uparrow$ \\
\midrule
\multicolumn{2}{c}{Open-source VLMs} \\
\ayasmalllogo~~\ayasmall       & 100.00 \\
\ayalargelogo~~\ayalarge       & 100.00 \\
\gemmasmalllogo~~\gemmasmall   & 94.10 \\
\gemmalargelogo~~\gemmalarge   & 97.47 \\
\qwensmalllogo~~\qwensmall     & 99.10 \\
\qwenlargelogo~~\qwenlarge     & 99.66 \\
\llamascoutlogo~~\llamascout   & 94.50 \\
\llamavericklogo~~\llamamaverick & 77.14 \\
\mistralsmalllogo~~\mistralsmall & 93.98 \\
\mistralmediumlogo~~\mistralmedium & 94.79 \\
\midrule
\rowcolor{gray!20} Model Mean & 95.07 \\
\midrule
\multicolumn{2}{c}{SOTA proprietary VLMs} \\
\geminiflashlogo~~\geminiflash & 98.09 \\
\sonnetfourlogo~~\sonnetfour   & \textbf{99.94} \\
\gptfourlogo~~\gptfour         & 99.66 \\
\gptthinklogo~~\gptthink       & 99.82 \\
\geminiprologo~~\geminipro     & 99.14 \\
\midrule
\rowcolor{gray!20} Model Mean & 99.33 \\
\bottomrule
\end{tabular}}
\caption{Both proprietary (mean 99.33\%) and open-source (mean 95.07\%) VLMs follow the required multiple-choice answer format reliably, so formatting is not a meaningful source of failures. The one notable exception is \llamavericklogo~\llamamaverick (77.14\%).}
\label{apptab:summary_extraction_rate}
\end{table}

\subsec{Results}
Proprietary VLMs achieve near-perfect extraction rates (mean 99.33\%; \cref{apptab:summary_extraction_rate}), and open-source VLMs are also reliable overall (mean 95.07\%), with 9 of 10 above 93\%. The only substantial outlier is \llamavericklogo~\llamamaverick (77.14\%); all other open-source models exceed 93\% and several (\ayasmalllogo~\ayasmall, \ayalargelogo~\ayalarge) achieve 100\%. Thus, answer-format compliance is not a meaningful contributor to the remaining failures for either group.

\subsection{Accuracy in the absence of visual evidence}\label{sec:results_knowledge_priors}

Having confirmed that answer extraction (\cref{sec:appendix_extraction_rate}) and calibration (\cref{sec:appendix_confidence_calibration}) do not explain the errors, we now examine whether visual evidence is actually essential for solving \vmmu. A common concern with multiple-choice multimodal benchmarks is that some accuracy may be recoverable without looking at the image, either through option-elimination heuristics, domain knowledge, or (in the worst case) contamination. To measure this residual, we evaluate all proprietary VLMs without visual evidence. A companion analysis (\cref{sec:results_noimage_heuristics}) then decomposes what drives the residual and shows that it reflects expected test-taking behavior rather than contamination.

\subsec{Experiments}
Starting from separated setting, we remove the visual evidence crop and provide only the text question and answer options (see \cref{fig:prompt_no_image}).
%This ablation is intentionally challenging because 93\% of VMMU questions are image-required (\cref{sec:results_visual_necessity}), i.e., the correct answer cannot be determined from the question text/options alone.

\begin{figure}
    \centering
    % \vspace{-0.8cm}
    \begin{tabular}{@{}c@{}}
        \includegraphics[width=\linewidth]{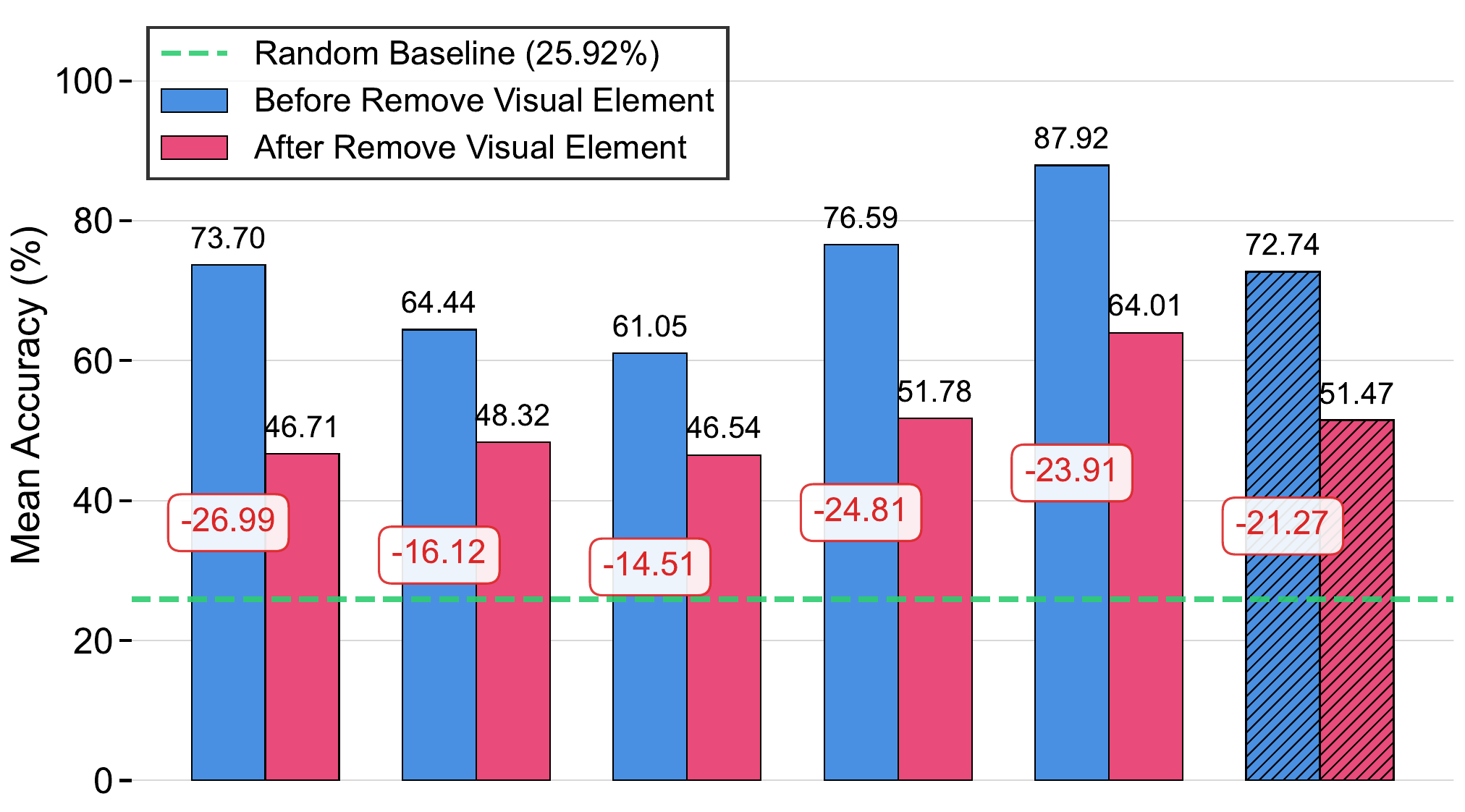}\\[-0.6ex]
        \makebox[\linewidth][l]{%
            \makebox[0pt][c]{\hspace{3cm}\geminiflashlogo}%
            \makebox[0pt][c]{\hspace{5.2cm}\sonnetfourlogo}%
            \makebox[0pt][c]{\hspace{7.4cm}\gptfourlogo}%
            \makebox[0pt][c]{\hspace{9.6cm}\gptthinklogo}%
            \makebox[0pt][c]{\hspace{11.8cm}\geminiprologo}%
            \makebox[0pt][c]{\hspace{13.8cm}Mean}%

        }
    \end{tabular}
    \caption{Accuracy with separated setting versus after removing the visual evidence. Accuracy drops by \decreasenoparent{21.27} percentage points on average, but remains above random chance (51.47\% vs.\ 25.92\%). This residual is decomposed in \cref{sec:results_noimage_heuristics}, where we show it reflects domain knowledge and option-elimination heuristics rather than contamination.}
    \vspace{-0.3cm}
    \label{fig:image_removal_impact}
\end{figure}

\subsec{Results}
Removing visual evidence significantly reduces accuracy for all VLMs (\decreasenoparent{21.27} percentage points; \cref{fig:image_removal_impact}), but performance does not collapse to chance (51.47\% vs.\ 25.92\%).
Notably, \geminiprologo~\geminipro still achieves 64.01\% accuracy without visual evidence.
% This indicates that thinking models are likely to recognize that the question is image-dependent and thus less able (or less willing, more random guesses) to rely on text-only shortcuts when the visual input is absent compared to non-thinking ones.
Qualitatively, remaining correct answers are typically produced by option elimination, domain knowledge, or incidental guessing (\cref{fig:geography_chart_types,fig:math_local_minimum,fig:iqtest_pyramid_pattern}), mirroring how human students answer multiple-choice exam items without full information.
In contrast to option shuffling (\cref{sec:results_shuffle}), which leaves accuracy essentially unchanged, this text-only ablation exposes the extent of such test-taking shortcuts. \cref{sec:results_noimage_heuristics} quantifies the contribution of these shortcuts and rules out contamination as an explanation.

\subsection{No-image accuracy reflects test-taking heuristics, not contamination}\label{sec:results_noimage_heuristics}

The 51.47\% mean accuracy observed when visual evidence is removed (\cref{sec:results_knowledge_priors}) could be misread as evidence that \vmmu is solvable from text alone or that the benchmark is contaminated. We perform three diagnostic checks, plus a hard-subset robustness analysis, that rule out this interpretation and attribute the residual to expected test-taking behavior.

\subsec{Experiments}
Starting from the no-image outputs of \cref{sec:results_knowledge_priors}, we manually categorize each response (human verification with LLM assistance) for two representative VLMs (\sonnetfour, \geminipro) into 4 types: (i) \emph{heuristic / domain-based} answers that eliminate implausible options or apply subject knowledge; (ii) \emph{random guessing}; (iii) \emph{self-reported memorization}, where the model explicitly claims to recall the question from exam repositories; and (iv) \emph{non-image-required} items. We then (a) report response-type breakdowns (\cref{tab:noimage_breakdown}), (b) measure accuracy \emph{within} each response type (\cref{tab:noimage_accuracy_by_type}), and (c) manually verify self-reported memorization claims and re-compute accuracy after filtering confirmed matches (\cref{tab:noimage_contamination_filter}).

\begin{table}[h!]
\centering
\resizebox{\columnwidth}{!}{%
\begin{tabular}{lcccc}
\toprule
Model & Heuristic/domain & Random & Self-memorization & Non-image-required \\
\midrule
\sonnetfourlogo~\sonnetfour & 63.11 & 29.08 & 0.00 & 7.26 \\
\geminiprologo~\geminipro   & 77.83 & 4.67  & 9.93 & 7.26 \\
\bottomrule
\end{tabular}%
}
\caption{Response-type breakdown (\%) on no-image outputs. Answers are dominated by domain knowledge and option-elimination heuristics, while self-reported memorization is rare.}
\label{tab:noimage_breakdown}
\end{table}

\subsec{Results}
We find that \textbf{(1) responses are dominated by heuristics, not memorization.} When the image is removed, 63.11\% (\sonnetfour) and 77.83\% (\geminipro) of responses are driven by domain knowledge and option elimination (see \cref{fig:geography_chart_types,fig:biology_phylogenetic_tree,fig:math_local_minimum}) or random (\cref{fig:iqtest_pyramid_pattern}), while self-reported memorization (see \cref{fig:driving_test_disabled_parking}) accounts for 0.00\% and 9.93\% respectively (\cref{tab:noimage_breakdown}).
\textbf{(2) Self-reported memorization does not reliably help.} Even when \geminipro explicitly claims to recall a question, it answers correctly only 54.15\% of the time (\cref{tab:noimage_accuracy_by_type}), showing that such claims are noisy and are not equivalent to having memorized the correct answer.
\textbf{(3) Verified contamination is small and does not affect conclusions.} Manually verifying all 253 self-reported memorization claims yields only 56 exact matches with the original exam source (2.1\% of \vmmu). Removing these items changes mean accuracy by 0.06\% (\cref{tab:noimage_contamination_filter}), within the run-to-run variance reported in \cref{sec:appendix_runs_consistency}.

\begin{table}[h!]
\centering
\resizebox{\columnwidth}{!}{%
\begin{tabular}{lcccc}
\toprule
Model & Heuristic/domain & Random & Self-memorization & Non-image-required \\
\midrule
\sonnetfourlogo~\sonnetfour & 52.30 & 36.30 & --    & 77.84 \\
\geminiprologo~\geminipro   & 67.73 & 25.21 & 54.15 & 94.05 \\
\bottomrule
\end{tabular}%
}
\caption{Accuracy (\%) within each response type. Even when \geminipro self-reports memorization, it answers correctly only 54.15\% of the time, indicating these self-reports are noisy and do not reliably map to memorized ground-truth answers.}
\label{tab:noimage_accuracy_by_type}
\end{table}

\begin{table}[h!]
\centering
\resizebox{\columnwidth}{!}{%
\begin{tabular}{lcc}
\toprule
Model & Before filtering & After filtering ($\Delta$) \\
\midrule
\geminiflashlogo~\geminiflash & 71.34 & 71.48 \increase{0.14} \\
\sonnetfourlogo~\sonnetfour   & 50.58 & 50.44 \decrease{0.14} \\
\gptfourlogo~\gptfour         & 51.58 & 51.71 \increase{0.13} \\
\gptthinklogo~\gptthink       & 73.98 & 74.03 \increase{0.05} \\
\geminiprologo~\geminipro     & 86.33 & 86.43 \increase{0.10} \\
\midrule
\textbf{Mean} & \textbf{66.76} & \textbf{66.82} \increase{0.06} \\
\bottomrule
\end{tabular}%
}
\caption{Overall accuracy (\%) on  before and after removing the 56 items (2.1\% of \vmmu) whose self-reported memorization was verified against the original exam source. Mean accuracy changes by only 0.06 pp, within run-to-run variance (\cref{sec:appendix_runs_consistency}).}
\label{tab:noimage_contamination_filter}
\end{table}

\begin{table}[h!]
\centering
\resizebox{\columnwidth}{!}{%
\begin{tabular}{lcc}
\toprule
Model & Full set (2{,}548) & Hard subset (2{,}051) ($\Delta$) \\
\midrule
\geminiflashlogo~\geminiflash & 71.34 & 66.92 \decrease{4.42} \\
\sonnetfourlogo~\sonnetfour   & 50.58 & 46.35 \decrease{4.23} \\
\gptfourlogo~\gptfour         & 51.58 & 46.74 \decrease{4.84} \\
\gptthinklogo~\gptthink       & 73.98 & 69.94 \decrease{4.04} \\
\geminiprologo~\geminipro     & 86.33 & 84.15 \decrease{2.18} \\
\midrule
\textbf{Mean} & \textbf{66.76} & \textbf{62.82} \decrease{3.94} \\
\bottomrule
\end{tabular}%
}
\caption{Accuracy (\%) on the full benchmark versus on the hard subset that conservatively removes the 497 questions all 5 proprietary VLMs answer correctly in the no-image condition. Accuracy drops for every model while preserving their relative ranking, confirming the benchmark remains challenging after the filter.}
\label{tab:noimage_hard_subset}
\end{table}

\paragraph{Hard-subset robustness.} As a stricter robustness check, we construct a hard subset by conservatively removing every question that \emph{all five} proprietary VLMs answer correctly in the no-image condition (n=497). On the remaining 2{,}051 questions, mean accuracy drops from 66.76\% to 62.82\% (\decreasenoparent{3.94} percentage points), with per-model drops ranging from 2.18 to 4.84 pp (\cref{tab:noimage_hard_subset}). Crucially, the relative ranking of models is preserved (\geminipro $>$ \gptthink $>$ \geminiflash $>$ \gptfour $>$ \sonnetfour), and the benchmark remains far from saturated.

Taken together, these analyses indicate that above-chance no-image accuracy reflects expected exam-style test-taking behavior (option elimination, domain knowledge, occasional guessing), not contamination or a failure of the benchmark's multimodal design.

\subsection{Over-reliance on priors causes failures when visual evidence conflicts}\label{sec:results_prior_overreliance}

Beyond sustaining above-chance no-image accuracy, reliance on priors also has a behavioral cost: when the image contradicts the model's prior expectations, the model may anchor on the prior and ignore the visual evidence \cite{vlmsbiased}.

\subsec{Experiments}
Using the per-question predictions from the no-image condition (\cref{sec:results_knowledge_priors}) and the separated setting (\cref{sec:results_split_mm}), we define a \emph{sticky-prior error} as a question where (i) the prediction is incorrect and (ii) it matches the no-image prediction. Such cases indicate that providing the visual evidence did not change the model's answer, consistent with the model committing to a text-only prior and disregarding the image. We report the sticky-prior error rate per model, both as a fraction of all questions and as a fraction of all errors.

\begin{table}[h!]
\centering
\resizebox{\columnwidth}{!}{%
\begin{tabular}{lccc}
\toprule
Model & Separated error rate & Sticky-prior errors & Sticky/All errors \\
\midrule
\geminiflashlogo~\geminiflash & 26.30 & 8.21  & 33.47 \\
\sonnetfourlogo~\sonnetfour   & 35.56 & 14.39 & 42.64 \\
\gptfourlogo~\gptfour         & 38.95 & 15.18 & 41.90 \\
\gptthinklogo~\gptthink       & 23.41 & 9.57  & 44.59 \\
\geminiprologo~\geminipro     & 12.08 & 5.54  & \textbf{47.16} \\
\midrule
\textbf{Mean} & \textbf{27.26} & \textbf{10.58} & \textbf{41.95} \\
\bottomrule
\end{tabular}%
}
\caption{Sticky-prior failure analysis (\%). \emph{Sticky-prior errors} are errors whose predicted option equals the no-image prediction, indicating that providing the image did not move the model off its prior. On average, 41.95\% of all errors are sticky-prior, i.e., nearly half of the remaining errors are cases where the visual evidence did not change the model's answer.}
\label{tab:sticky_prior_errors}
\end{table}

\subsec{Results}
Across the 5 proprietary VLMs, \textbf{41.95\% of all errors are sticky-prior errors on average} (\cref{tab:sticky_prior_errors}): nearly half of the remaining failures are cases where providing the image did not move the model off its prior, consistent with the visual input being effectively ignored. Notably, the strongest model (\geminiprologo~\geminipro, with the lowest overall error rate at 12.08\%) has the \emph{highest} sticky-prior share (47.16\%), indicating that even top VLMs are disproportionately susceptible to prior over-reliance when they do fail. \geminiflashlogo~\geminiflash is the only model below 40\% (33.47\%), suggesting its errors are more diverse and less dominated by prior commitment. The memorization excerpt in \cref{sec:results_noimage_heuristics} is a clear qualitative instance: the model commits to a prior answer inferred from familiar exam sources and does not revise it when reading the figure. Similar patterns recur in domains with strong conventional expectations (\eg, standard diagrams in physics and chemistry). This finding is completely aligned with \cite{vlmsbiased}, which showed that VLMs are heavily biased due to relying on knowledge priors that even ignore visual evidence.

Two implications follow. First, improving \vmmu performance is not solely a matter of better perception. It also requires models that down-weight priors when the image carries contradicting evidence. Second, the no-image ablation (\cref{sec:results_knowledge_priors}) should be read not as evidence of contamination, but as a diagnostic of how strong the text-only prior is, with \cref{tab:sticky_prior_errors} translating that prior into a measurable rate of behavioral failure under visual conflict.

\clearpage
%====================================================================================

\onecolumn
\section{Models and access details}\label{section:detail_experiment}

\begin{table*}[!htbp]
\centering
\resizebox{\textwidth}{!}{%
\begin{tabular}{llcl}
\toprule
\textbf{Model} & \textbf{Model Name} & \textbf{Temperature} & \textbf{Platform} \\
\midrule
\multicolumn{4}{c}{\textbf{Open-source VLMs}} \\ 
\ayasmalllogo~\ayasmall \cite{aya} & \texttt{c4ai-aya-vision-8b} & 0.0 & Cohere \\
\ayalargelogo~\ayalarge \cite{aya} & \texttt{c4ai-aya-vision-32b} & 0.0 & Cohere \\
\gemmasmalllogo~\gemmasmall \cite{gemma} & \texttt{gemma-3-4b-it} & 0.0 & Google AI Studio \\
\gemmalargelogo~\gemmalarge \cite{gemma} & \texttt{gemma-3-27b-it} & 0.0 & Google AI Studio \\
\mistralsmalllogo~\mistralsmall \cite{mistralsmall} & \texttt{mistral-small-3.2-24b-instruct} & 0.0 & OpenRouter\\
\mistralmediumlogo~\mistralmedium \cite{mistralmedium} & \texttt{mistral-medium-3} & 0.0 & OpenRouter \\
\qwensmalllogo~\qwensmall \cite{qwen2.5} & \texttt{qwen/qwen2.5-vl-32b-instruct} & 0.0 & OpenRouter \\
\qwenlargelogo~\qwenlarge \cite{qwen2.5} & \texttt{qwen/qwen2.5-vl-72b-instruct} & 0.0 & OpenRouter \\
\llamavericklogo~\llamamaverick \cite{llama4} & \texttt{llama-4-maverick} & 0.0 & OpenRouter \\
\llamascoutlogo~\llamascout \cite{llama4} & \texttt{llama-4-scout} & 0.0 & OpenRouter \\
\midrule
\multicolumn{4}{c}{\textbf{Non-thinking proprietary VLMs}} \\
\geminiflashlogo~\geminiflash \cite{gemini2_5} & \texttt{gemini-2.5-flash}  & 0.0 & Google AI Studio \\
\sonnetfourlogo~\sonnetfour & \texttt{claude-sonnet-4-20250514} & 0.0 & Anthropic \\
\gptfourlogo~\gptfour \cite{gpt4_1} & \texttt{gpt-4.1-2025-04-14} & 0.0 & OpenAI \\
\midrule
\multicolumn{4}{c}{\textbf{Thinking proprietary VLMs}} \\
\gptfivelogo~\gptfive* \cite{gpt5} & \texttt{gpt-5-2025-08-07} & N/A & OpenAI \\
\opusfour* \cite{opus4_6} & \texttt{claude-opus-4-6} & N/A & Anthropic \\
\gptthinklogo~\gptthink* \cite{openai2025o3o4mini} & \texttt{o3-2025-04-16} & N/A & OpenAI \\
\geminiprologo~\geminipro \cite{gemini_3} & \texttt{gemini-3.0-pro} & N/A & Google AI Studio/OpenRouter \\
\bottomrule
*reasoning\_effort: medium (default thinking mode setting)
\end{tabular}%
} 
\caption{VLMs configuration and platform details.}
\label{tab:vlm_models_config}

\vspace{0.3em}

\end{table*}

\clearpage

% We evaluate 4 state-of-the-art VLMs using the official APIs of each model with default settings, including one \textbf{thinking model} (\gptthink) and three \textbf{non-thinking models} (\sonnetfour, \gptfour, \geminiflash), along with 10 open-source VLMs to compare their performance and test their capabilities on the same dataset (see \cref{tab:vlm_models_config}). 

% \clearpage

%================================================================================

\section{Details of the prompts}
\label{section:prompt}

% The following are the prompts (see \cref{fig:prompt_question,fig:prompt_question2,fig:prompt_ocr,fig:prompt_ocr2}) used throughout the \vmmu experiments (\ie prompts for answering single multiple-choice questions, multiple questions in one image, and performing OCR on images in the dataset). All prompts are designed with bilingual English-Vietnamese versions and have specific instructions for output formatting to ensure consistency during the evaluation process.

\begin{figure*}[h!]
\centering
\begin{AIbox}{Original setting prompt}

\textbf{English prompt}:  
Answer the multiple-choice question in the image. Put your choice in curly brackets with the option letter, \eg \{Z\}.

\textbf{Vietnamese prompt}:  
Trả lời câu hỏi trắc nghiệm trong hình. Ghi đáp án bạn chọn trong dấu ngoặc nhọn với chữ cái tương ứng, ví dụ: \{Z\}.

\end{AIbox}
\caption{The Vietnamese and English prompts used for multiple-choice question answering tasks.}
\label{fig:prompt_question}
\end{figure*}

%====================================
\begin{figure*}[h!]
\centering
\begin{AIbox}{Original setting prompt (Multiple answers)}

\textbf{English prompt}:  
Answer the multiple-choice questions in the image. List your choices in curly brackets using option letters, \eg \{Y, Z\}.

\textbf{Vietnamese prompt}:  
Trả lời các câu hỏi trắc nghiệm trong hình. Liệt kê các đáp án bạn chọn trong dấu ngoặc nhọn, sử dụng chữ cái, ví dụ: \{Y, Z\}.

\end{AIbox}
\caption{The Vietnamese and English prompts used for question answering tasks on images containing multiple questions.}
\label{fig:prompt_question2}
\end{figure*}

%=========================================
\begin{figure*}[h!]
\centering
\begin{AIbox}{OCR prompt}
\textbf{English prompt:}  
Perform OCR on the image and return the result in this \texttt{JSON} format:
\begin{quote}
\begin{ttfamily}
\begin{flushleft}
\{\\
\quad ``Text``: ``Extract only the textual content, excluding figures and tables. Use Unicode math symbols (\eg  $\sqrt{}$, $\pi$) and '/' for fractions (\eg 1/5).``,\\
\quad ``Image description``: ``If figures or tables are present, describe them in Vietnamese; otherwise, leave blank.``\\
\}
\end{flushleft}
\end{ttfamily}
\end{quote}

\textbf{Vietnamese prompt:}  
Thực hiện OCR trên hình ảnh và trả về kết quả theo định dạng \texttt{JSON} sau:
\begin{quote}
\begin{ttfamily}
\begin{flushleft}
\{\\
\quad ``Text``: ``Chỉ trích xuất phần văn bản, bỏ qua hình vẽ và bảng biểu. Sử dụng ký hiệu toán học Unicode (\eg $\sqrt{}$, $\pi$) và '/' cho phân số (ví dụ: 1/5).``,\\
\quad ``Image description``: ``Nếu có hình vẽ hoặc bảng biểu, mô tả chúng bằng tiếng Việt; nếu không có, để trống.``\\
\}
\end{flushleft}
\end{ttfamily}
\end{quote}
\end{AIbox}
\caption{The Vietnamese and English prompts used for OCR tasks on individual question images in the dataset.}
\label{fig:prompt_ocr}
\end{figure*}

%============================================================================================
\begin{figure*}[h!]
\centering
\begin{AIbox}{OCR prompt (Multiple Answers)}
\textbf{English prompt:}  
Perform OCR on the image and return the result in this \texttt{JSON} format:
\begin{quote}
\begin{ttfamily}
\begin{flushleft}
\{\\
\quad ``Text``: ``Extract only the textual content, excluding figures and tables. Use Unicode math symbols (\eg $\sqrt{}$, $\pi$) and '/' for fractions (\eg 1/5).``,\\
\quad ``Image description``: ``If figures or tables are present, describe them in Vietnamese; otherwise, leave blank.``\\
\}
\end{flushleft}
\end{ttfamily}
\end{quote}
\textbf{Vietnamese prompt:}  
Thực hiện OCR trên hình ảnh và trả về kết quả theo định dạng \texttt{JSON} sau:
\begin{quote}
\begin{ttfamily}
\begin{flushleft}
\{\\
\quad ``Text``: ``Chỉ trích xuất phần văn bản, bỏ qua hình vẽ và bảng biểu. Sử dụng ký hiệu toán học Unicode (\eg $\sqrt{}$, $\pi$) và '/' cho phân số (ví dụ: 1/5).``,\\
\quad ``Image description``: ``Nếu có hình vẽ hoặc bảng biểu, mô tả chúng bằng tiếng Việt; nếu không có, để trống.``\\
\}
\end{flushleft}
\end{ttfamily}
\end{quote}
\end{AIbox}
\caption{The prompts used for OCR tasks on images containing two questions in the dataset.}
\label{fig:prompt_ocr2}
\end{figure*}

%=======================================
\begin{figure*}[h!]
\centering
\begin{AIbox}{Separated setting prompt}
\textbf{English prompt}:  
[Question content]\\
Based on the information in the image, answer the multiple-choice question above. Write your chosen answer in double curly brackets with the corresponding letter, for example: \{\{Z\}\}.

\textbf{Vietnamese prompt}:  
[Nội dung câu hỏi]\\
Dựa vào thông tin trong hình để trả lời câu hỏi trắc nghiệm ở trên. Ghi đáp án bạn chọn trong dấu ngoặc nhọn kép với chữ cái tương ứng, ví dụ: \{\{Z\}\}.
\end{AIbox}
\caption{The Vietnamese and English prompts used for question answering tasks on separated setting questions.}
\label{fig:prompt_split_mm}
\end{figure*}
%========================================================

\begin{figure}[h!]
\centering
\begin{AIbox}{Separated  setting prompt (after removing visual elements)}
\textbf{English prompt}:  
[Question content]
Based on the information provided, answer the multiple-choice question above. In all cases, you must select exactly one valid answer using the corresponding letter, even when data is insufficient or certain inference cannot be made. Write your chosen answer in double curly brackets with the corresponding letter, for example: \{\{Z\}\}.

\textbf{Vietnamese prompt}:  
[Nội dung câu hỏi]\\
Dựa vào thông tin được cung cấp để trả lời câu hỏi trắc nghiệm ở trên. Trong mọi trường hợp, bạn phải chọn đúng một đáp án hợp lệ bằng chữ cái tương ứng, kể cả khi thiếu dữ liệu hoặc không thể suy luận chắc chắn. Ghi đáp án bạn chọn trong dấu ngoặc nhọn kép với chữ cái tương ứng, ví dụ: \{\{Z\}\}.
\end{AIbox}
\caption{The Vietnamese and English prompts used for question answering tasks when image is required but not provided in the experiment.}
\label{fig:prompt_no_image}
\end{figure}

\clearpage

%=====================================================================
\twocolumn

\section{Dataset curation details }
\label{appsec:dataset_construction_method}

\subsection{Overview}
\label{appsec:dataset_overview}
We curate \vmmulogo~\vmmu with a semi-automated pipeline that converts heterogeneous public documents into
individual question images, then applies human verification for final quality control (\cref{fig:dataset_collection}).
The pipeline has 3 goals: (i) reliably extract question boundaries (including multi-page items),
(ii) retain only \emph{genuine} multimodal questions (where non-text visual evidence is necessary),
and (iii) produce consistent artifacts for evaluation (original and separated settings).

\subsection{Data source}\label{appsec:dataset_source}

We collect question sets from publicly available repositories and official releases (\cref{tab:vmmu_sources}).
% For STEM tasks (\mathlogo~Math, \physiclogo~Physics, \chemistrylogo~Chemistry, \biologylogo~Biology, \geographylogo~Geography), sources include official tests, MOET sample tests,
% and provincial mock tests. For \drivinglogo~Driving Test, we use official question banks and regulation-aligned materials.
% For \iqlogo~IQ Test, we manually curate public items.

\paragraph{Tasks 1-5: Mathematics, Physics, Chemistry, Biology, Geography} We systematically collected questions using Selenium web scraping from public repositories. Our dataset comprises questions from 3 primary sources: (1) Official Exam Tests from the Vietnamese National High School Graduation Examination, (2) Official Sample Tests issued by the Ministry of Education and Training (MOET), and (3) Mock Exam Tests from Provincial Departments of Education.

\paragraph{Task 6: Driving Test}  The Vietnamese Ministry of Transport (MOT) maintains official question banks containing 250 questions for A1 licenses (motorcycles) and 600 questions for B2 licenses (automobiles). To achieve A1 level requires a minimum of 21/25 correct answers (84\%), while B2 level requires 32/35 correct answers (91.4\%) and must not make errors on critical questions. From this broader collection of 850 total questions, we selected 367 multimodal multiple-choice questions that contain visual elements integrated with Vietnamese text. Each selected question presents traffic scenarios through illustrations, accompanied by Vietnamese text describing the situation and 2-4 multiple-choice options. 

\paragraph{Task 7: IQ Test} We collected 240 IQ test questions from \texttt{https://vndoc.com} through manual screenshot capture. These questions span various cognitive domains (\eg, spatial reasoning, pattern completion, logical sequences, numerical reasoning, abstract thinking). Each question presents 4-8 multiple-choice options embedded within the image.

\subsection{Collection and cleanup}
\label{appsec:dataset_collection}

\paragraph{Crawling and normalization}
We use a Selenium-based crawler to download documents in \texttt{.pdf/.doc/.docx}. All Word files are converted to PDF to standardize downstream processing.

\paragraph{Removing trailing solutions/answer keys}
Many documents append answer keys or solution explanations after the questions.
We drop these sections with a lightweight PDF segmentation step (\texttt{pypdf}): we scan pages for
high-precision termination cues (\eg, ``HẾT''; \textit{THE END}) and fallback cues marking auxiliary sections
(\eg, ``BẢNG ĐÁP ÁN''; \textit{ANSWER KEY}, and ``LỜI GIẢI CHI TIẾT''; \textit{DETAILED SOLUTION}).
We keep pages from the beginning up to the first matched cue.

\begin{table*}[h!]
\centering
\renewcommand{\arraystretch}{1.15}
\resizebox{\textwidth}{!}{%
\begin{tabular}{p{0.25\textwidth} p{0.4\textwidth} p{0.33\textwidth}}
\toprule
\textbf{Domain(s)} & \textbf{Source} & \textbf{Platform} \\
\midrule

\multirow{3}{*}{\begin{tabular}[t]{@{}l@{}}
\mathlogo~Mathematics, \physiclogo~Physics,\\
\chemistrylogo~Chemistry,\\
\biologylogo~Biology, \geographylogo~Geography
\end{tabular}}
& Official exam tests from the Vietnamese National High School Graduation Examination.
& \multirow{3}{*}{\begin{tabular}[t]{@{}l@{}}
\url{https://thuvienhoclieu.com}\\
\url{https://vndoc.vn}
\end{tabular}} \\
& Official sample tests issued by the Ministry of Education and Training (MOET). & \\
& Mock exam tests released by Provincial Departments of Education. & \\

\midrule

\drivinglogo~Driving Test
& Official question banks and regulation-aligned materials from the traffic police department and the Ministry of Transportation (MOT).
& \begin{tabular}[t]{@{}l@{}}
\url{https://daotaohoclaixeoto.com}\\
\url{https://tracuuphapluat.info}
\end{tabular} \\

\midrule

\iqlogo~IQ Test
& Manual collection from publicly available sources.
& \url{https://vndoc.com} \\
\bottomrule
\end{tabular}%
}
\caption{\vmmu data sources by subject group.}
\label{tab:vmmu_sources}
\end{table*}

\subsection{PDF rendering to page images}
\label{appsec:dataset_render}
We render each PDF page to a PNG image using \texttt{pdf2image} at 300 DPI. Each page image stores
document/page identifiers and pixel dimensions in a metadata file to support traceability during cropping.

\subsection{Question boundary detection and extraction}
\label{appsec:dataset_extraction}
\paragraph{Detecting question starts}
We run Tesseract OCR~\cite{tesseract} to obtain line-level text and bounding boxes. We then detect question starts via
Vietnamese layout markers using regex patterns (\eg \texttt{``Câu\textbackslash s?\textbackslash d+[:.]''}, where ``Câu'' means \textit{Question}).
We additionally detect group directives (e.g., ``Dùng dữ kiện cho câu 1--5''; \textit{Use the given data for Questions 1--5}) to preserve shared context for question groups.

\paragraph{Cropping and multi-page stitching}
Given detected markers, we construct a crop plan that maps each question to a page region.
If a question continues onto the next page, we stitch the cropped fragments vertically (\texttt{cv2.vconcat})
to form a single question image.

\subsection{Automatic modality classification (text-only vs. genuine multimodal)}
\label{appsec:dataset_modality}
To filter out weakly multimodal items (text-only screenshots), we classify each extracted question image
based on the presence of \emph{non-text visual structures} (\eg, charts/diagrams/tables/maps).
We binarize the image (grayscale + adaptive thresholding), detect connected components via contours,
and apply simple geometric rules to separate text-like blocks from non-text graphics.
If at least one sufficiently large, non-text-like region is detected, we label the item as multimodal, otherwise, it is labeled text-only.

\subsection{Human verification and ground-truth finalization}
\label{appsec:dataset_human}
Automated extraction is followed by human verification to ensure correct crop boundaries, correct modality
labels, and correct answer keys.

\begin{figure*}[h!]
    \centering
    \includegraphics[width=\textwidth]{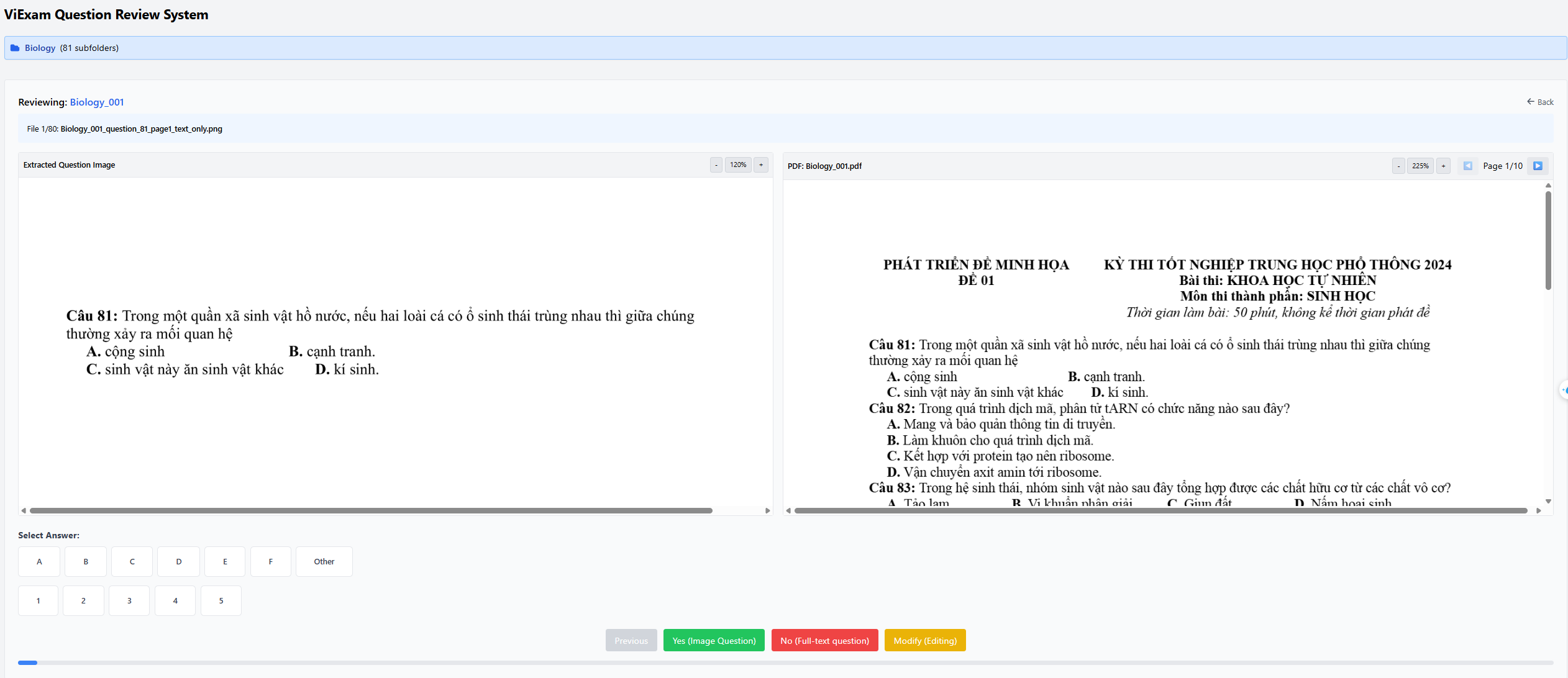}
    \caption{\vmmulogo~\vmmu question review system for human verification and ground-truth annotation. The interface shows an automatically extracted question crop (left) alongside the original source document rendered in-browser (right) for side-by-side checking. Annotators validate whether visual evidence is required, flag cropping errors, and record the correct answer choice.}

    \label{fig:vmmu_review_tool}
\end{figure*}

\paragraph{Review interface}
We built a review tool (React + TailwindCSS) that displays extracted question
images alongside the original source document for side-by-side inspection (\cref{fig:vmmu_review_tool}). Annotators can:
(i) accept/reject the multimodal label, (ii) flag cropping issues for manual repair, and (iii) record the correct answer.

\paragraph{Cross-verification}
For final answer keys, 3 native speakers independently verify each label.
We accept an answer only under unanimous agreement, and disagreements are resolved via discussion and re-checking
against the original document.

All annotators are co-authors of this paper, aged between 20 and 30, and provided informed consent for their participation and the intended use of their annotations.

\subsection{Release license}
\label{appsec:license}
\vmmu is released under the Creative Commons Attribution-NonCommercial 4.0 International license (CC BY-NC 4.0, \url{https://creativecommons.org/licenses/by-nc/4.0/}), which permits non-commercial use with attribution. Copyright remains with the original exam-material creators, as noted in the \nameref{sec:ethics}.

\clearpage
\onecolumn
\section{Qualitative results}
\label{section:qualitative}

\subsection{Math: Qualitative results}\label{sec:math_qualitative_results}
% ============================================
\begin{figure*}[h!]
\centering
% \begin{minipage}{0.95\textwidth}
    \begin{AIbox}{Math}

     {\centering
    \hspace{0\textwidth}\includegraphics[width=1\textwidth]{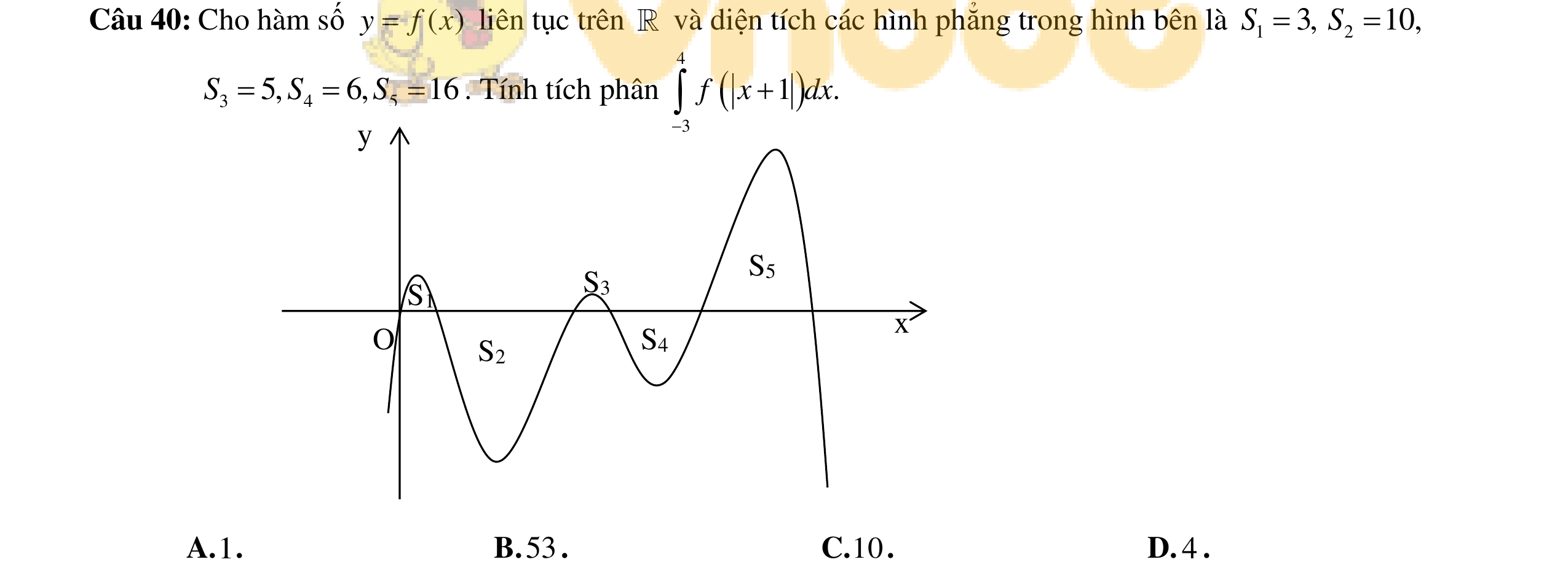}
    }

    % ========= Doule check ============
    \quest Given a continuous function \(y=f(x)\) on \(\mathbb{R}\) with the graph shown and the areas labeled in the figure
    \(S_1=3,\ S_2=10,\ S_3=5,\ S_4=6,\ S_5=16\). Compute
    \(\displaystyle \int_{-3}^{4} f\!\left(|x+1|\right)\,dx.\) \\
    A. 1 \quad B. 53 \quad C. 10 \quad D. 4

     \prompt Answer the multiple-choice question in the image. Put your choice in curly brackets with the option letter, \eg \{Z\}

     \groundtruth D \textcolor{ForestGreen}{\cmark}

    \hrulefill

    \geminiflashlogo~~\geminiflash: \{5\} \textcolor{red}{\xmark}

    \sonnetfourlogo~~\sonnetfour: \{C\} \textcolor{red}{\xmark}

    \gptfourlogo~~\gptfour: \{B\} \textcolor{red}{\xmark}

    \gptthinklogo~~\gptthink: \{B\} \textcolor{red}{\xmark}

    \geminiprologo~~\geminipro: \{A\} \textcolor{red}{\xmark}

    \end{AIbox}
% \end{minipage}

\caption{VLMs fail to correctly evaluate an integral defined by a graphical function combined with an absolute-value transformation. Notably, Gemini-2.5-Flash also fails to produce an answer in the required format.}
\label{fig:math2}
\end{figure*}

%====================
% ============================================
\begin{figure*}[h!]
\centering
% \begin{minipage}{0.95\textwidth}
    \begin{AIbox}{Math}

     {\centering
    \hspace{0\textwidth}\includegraphics[width=1\textwidth]{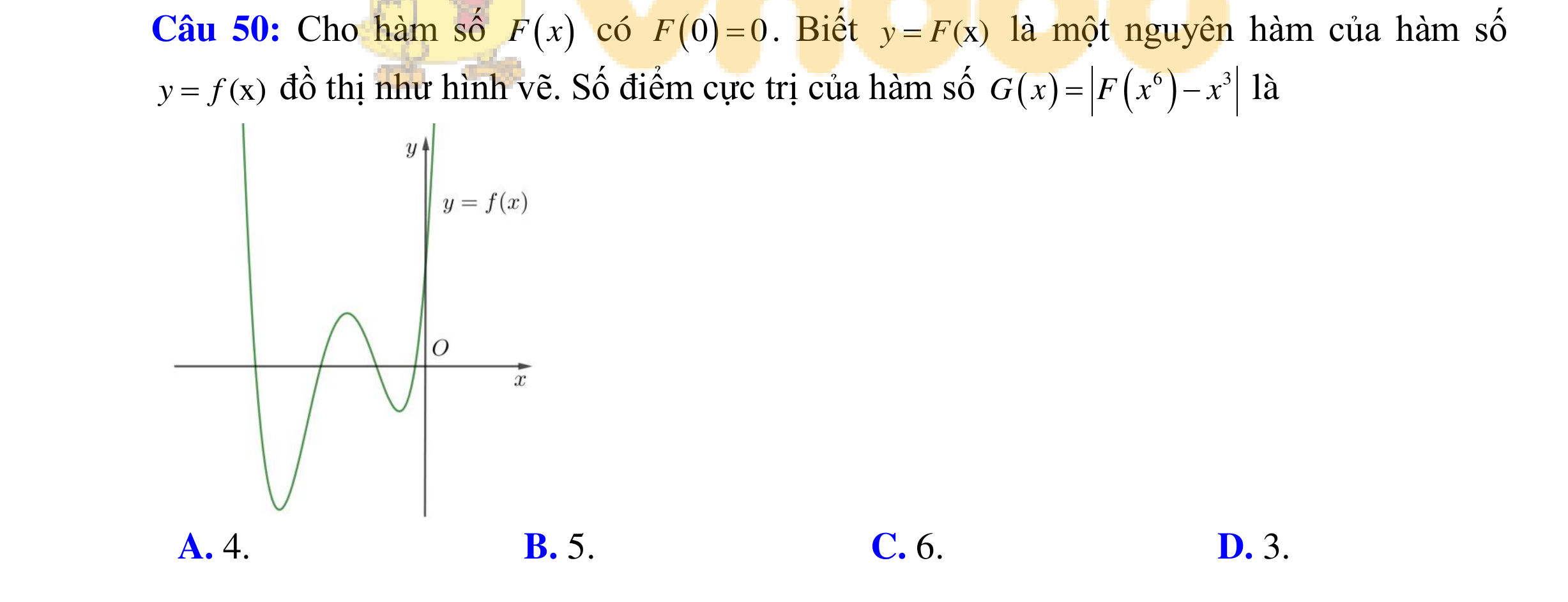}
    }

    % ========= Doule check ============
    \quest Let \(F(x)\) satisfy \(F(0)=0\). Suppose \(y=F(x)\) is an antiderivative of \(y=f(x)\) whose graph is shown.
    The number of local extrema of the function \(G(x)=\left|F(x^{6})-x^{3}\right|\) is \\
    A. 4 \quad B. 5 \quad C. 6 \quad D. 3

    \prompt Answer the multiple-choice question in the image. Put your choice in curly brackets with the option letter, \eg \{Z\}

    \groundtruth D \textcolor{ForestGreen}{\cmark}

    \hrulefill

    \geminiflashlogo~~\geminiflash: \{1\} \textcolor{red}{\xmark}
    
    \sonnetfourlogo~~\sonnetfour: \{B\} \textcolor{red}{\xmark}
    
    \gptfourlogo~~\gptfour: \{C\} \textcolor{red}{\xmark}
    
    \gptthinklogo~~\gptthink: \{B\} \textcolor{red}{\xmark}
    
    \geminiprologo~~\geminipro: \{A\} \textcolor{red}{\xmark}

    \end{AIbox}
% \end{minipage}

\caption{VLMs fail to correctly count the local extrema of an absolute-value composite function defined through an antiderivative and a provided graph. Gemini-2.5-Flash additionally fails to comply with the required answer format.}
\label{fig:math90}
\end{figure*}

%=================================
% \begin{figure*}[h!]
% \centering
%     \begin{AIbox}{Physics}

%      {\centering
%     \hspace{0\textwidth}\includegraphics[width=1\textwidth]{images/Physics_0063_test_029_question_17_page_3_key_C_vqa.png}
%     }

%     % ========= Double check ============
%     \quest A mechanical wave is travelling in the positive \(Ox\) direction as shown. What is the wavelength? \\
%     A. 120cm \quad B. 30cm \quad C. 90cm \quad D. 60cm

%     \prompt Answer the multiple-choice question in the image. Put your choice in curly brackets with the option letter, \eg \ \{Z\}

%     \groundtruth C \textcolor{ForestGreen}{\cmark}

%     \hrulefill

%     \geminiflash: \{D\} \textcolor{red}{\xmark}

%     \sonnetfour: \{B\} \textcolor{red}{\xmark}

%     \gptfour: \{D\} \textcolor{red}{\xmark}

%     \gptthink: \{D\} \textcolor{red}{\xmark}

%     \end{AIbox}

% \caption{VLMs fail on tasks of finding wavelengths in mechanical wave problems.
% }
% \label{fig:physics1}
% \end{figure*}

\clearpage
\subsection{Physics: Qualitative results}\label{sec:physics_qualitative_results}

% ============================================
\begin{figure*}[h!]
\centering
% \begin{minipage}{0.95\textwidth}
    \begin{AIbox}{Physics}

     {\centering
    \hspace{0\textwidth}\includegraphics[width=1\textwidth]{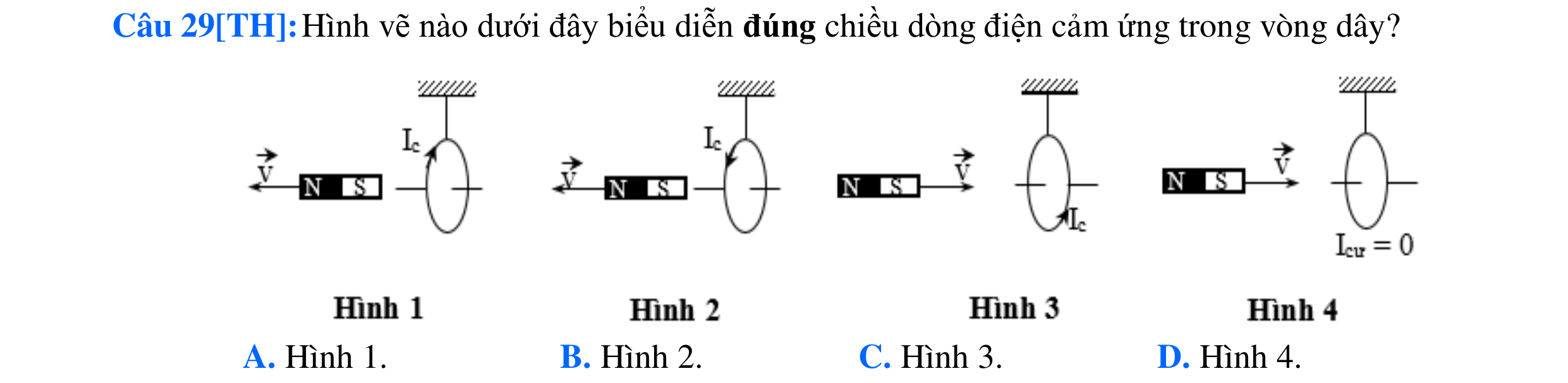}
    }

    % ========= Doule check ============
    \quest Which diagram below correctly represents the direction of the induced current in the loop? \\
    A. Figure 1 \quad B. Figure 2 \quad C. Figure 3 \quad D. Figure 4

    \prompt Answer the multiple-choice question in the image. Put your choice in curly brackets with the option letter, \eg \{Z\}

    \groundtruth A \textcolor{ForestGreen}{\cmark}

    \hrulefill

    \geminiflashlogo~~\geminiflash: \{C\} \textcolor{red}{\xmark}
    
    \sonnetfourlogo~~\sonnetfour: \{D\} \textcolor{red}{\xmark}
    
    \gptfourlogo~~\gptfour: \{B\} \textcolor{red}{\xmark}
    
    \gptthinklogo~~\gptthink: \{B\} \textcolor{red}{\xmark}
    
    \geminiprologo~~\geminipro: \{C\} \textcolor{red}{\xmark}

    \end{AIbox}
% \end{minipage}

\caption{VLMs struggle with electromagnetic induction: identifying the correct direction of induced current from the motion of a magnet relative to a loop.}
\label{fig:physics1}
\end{figure*}

% ============================================
\begin{figure*}[h!]
\centering
% \begin{minipage}{0.95\textwidth}
    \begin{AIbox}{Physics}

     {\centering
    \hspace{0\textwidth}\includegraphics[width=1\textwidth]{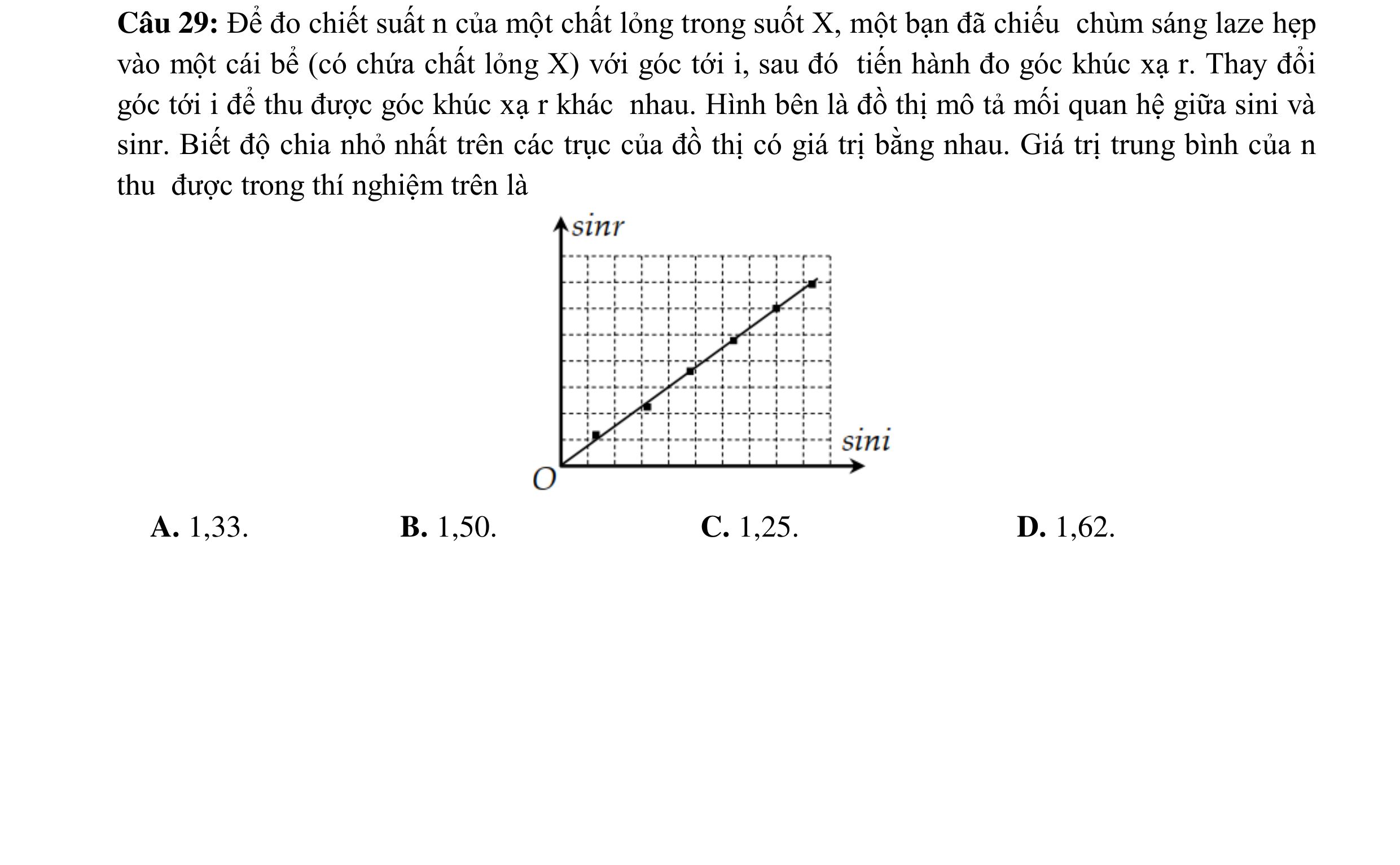}
    }

    % ========= Doule check ============
    \quest To measure the refractive index \(n\) of a transparent liquid \(X\), a narrow laser beam is incident on a container
    (filled with liquid \(X\)) at angle \(i\), and the refraction angle \(r\) is measured. Varying \(i\) yields different \(r\).
    The plot shows the relationship between \(\sin i\) and \(\sin r\), and the smallest grid spacing on both axes is the same.
    The average value of \(n\) obtained from the experiment is \\
    A. 1.33 \quad B. 1.50 \quad C. 1.25 \quad D. 1.62

    \prompt Answer the multiple-choice question in the image. Put your choice in curly brackets with the option letter, \eg \{Z\}

    \groundtruth A \textcolor{ForestGreen}{\cmark}

    \hrulefill

    \geminiflashlogo~~\geminiflash: \{C\} \textcolor{red}{\xmark}

    \sonnetfourlogo~~\sonnetfour: \{C\} \textcolor{red}{\xmark}

    \gptfourlogo~~\gptfour: \{B\} \textcolor{red}{\xmark}

    \gptthinklogo~~\gptthink: \{B\} \textcolor{red}{\xmark}

    \geminiprologo~~\geminipro: \{B\} \textcolor{red}
{\xmark}

    \end{AIbox}
% \end{minipage}

\caption{ VLMs struggle with interpreting refraction graphs, estimating the refractive index of a liquid from a plot of $\sin i$ versus $\sin r$.}

\label{fig:physics2}
\end{figure*}

%=======================================
% ============================================

\clearpage
\subsection{Chemistry: Qualitative results}\label{sec:chemistry_qualitative_results}

\begin{figure*}[h!]
\centering
% \begin{minipage}{0.95\textwidth}
    \begin{AIbox}{Chemistry}

     {\centering
    \hspace{0\textwidth}\includegraphics[width=1\textwidth]{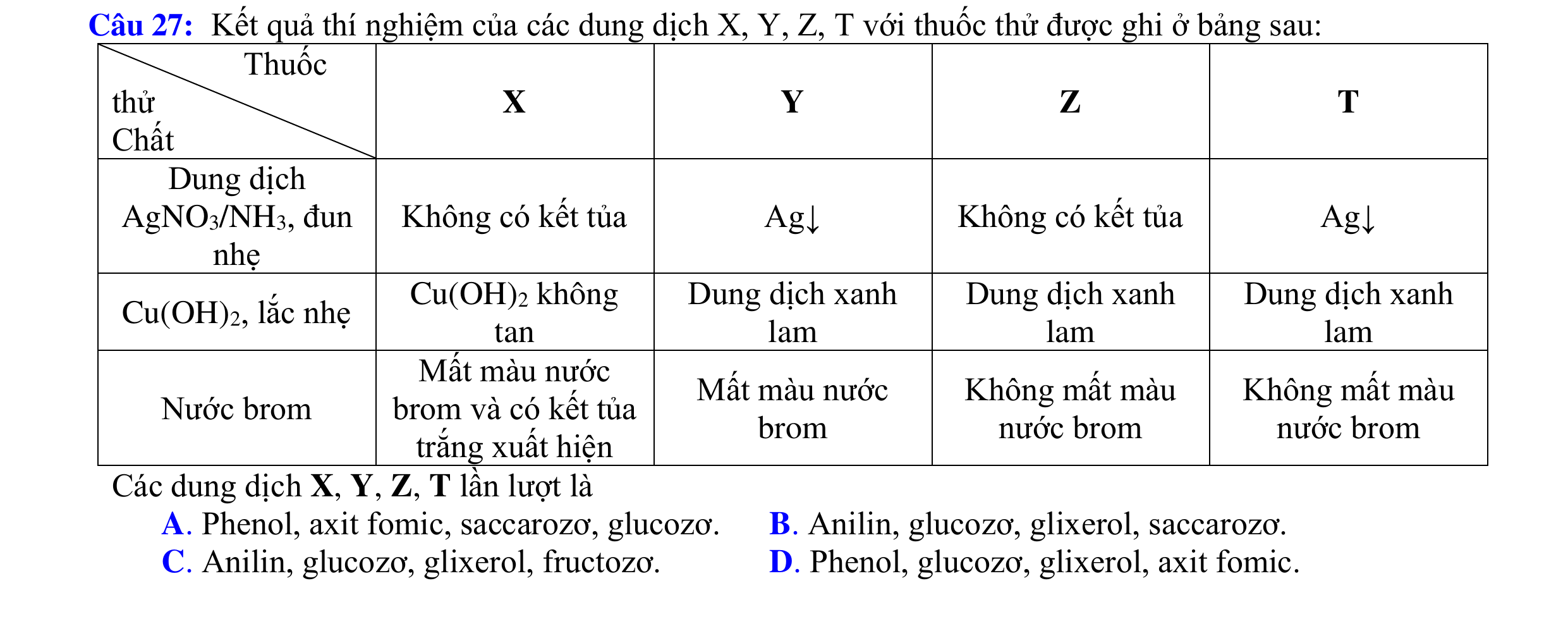}
    }

    % ========= Doule check ============
    \quest The experimental results for solutions \(X, Y, Z, T\) with different reagents are shown in the table.
    Identify the substances corresponding to \(X, Y, Z, T\) in order. \\
    A. Phenol, formic acid, sucrose, glucose. \quad
    B. Aniline, glucose, glycerol, sucrose. \quad
    C. Aniline, glucose, glycerol, fructose. \quad
    D. Phenol, glucose, glycerol, formic acid.

    \prompt Answer the multiple-choice question in the image. Put your choice in curly brackets with the option letter, \eg \{Z\}

    \groundtruth B \textcolor{ForestGreen}{\cmark}

    \hrulefill

    \geminiflashlogo~~\geminiflash: \{C\} \textcolor{red}{\xmark}

    \sonnetfourlogo~~\sonnetfour: \{D\} \textcolor{red}{\xmark}

    \gptfourlogo~~\gptfour: \{A\} \textcolor{red}{\xmark}

    \gptthinklogo~~\gptthink: \{D\} \textcolor{red}
{\xmark}

    \geminipro: \{C\} \textcolor{red}{\xmark}

    \end{AIbox}
% \end{minipage}

\caption{VLMs fail to reason over a multi-reagent qualitative analysis table to identify unknown organic compounds.}
\label{fig:chemistry1}
\end{figure*}

% ============================================
\begin{figure*}[h!]
\centering
% \begin{minipage}{0.95\textwidth}
    \begin{AIbox}{Chemistry}

     {\centering
    \hspace{0\textwidth}\includegraphics[width=1\textwidth]{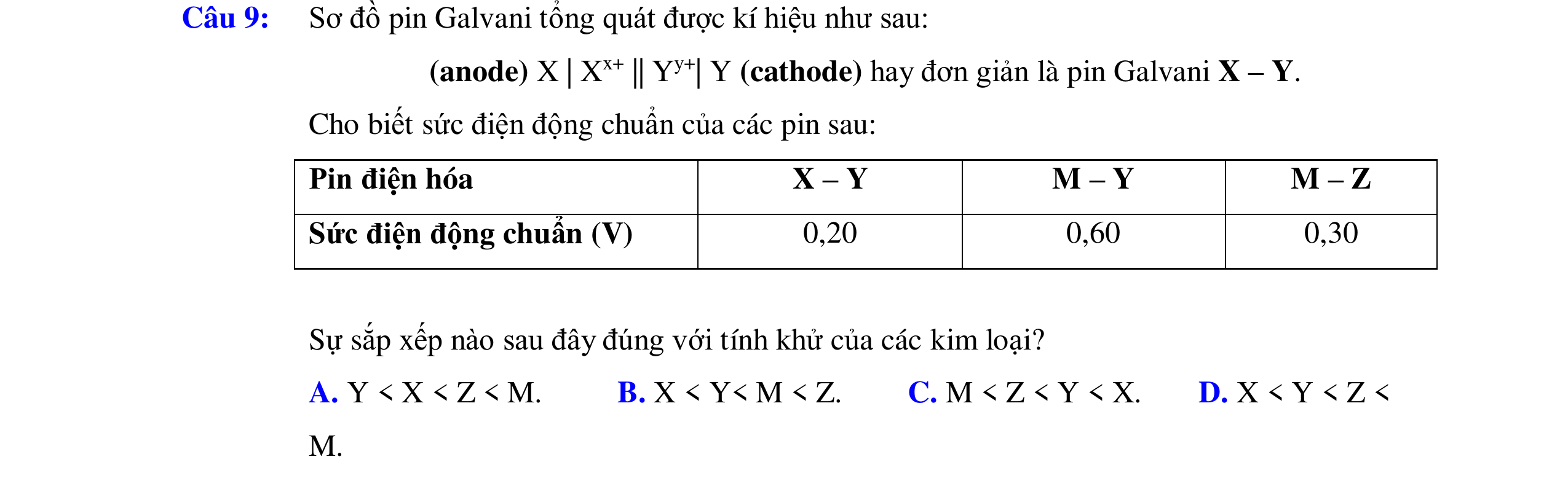}
    }

    % ========= Doule check ============
    \quest A galvanic cell is denoted by \((\text{anode})\ X\mid X^{x+}\parallel Y^{y+}\mid Y\ (\text{cathode})\), i.e., the cell \(X-Y\).
    Given the standard EMFs of the following cells: \(X-Y = 0.20\,\text{V}\), \(M-Y = 0.60\,\text{V}\), and \(M-Z = 0.30\,\text{V}\).
    Which ordering correctly describes the reducing strength of the metals? \\
    A. \(Y<X<Z<M\). \quad
    B. \(X<Y<M<Z\). \quad
    C. \(M<Z<Y<X\). \quad
    D. \(X<Y<Z<M\).

    \prompt Answer the multiple-choice question in the image. Put your choice in curly brackets with the option letter, \eg \{Z\}

    \groundtruth C \textcolor{ForestGreen}{\cmark}

    \hrulefill

    \geminiflash: \{A\} \textcolor{red}{\xmark}

    \sonnetfour: \{B\} \textcolor{red}{\xmark}

    \gptfour: \{D\} \textcolor{red}{\xmark}

    \gptthink: \{A\} \textcolor{red}{\xmark}

    \geminipro: \{A\} \textcolor{red}{\xmark}

    \end{AIbox}
% \end{minipage}

\caption{VLMs fail to correctly reason over a pedigree involving X-linked recessive mutations and genetic recombination, resulting in incorrect evaluation of multiple statements.}
\label{fig:chemistry2}
\end{figure*}

\clearpage
\subsection{Biology: Qualitative results}\label{sec:biology_qualitative_results}

% ============================================
\begin{figure*}[h!]
\centering
% \begin{minipage}{0.95\textwidth}
    \begin{AIbox}{Biology}

     {\centering
    \hspace{0\textwidth}\includegraphics[width=1\textwidth]{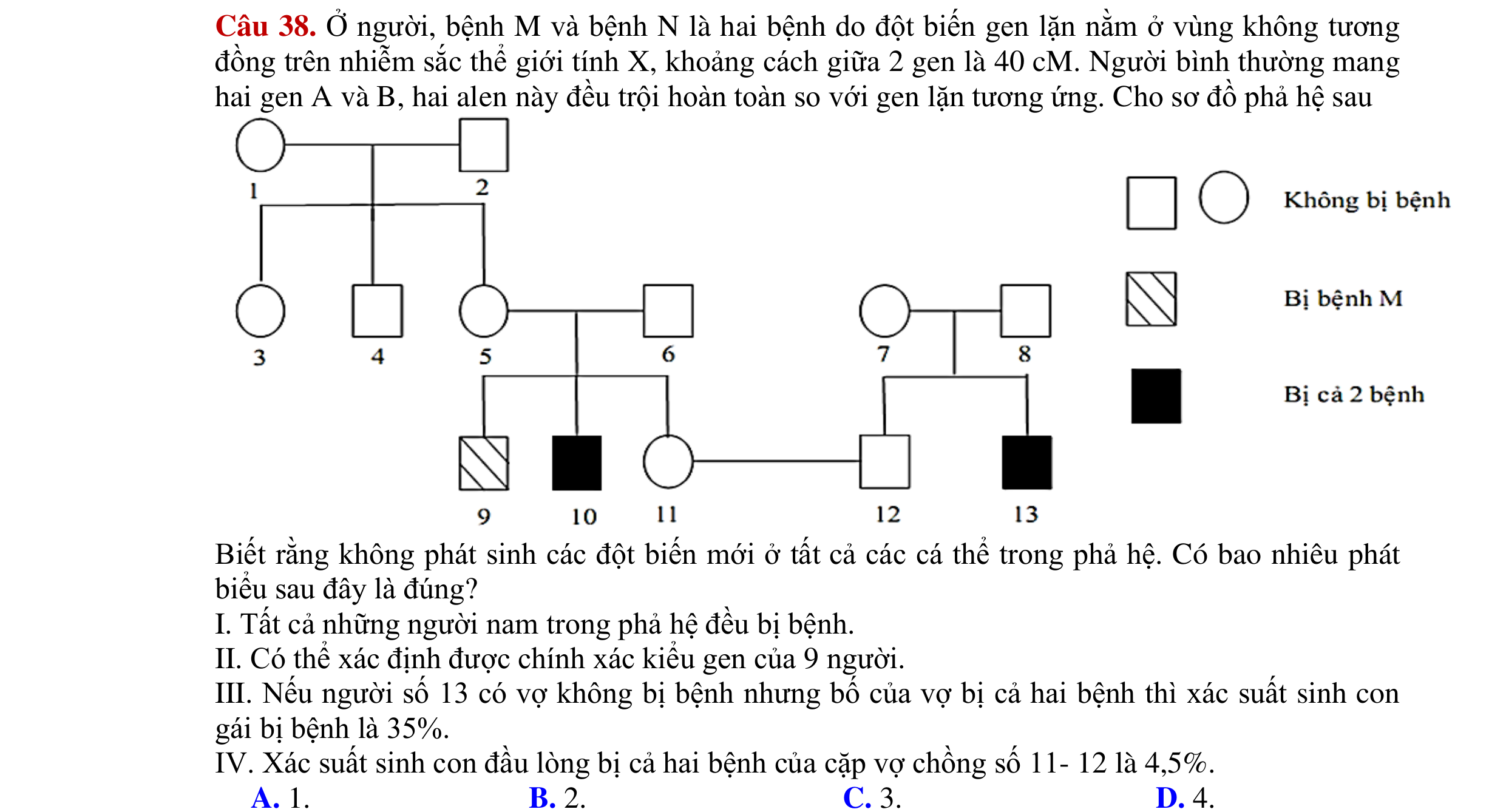}
    }

    % ========= Doule check ============
    \quest In humans, diseases \(M\) and \(N\) are caused by recessive mutations in two non-homologous regions on the X chromosome,
    with the distance between the two genes being 40 cM. Normal individuals carry alleles \(A\) and \(B\), both completely dominant
    over their corresponding recessive alleles. Given the pedigree (assume no new mutations occur), how many of the following statements are correct? \\
    I. All males in the pedigree are affected. \\
    II. It is possible to determine the exact genotypes of 9 individuals. \\
    III. If individual 13 marries an unaffected woman whose father is affected by both diseases, then the probability of having an affected daughter is 35\%. \\
    IV. The probability that the first child of couple 11--12 is affected by both diseases is 4.5\%. \\
    A. 1 \quad B. 2 \quad C. 3 \quad D. 4

    \prompt Answer the multiple-choice question in the image. Put your choice in curly brackets with the option letter, \eg \{Z\}

    \groundtruth C \textcolor{ForestGreen}{\cmark}

    \hrulefill

    \geminiflashlogo~~\geminiflash: \{B\} \textcolor{red}{\xmark}

    \sonnetfourlogo~~\sonnetfour: \{B\} \textcolor{red}{\xmark}

    \gptfourlogo~~\gptfour: \{B\} \textcolor{red}{\xmark}

    \gptthinklogo~~\gptthink: \{B\} \textcolor{red}{\xmark}

    \geminiprologo~~\geminipro: \{A\} \textcolor{red}{\xmark}

    \end{AIbox}
% \end{minipage}

\caption{VLMs fail on pedigree-based genetics reasoning involving X-linked recessive mutations and recombination, leading to systematic errors in counting correct statements.}
\label{fig:biology1}
\end{figure*}

% ============================================
\begin{figure*}[h!]
\centering
% \begin{minipage}{0.95\textwidth}
    \begin{AIbox}{Biology}

     {\centering
    \hspace{0\textwidth}\includegraphics[width=1\textwidth]{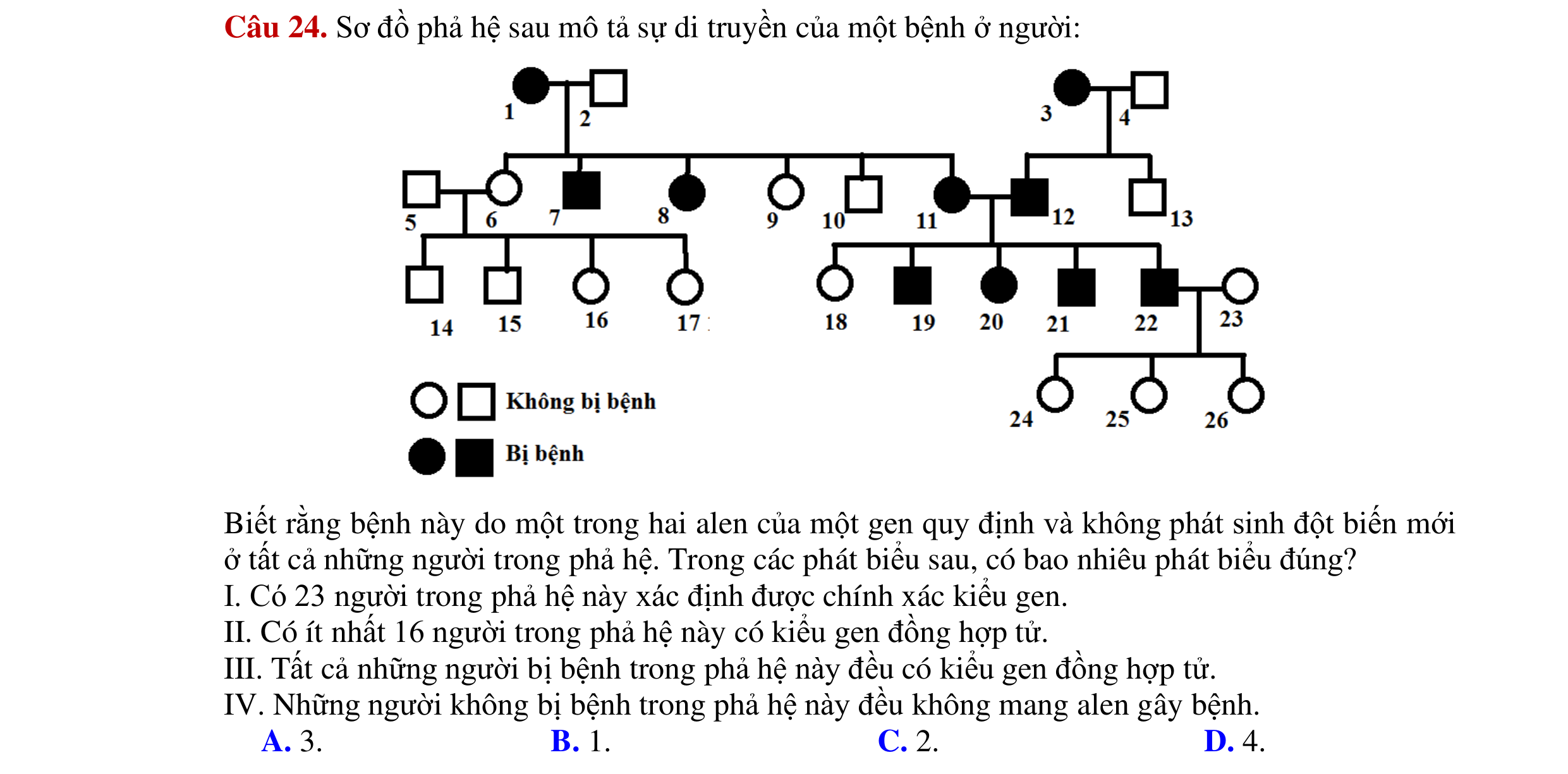}
    }

    % ========= Doule check ============
    \quest The pedigree describes the inheritance of a human disease caused by one of the alleles of a gene, and no new mutations occur.
    Among the following statements, how many are correct? \\
    I. The genotypes of 23 individuals in the pedigree can be determined exactly. \\
    II. At least 16 individuals in the pedigree are homozygous. \\
    III. All affected individuals in the pedigree are homozygous. \\
    IV. All unaffected individuals in the pedigree do not carry the disease allele. \\
    A. 3 \quad B. 1 \quad C. 2 \quad D. 4

    \prompt Answer the multiple-choice question in the image. Put your choice in curly brackets with the option letter, \eg \{Z\}

    \groundtruth D \textcolor{ForestGreen}{\cmark}

    \hrulefill

    \geminiflashlogo~~\geminiflash: \{C\} \textcolor{red}{\xmark}

    \sonnetfourlogo~~\sonnetfour: \{C\} \textcolor{red}{\xmark}

    \gptfourlogo~~\gptfour: \{C\} \textcolor{red}{\xmark}

    \gptthinklogo~~\gptthink: \{A\} \textcolor{red}{\xmark}

    \geminiprologo~~\geminipro: \{A\} \textcolor{red}{\xmark}

    \end{AIbox}
% \end{minipage}

\caption{VLMs fail on pedigree interpretation and genotype reasoning, leading to incorrect counting of valid statements about inheritance.}
\label{fig:biology2}
\end{figure*}

%===========================

\clearpage
\subsection{Geography: Qualitative results}\label{sec:geography_qualitative_results}
%========================
\begin{figure*}[h!]
\centering
    \begin{AIbox}{Geography}

     {\centering
    \hspace{0\textwidth}\includegraphics[width=1\textwidth]{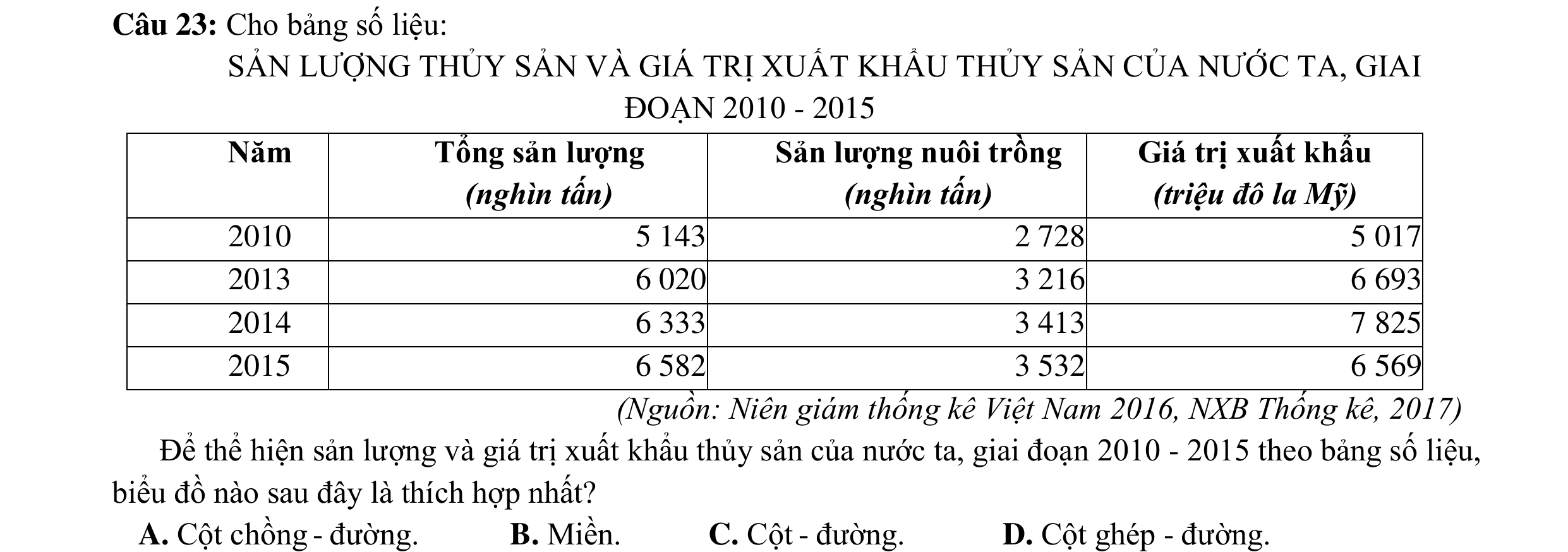}
    }

    % ========= Double check ============
    \quest  
    Using the table of fish-production volume and export value for Viet Nam (2010 – 2015), which type of chart is \emph{most appropriate} to display all three data series?\\
    A. Stacked bar chart - line chart. \quad B. Area chart. \quad C. Bar chart - line chart. \quad D. Grouped chart - line chart.

    \prompt Answer the multiple-choice question in the image. Put your choice in curly brackets with the option letter, \eg \ \{Z\}

    \groundtruth A \textcolor{ForestGreen}{\cmark}

    \hrulefill

    \geminiflashlogo~~\geminiflash: \{D\} \textcolor{red}{\xmark}

    \sonnetfourlogo~~\sonnetfour: \{D\} \textcolor{red}{\xmark}

    \gptfourlogo~~\gptfour: \{D\} \textcolor{red}{\xmark}

    \gptthinklogo~~\gptthink: \{C\} \textcolor{red}{\xmark}

    \geminiprologo~~\geminipro: \{C\} \textcolor{red}{\xmark}

    \end{AIbox}

\caption{Failure case where VLMs incorrectly choose chart types for displaying multiple data series in a geography context.}
\label{fig:geography1}
\end{figure*}

%===============
% ============================================
\begin{figure*}[h!]
\centering
% \begin{minipage}{0.95\textwidth}
    \begin{AIbox}{Geography}

     {\centering
    \hspace{0\textwidth}\includegraphics[width=1\textwidth]{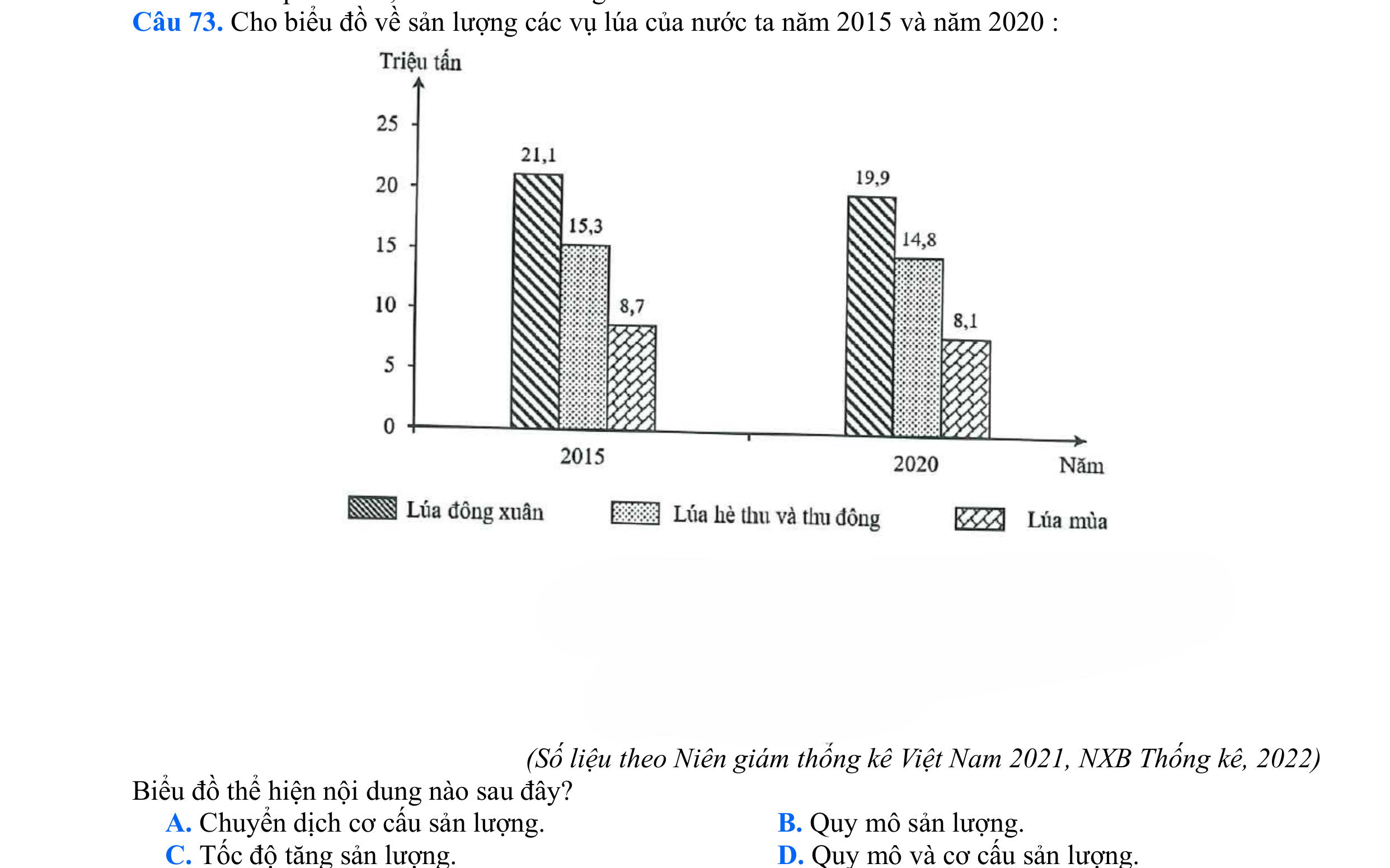}
    }

    % ========= Doule check ============
    \quest The bar chart shows Vietnam's rice output by season in 2015 and 2020 (million tons).
    What content does the chart best represent? \\
    A. Structural change in output composition. \quad
    B. Scale of output. \quad
    C. Output growth rate. \quad
    D. Both output scale and composition.

    \prompt Answer the multiple-choice question in the image. Put your choice in curly brackets with the option letter, \eg \{Z\}

    \groundtruth C \textcolor{ForestGreen}{\cmark}

    \hrulefill

    \geminiflashlogo~~\geminiflash: \{D\} \textcolor{red}{\xmark}

    \sonnetfourlogo~~\sonnetfour: \{A\} \textcolor{red}{\xmark}

    \gptfourlogo~~\gptfour: \{D\} \textcolor{red}{\xmark}

    \gptthinklogo~~\gptthink: \{D\} \textcolor{red}{\xmark}

    \geminiprologo~~\geminipro: \{B\} \textcolor{red}{\xmark}

    \end{AIbox}
% \end{minipage}

\caption{VLMs fail to interpret what a grouped bar chart is intended to convey (growth rate vs.\ scale/composition).}
\label{fig:geography2}
\end{figure*}

%======================
\clearpage
\subsection{Driving Test: Qualitative results}\label{sec:driving_test_qualitative_results}

% ============================================
\begin{figure*}[h!]
\centering
% \begin{minipage}{0.95\textwidth}
    \begin{AIbox}{DrivingTest}

     {\centering
    \hspace{0\textwidth}\includegraphics[width=1\textwidth]{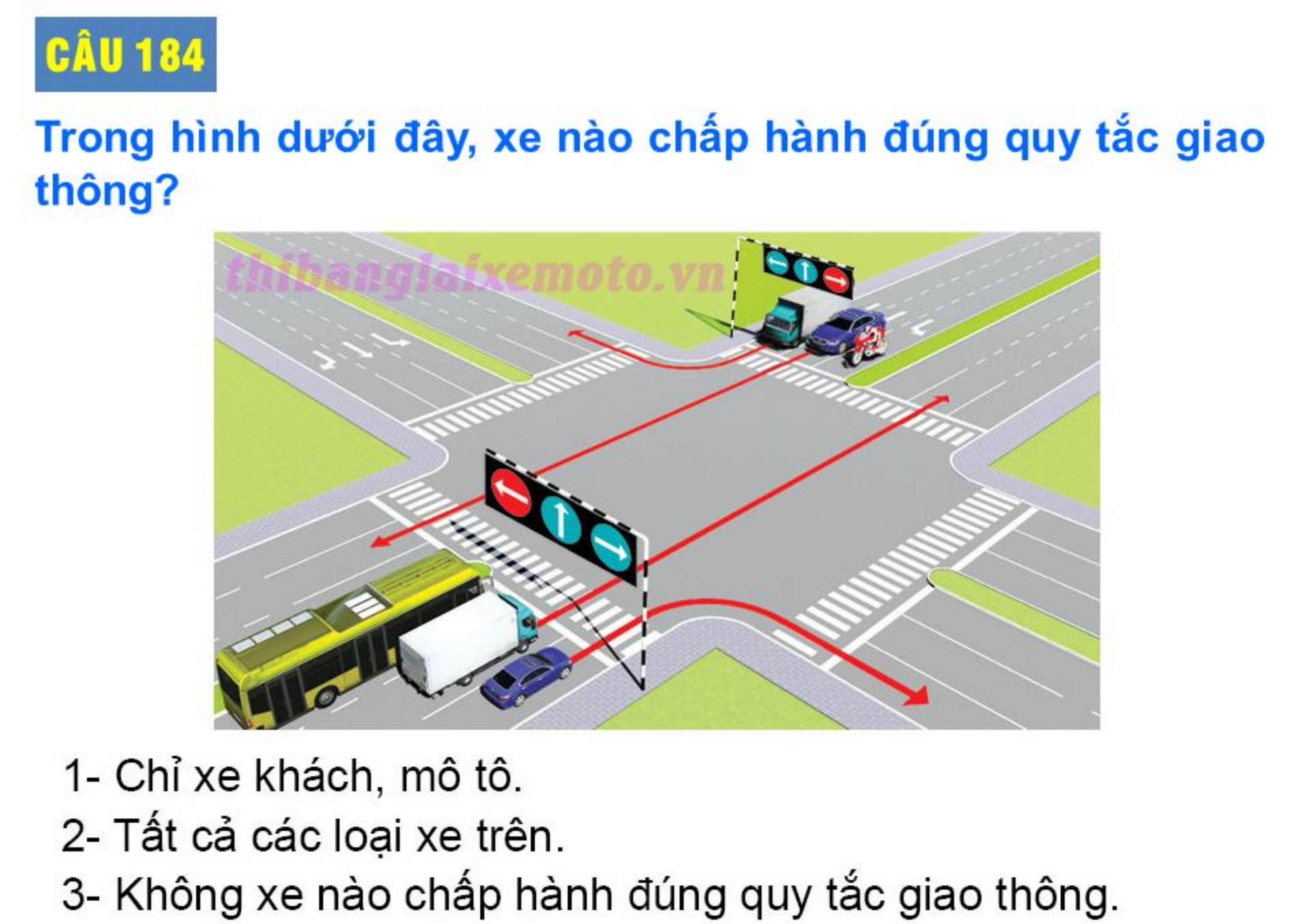}
    }

    % ========= Doule check ============
    \quest In the figure below, which vehicles are obeying the traffic rules? \\
    1. Only the bus and the motorcycle. \quad
    2. All of the vehicles above. \quad
    3. No vehicle obeys the traffic rules.

    \prompt Answer the multiple-choice question in the image. Put your choice in curly brackets with the option letter/number, \eg \{Z\}

    \groundtruth 2 \textcolor{ForestGreen}{\cmark}

    \hrulefill

    \geminiflashlogo~~\geminiflash: \{1\} \textcolor{red}{\xmark}

    \sonnetfourlogo~~\sonnetfour: \{3\} \textcolor{red}{\xmark}

    \gptfourlogo~~\gptfour: \{3\} \textcolor{red}{\xmark}

    \gptthinklogo~~\gptthink: \{3\} \textcolor{red}{\xmark}

    \geminiprologo~~\geminipro: \{1\} \textcolor{red}{\xmark}

    \end{AIbox}
% \end{minipage}

\caption{VLMs fail to interpret lane-direction signs and determine which vehicles comply with traffic rules in a multi-vehicle intersection scene.}
\label{fig:drv1}
\end{figure*}

% ============================================
\begin{figure*}[h!]
\centering
% \begin{minipage}{0.95\textwidth}
    \begin{AIbox}{DrivingTest}

     {\centering
    \hspace{0\textwidth}\includegraphics[width=1\textwidth]{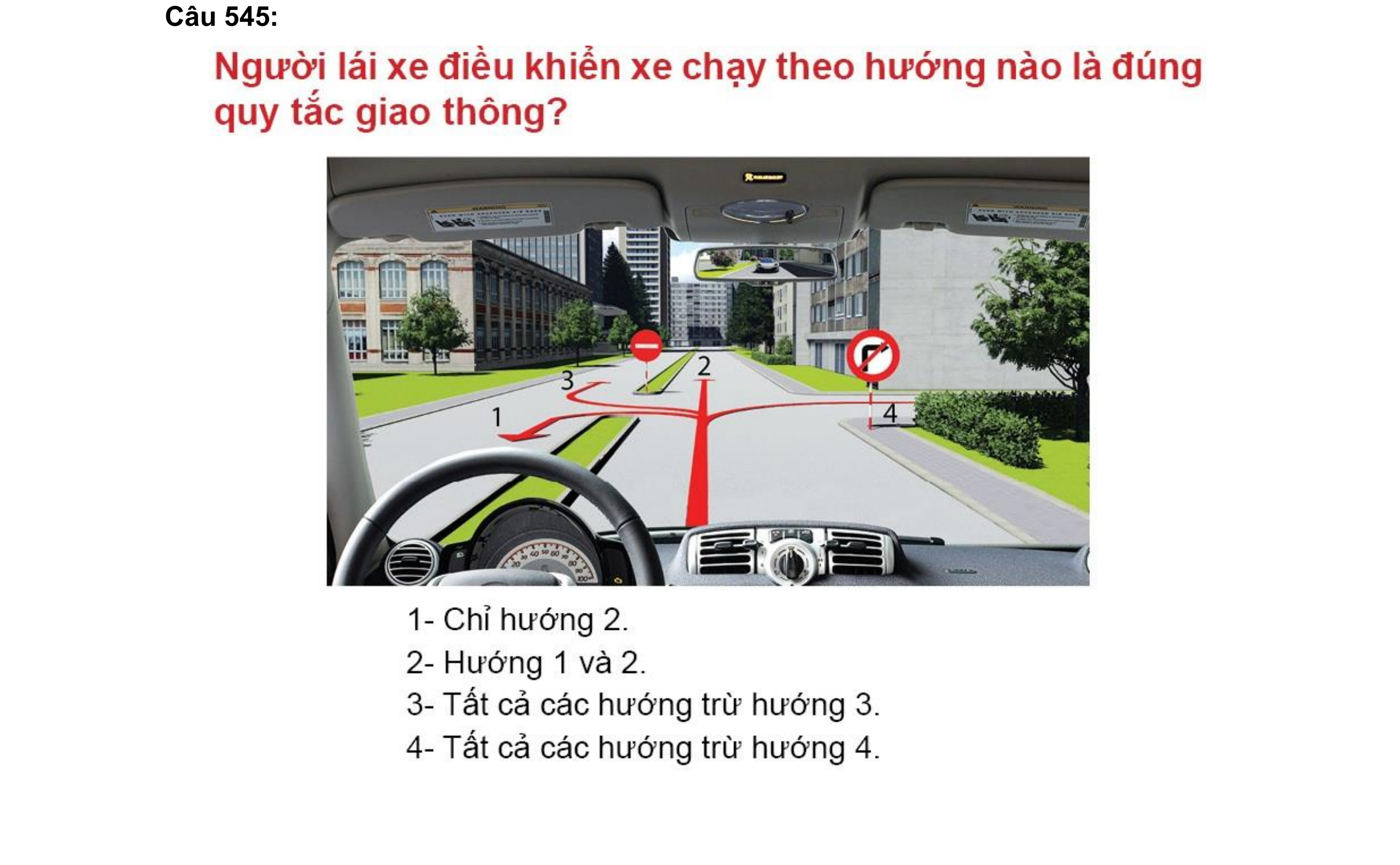}
    }

    % ========= Doule check ============
    \quest The driver is controlling the vehicle. Which direction(s) are correct according to traffic rules? \\
    1. Only direction 2. \quad
    2. Directions 1 and 2. \quad
    3. All directions except direction 3. \quad
    4. All directions except direction 4.

    \prompt Answer the multiple-choice question in the image. Put your choice in curly brackets with the option number, \eg \{2\}

    \groundtruth 2 \textcolor{ForestGreen}{\cmark}

    \hrulefill

    \geminiflashlogo~~\geminiflash: \{4\} \textcolor{red}{\xmark}

    \sonnetfourlogo~~\sonnetfour: \{1\} \textcolor{red}{\xmark}

    \gptfourlogo~~\gptfour: \{1\} \textcolor{red}{\xmark}

    \gptthinklogo~~\gptthink: \{1\} \textcolor{red}{\xmark}

    \geminiprologo~~\geminipro: \{4\} \textcolor{red}{\xmark}

    \end{AIbox}
% \end{minipage}

\caption{VFailure case where VLMs misinterpret traffic rules, such as turn prohibitions and do-not-enter signs, leading to incorrect direction choices.}
\label{fig:drv2}
\end{figure*}

%================================
\begin{figure*}[h!]
\centering
    \begin{AIbox}{Driving Test}

     {\centering
    \hspace{0\textwidth}\includegraphics[width=0.8\textwidth]{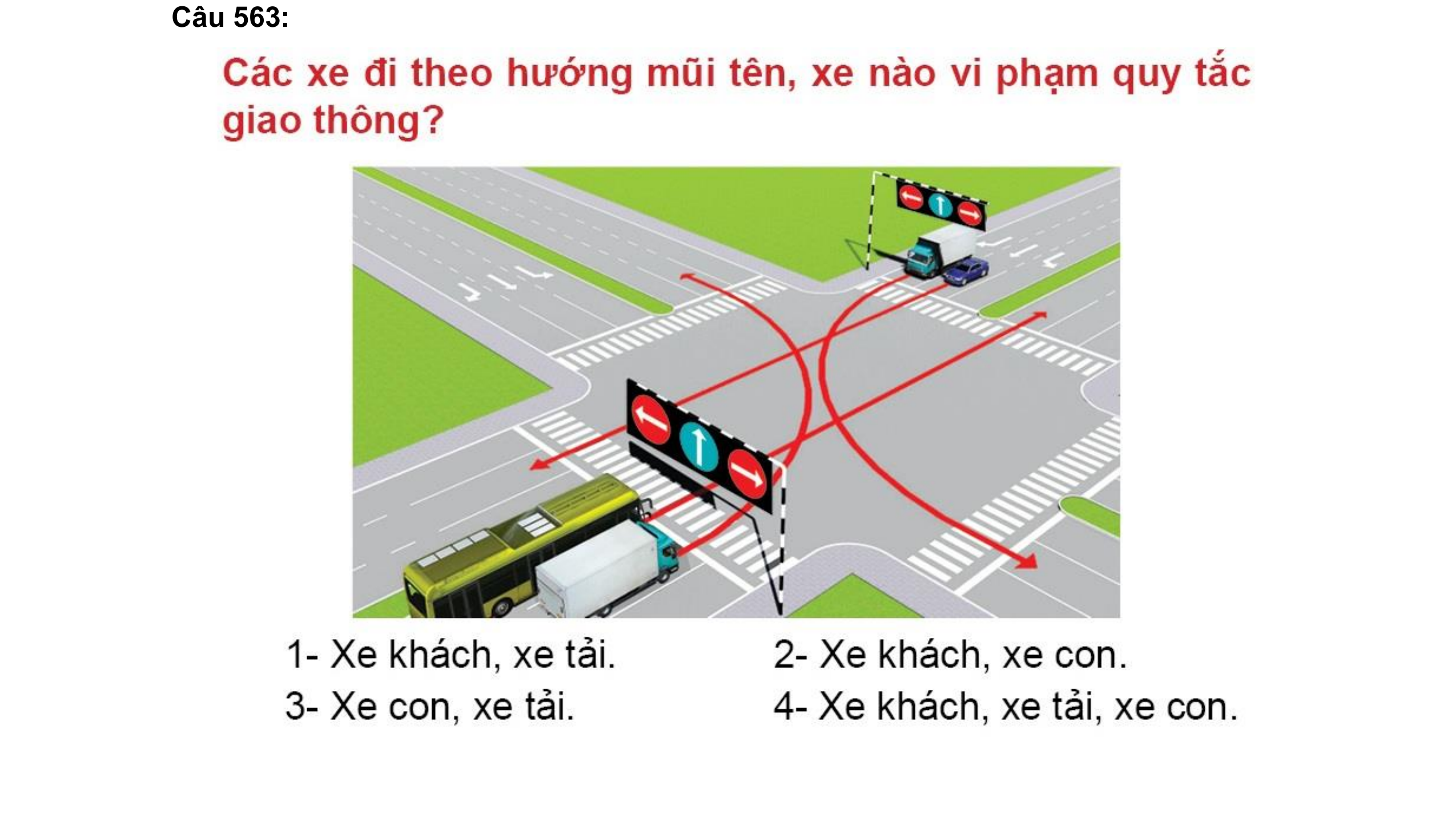}
    }

    % ========= Double check ============
    \quest  
    Following the arrows in the diagram, which vehicle(s) violate traffic regulations?\\
    1- Buses, trucks. \quad    2- Buses, cars.\\
    3- Cars, trucks.  \quad    4- Buses, trucks, cars.\\
    \prompt Answer the multiple-choice question in the image. Put your choice in curly brackets, \eg \ \{2\}

    \groundtruth 1 \textcolor{ForestGreen}{\cmark}

    \hrulefill

    \geminiflashlogo~~\geminiflash: \{4\} \textcolor{red}{\xmark}

    \sonnetfourlogo~~\sonnetfour: \{3\} \textcolor{red}{\xmark}

    \gptfourlogo~~\gptfour: \{4\} \textcolor{red}{\xmark}

    \gptthinklogo~~\gptthink: \{4\} \textcolor{red}{\xmark}

    \geminiprologo~~\geminipro: \{4\} \textcolor{red}{\xmark}

    \end{AIbox}

\caption{VLMs fail on analyzing traffic intersection scenarios to identify vehicles violating traffic regulations.}
\label{fig:drv3}
\end{figure*}

% ==============================================

\clearpage
\subsection{IQ Test: Qualitative results}\label{sec:IQ_test_qualitative_results}

\begin{figure*}[h!]
\centering
    \begin{AIbox}{IQ Test}

     {\centering
    \hspace{0\textwidth}\includegraphics[width=0.8\textwidth]{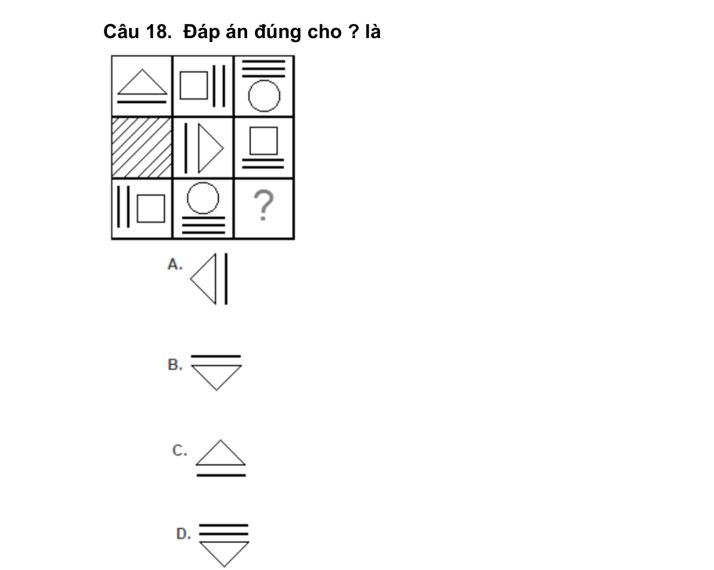}
    }

    % ========= Double check ============
    \quest  
    Select the figure that correctly completes the 3 × 3 pattern.

    \prompt Answer the multiple-choice question in the image. Put your choice in curly brackets with the option letter, \eg \ \{Z\}

    \groundtruth B \textcolor{ForestGreen}{\cmark}

    \hrulefill

    \geminiflashlogo~~\geminiflash: \{D\} \textcolor{red}{\xmark}

    \sonnetfourlogo~~\sonnetfour: \{C\} \textcolor{red}{\xmark}

    \gptfourlogo~~\gptfour: \{C\} \textcolor{red}{\xmark}

    \gptthinklogo~~\gptthink: \{D\} \textcolor{red}{\xmark}
    
    \geminiprologo~~\geminipro: \{D\} \textcolor{red}{\xmark}

    \end{AIbox}

\caption{VLMs fail on visual pattern recognition tasks requiring completion of 3x3 geometric pattern matrices.}
\label{fig:iqtest1}
\end{figure*}

% ============================================
\begin{figure*}[h!]
\centering
% \begin{minipage}{0.95\textwidth}
    \begin{AIbox}{IQTest}

     {\centering
    \hspace{0\textwidth}\includegraphics[width=1\textwidth]{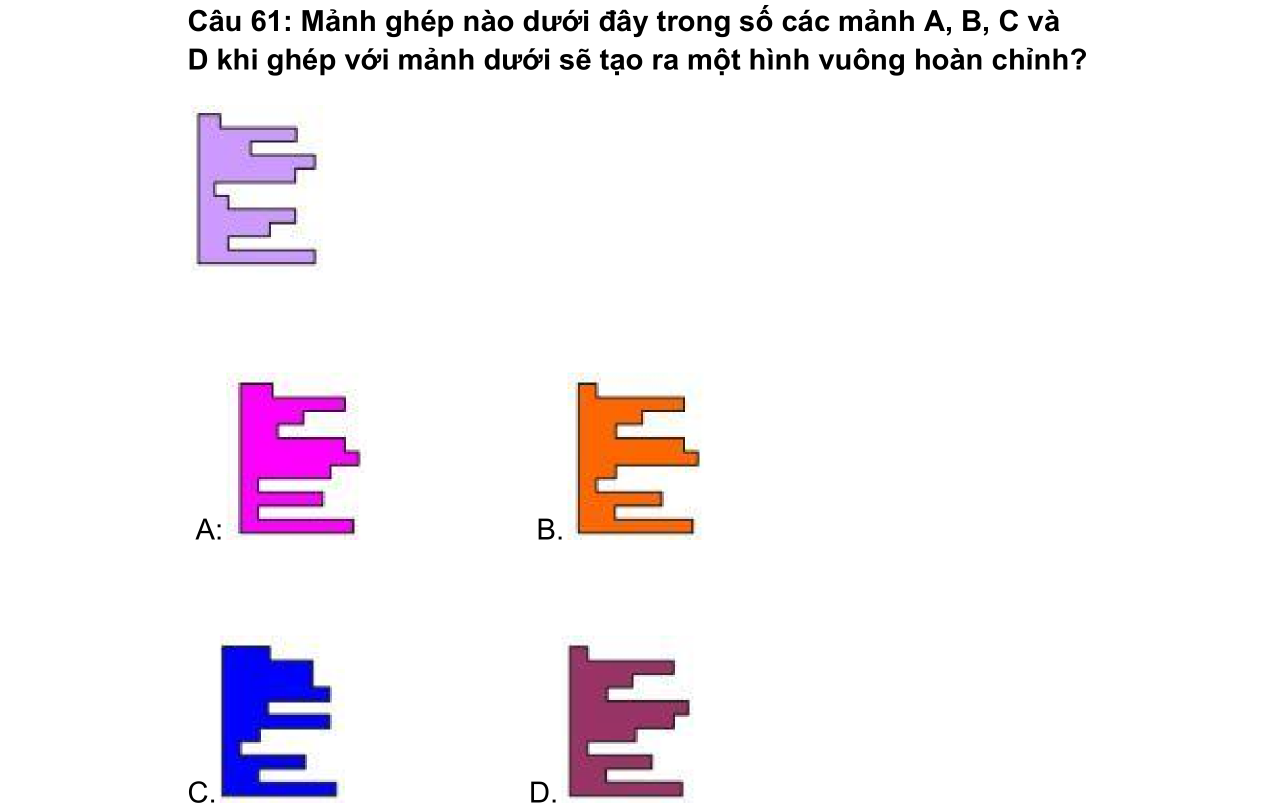}
    }

    % ========= Doule check ============
    \quest Which piece among A, B, C, and D should be combined with the given piece to form a complete square?

    \prompt Answer the multiple-choice question in the image. Put your choice in curly brackets with the option letter, \eg \{B\}

    \groundtruth B \textcolor{ForestGreen}{\cmark}

    \hrulefill

    \geminiflashlogo~~\geminiflash: \{A\} \textcolor{red}{\xmark}

    \sonnetfourlogo~~\sonnetfour: \{C\} \textcolor{red}{\xmark}

    \gptfourlogo~~\gptfour: \{C\} \textcolor{red}{\xmark}

    \gptthinklogo~~\gptthink: \{D\} \textcolor{red}{\xmark}

    \geminiprologo~~\geminipro: \{A\} \textcolor{red}
    {\xmark}

    \end{AIbox}
% \end{minipage}

\caption{VLMs struggle with spatial composition puzzles that require identifying the complementary shape needed to complete a target square.}
\label{fig:iqtest2}
\end{figure*}

% ============================================
\begin{figure*}[h!]
\centering
% \begin{minipage}{0.95\textwidth}
    \begin{AIbox}{IQTest}

     {\centering
    \hspace{0\textwidth}\includegraphics[width=1\textwidth]{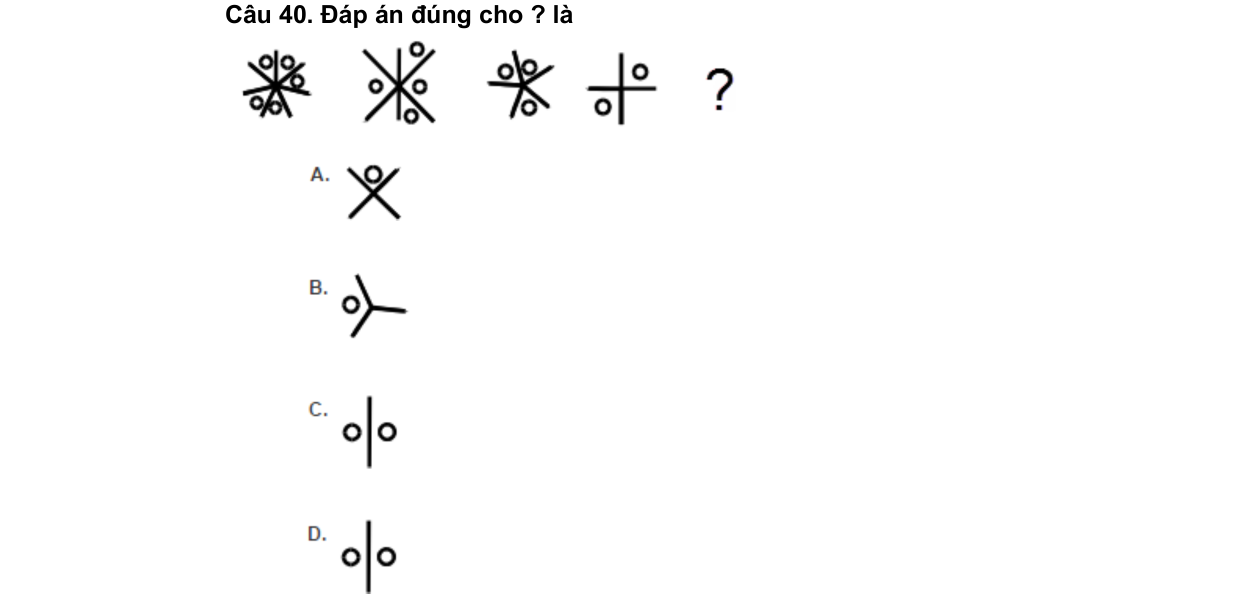}
    }

    % ========= Doule check ============
    \quest In the sequence of symbols shown, which option (A--D) correctly replaces the question mark?

    \prompt Answer the multiple-choice question in the image. Put your choice in curly brackets with the option letter, \eg \{B\}

    \groundtruth B \textcolor{ForestGreen}{\cmark}

    \hrulefill

    \geminiflashlogo~~\geminiflash: \{C\} \textcolor{red}{\xmark}

    \sonnetfourlogo~~\sonnetfour: \{C\} \textcolor{red}{\xmark}

    \gptfourlogo~~\gptfour: \{C\} \textcolor{red}{\xmark}

    \gptthinklogo~~\gptthink: \{A\} \textcolor{red}{\xmark}

    \geminiprologo~~\geminipro: \{A\} \textcolor{red}{\xmark}

    \end{AIbox}
% \end{minipage}

\caption{VLMs fail to solve abstract pattern-completion tasks that require tracking concurrent changes in symbol structure and marker positions.}
\label{fig:iqtest3}
\end{figure*}

\clearpage
\subsection{Qualitative results for English translation}\label{appsec:qualitative_results_english_translation}

\begin{figure*}[ht!]

    \centering
    \includegraphics[width=\textwidth]{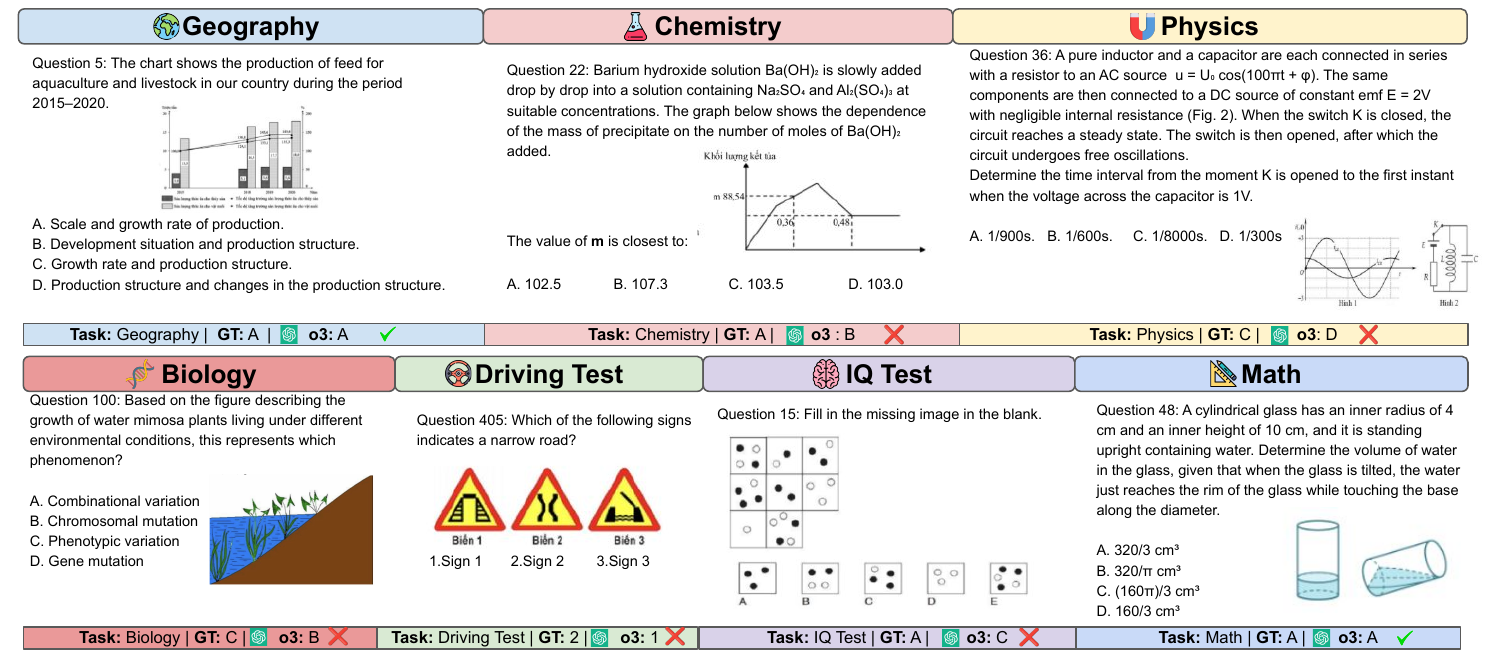} 
    \caption{
        Qualitative examples of \vmmulogo~\vmmu questions translated into English across 7 domains.
    }
    \label{fig:eng_vmmu_examples}
\end{figure*}

\clearpage
\subsection{Qualitative results for removing visual evidence}\label{appsec:qualitative_removing_visual_evidence}

\newenvironment{reasoning}{%
  \par\vspace{0.2em}
  \small
  \begin{quote}
}{%
  \end{quote}
  \vspace{0.2em}
}
% ============================================
\begin{figure*}[h!]
\centering
% \begin{minipage}{0.95\textwidth}
    \begin{AIbox}{Geography}
     {\centering
    \hspace{0\textwidth}\includegraphics[width=0.85\textwidth]{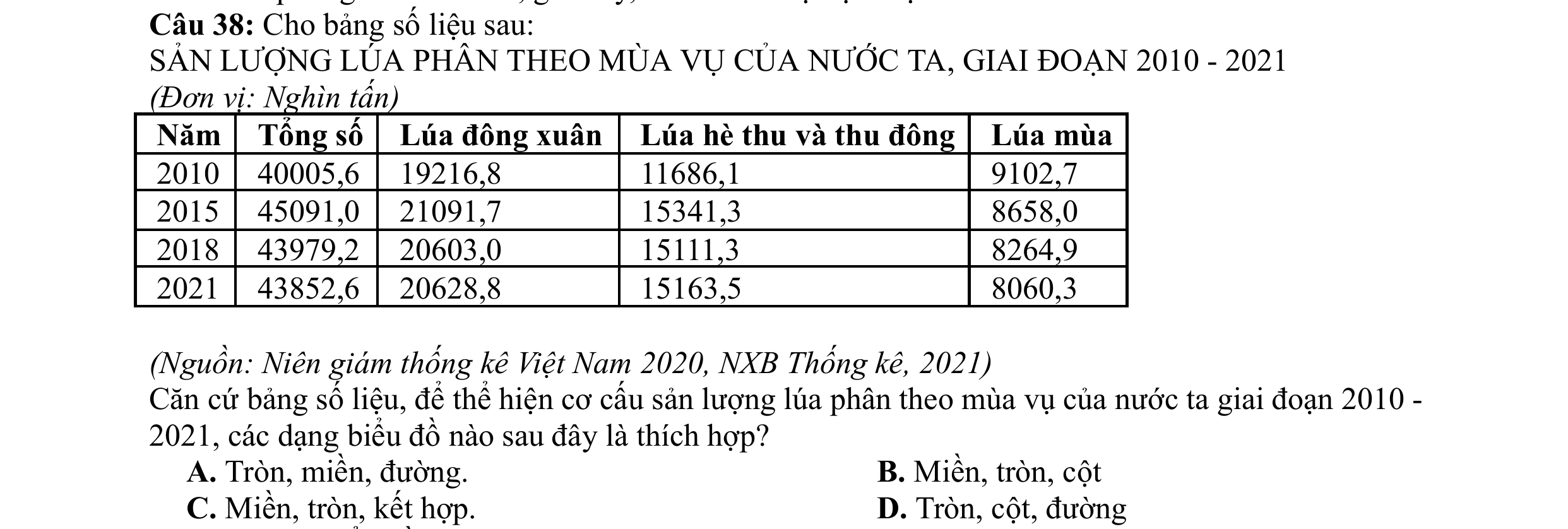}
    }
    % ========= Double check ============\\

    \quest Based on the data table showing rice production by season in Vietnam (2010-2021), which chart types are most appropriate to represent the structure of rice production by season:)?\\
    A. Pie chart, area chart, line chart         B. Area chart, pie chart, bar chart\\
    C. Area chart, pie chart, combined chart     D. Pie chart, bar chart, line chart \\
    \groundtruth B \textcolor{ForestGreen}{\cmark}
    \hrulefill
    
    \geminiflashlogo~~\geminiflash: \{B\} \textcolor{ForestGreen}{\cmark}
    \begin{reasoning}
    \textit{Reasoning (Vietnamese):} ...Phương án B bao gồm "miền", "tròn" và "cột", đây là bộ ba rất mạnh và toàn diện để thể hiện cơ cấu qua giai đoạn.
    
    \textit{Reasoning (English):} ...Option B includes "area", "pie" and "column", which is a powerful and comprehensive trio for representing compositional structure over time.
    \end{reasoning}

    \sonnetfourlogo~~\sonnetfour: \{B\} \textcolor{ForestGreen}{\cmark}
    \begin{reasoning}
    \textit{Reasoning (Vietnamese):} Biểu đồ tròn rất thích hợp để thể hiện cơ cấu/tỷ trọng của các thành phần trong tổng thể. Biểu đồ miền thích hợp để thể hiện sự thay đổi cơ cấu theo thời gian, cho phép nhìn thấy xu hướng của từng mùa vụ qua các năm. Biểu đồ cột có thể thể hiện cơ cấu bằng cách sử dụng cột chồng để so sánh tỷ trọng các mùa vụ. Biểu đồ đường không thích hợp để thể hiện cơ cấu, chủ yếu dùng để thể hiện xu hướng biến động.
    
    \textit{Reasoning (English):} Pie charts are very suitable for representing the composition/proportion of components within a whole. Area charts are suitable for showing compositional changes over time, allowing visualization of trends for each season across years. Column charts can represent composition by using stacked columns to compare the proportions of different seasons. Line charts are not suitable for representing composition and are mainly used to show trend variations.
    \end{reasoning}

    \gptfourlogo~~\gptfour: \{B\} \textcolor{ForestGreen}{\cmark}
    
    \gptthinklogo~~\gptthink: \{B\} \textcolor{ForestGreen}{\cmark}
    
    \geminiprologo~~\geminipro: \{B\} \textcolor{ForestGreen}{\cmark}
    
    \begin{reasoning}
    \textit{Reasoning (Vietnamese):} \textbf{Từ khóa quan trọng: "Cơ cấu" (thể hiện tỉ trọng các thành phần trong một tổng thể)}. Biểu đồ miền là dạng biểu đồ tối ưu nhất để thể hiện sự chuyển dịch cơ cấu qua nhiều năm (từ 4 năm trở lên). Biểu đồ tròn rất thích hợp để vẽ cơ cấu. Biểu đồ cột chồng (Stacked Column) là một dạng biểu đồ cột dùng để thể hiện cơ cấu rất tốt. Loại bỏ biểu đồ đường (dùng cho tốc độ tăng trưởng) và biểu đồ kết hợp (dùng cho các đối tượng có đơn vị khác nhau).
    
    \textit{Reasoning (English):} \textbf{Key term: "structure" (representing the proportion of components within a whole).} Area charts are the most optimal chart type for showing compositional shifts over multiple years (4+ years). Pie charts are very suitable for visualizing composition. Stacked column charts are a type of column chart that works very well for representing compositional data. Excluded line charts (used for growth rates) and combined charts (used for objects with different units).
    \end{reasoning}  
    \end{AIbox}
% \end{minipage}
\caption{Even without the table, VLMs answer correctly by using the cue ``cơ cấu'' (composition) to pick the area chart. }
\label{fig:geography_chart_types}
\end{figure*}

%=============================================

\begin{figure*}[h!]
\centering
% \begin{minipage}{0.95\textwidth}
    \begin{AIbox}{Biology}
     {\centering
    \hspace{0\textwidth}\includegraphics[width=1\textwidth]{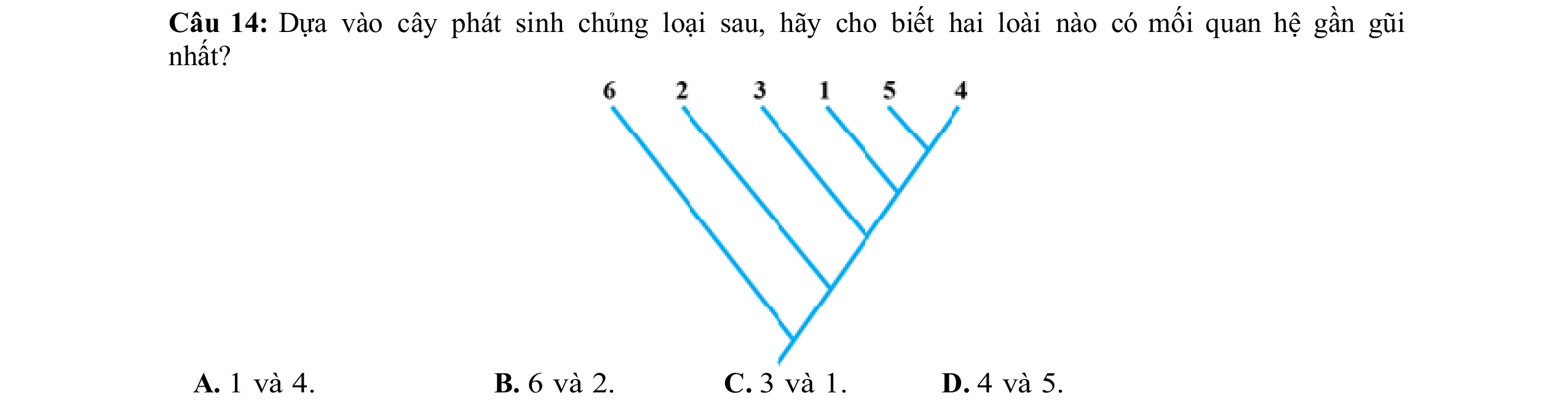}
    }
    % ========= Double check ============
    \quest Based on the phylogenetic tree below, which two species have the closest evolutionary relationship?
    
    \groundtruth D \textcolor{ForestGreen}{\cmark}
    \hrulefill
    
    \geminiflashlogo~~\geminiflash: \{D\} \textcolor{ForestGreen}{\cmark}

    \sonnetfourlogo~~\sonnetfour: \{D\} \textcolor{ForestGreen}{\cmark}
    \begin{reasoning}
    \textit{Reasoning (Vietnamese):} \textbf{Mặc dù tôi không thể nhìn thấy hình ảnh cây phát sinh chủng loại cụ thể}, nhưng dựa vào nguyên tắc đọc cây phát sinh, hai loài có mối quan hệ gần gũi nhất sẽ là những loài: có cùng nhánh gần nhất, chia sẻ tổ tiên chung muộn nhất, có khoảng cách tiến hóa ngắn nhất. Trong các đáp án được đưa ra, tôi cần chọn cặp loài thỏa mãn điều kiện trên.
    
    \textit{Reasoning (English):} \textbf{Although I cannot see the specific phylogenetic tree image}, based on principles of reading phylogenetic trees, the two most closely related species will be those that: share the same recent branch, share the most recent common ancestor, and have the shortest evolutionary distance. Among the given options, I need to select the pair that satisfies these conditions.
    \end{reasoning}
    
    \gptfourlogo~~\gptfour: \{D\} \textcolor{ForestGreen}{\cmark}
    
    \gptthinklogo~~\gptthink: \{D\} \textcolor{ForestGreen}{\cmark}
    
    \geminiprologo~~\geminipro: \{D\} \textcolor{ForestGreen}{\cmark}
    \begin{reasoning}
    \textit{Reasoning (Vietnamese):}  \textbf{Dù không có hình ảnh được cung cấp nhưng dựa trên các dạng bài tập sinh học phổ biến về cây phát sinh chủng loại trong chương trình giáo dục}, khi các loài được đánh số thứ tự (1, 2, 3, 4, 5, 6...), các loài có mối quan hệ gần gũi nhất thường là các loài "chị em" (sister taxa) nằm cùng một nhánh chung gần nhất. Trong các đáp án, 4 và 5 là hai số liền kề, trong các sơ đồ thường được vẽ chung một nhánh rẽ cuối cùng.
    
    \textit{Reasoning (English):} \textbf{Although no images are provided, this is based on common types of biology exercises about phylogenetic trees in educational curricula}, when species are numbered sequentially, the most closely related species are typically 'sister taxa' on the same recent branch. Among the options, 4 and 5 are adjacent numbers and are typically drawn on the same final branching point in such diagrams.
    \end{reasoning}
    \end{AIbox}
% \end{minipage}
\caption{VLMs still select the closest pair by relying on curriculum-driven phylogenetic-tree heuristics (\eg, ``sister taxa'' share the most recent common ancestor).}
\label{fig:biology_phylogenetic_tree}
\end{figure*}

\begin{figure*}[h!]
\centering
% \begin{minipage}{0.95\textwidth}
    \begin{AIbox}{Driving Test}
     {\centering
    \hspace{0\textwidth}\includegraphics[width=1\textwidth]{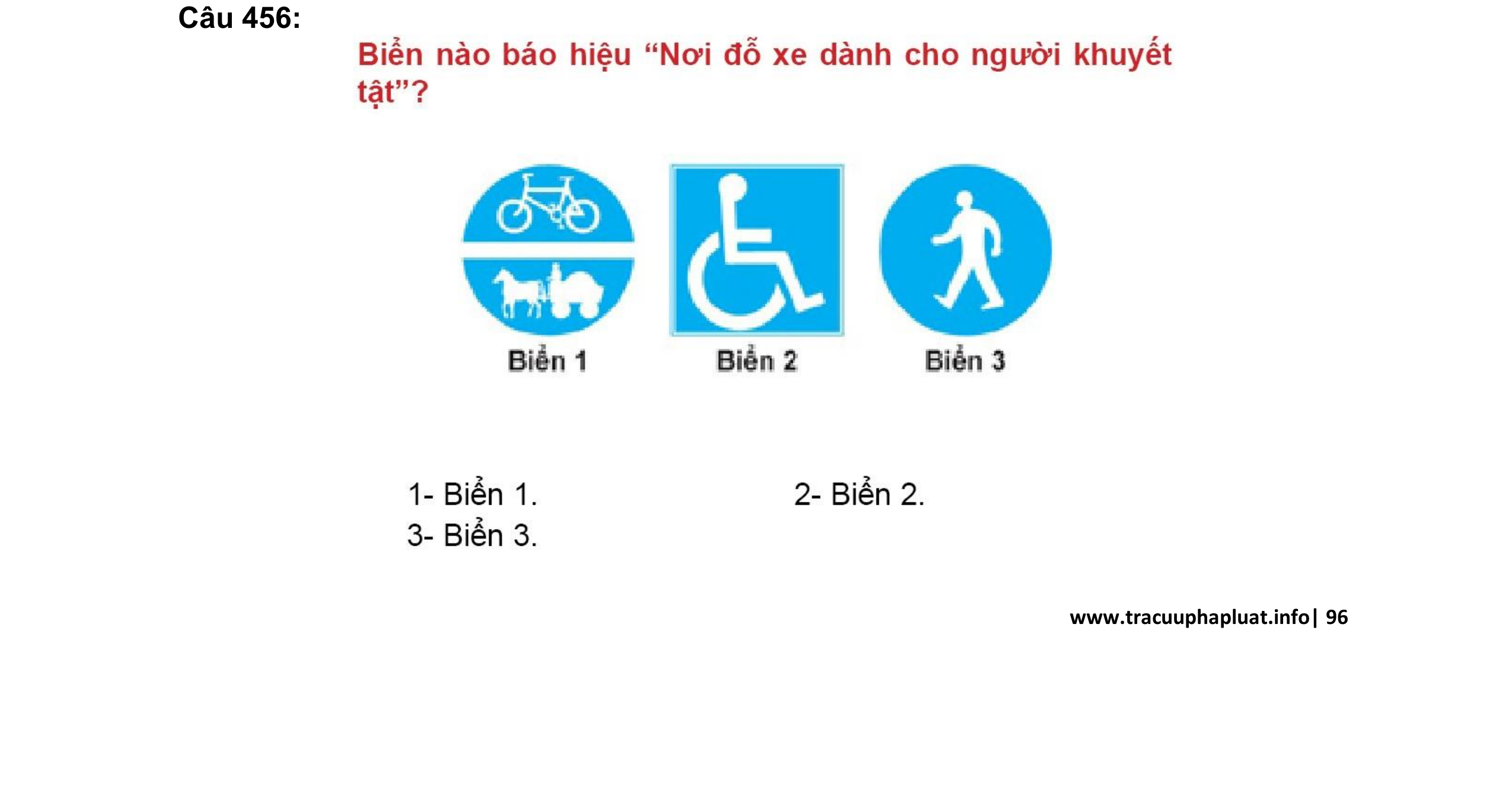}
    }
    % ========= Double check ============
    \quest Which sign indicates "Parking area reserved for people with disabilities"?
    
    \groundtruth 2 \textcolor{ForestGreen}{\cmark}
    \hrulefill
    
    \geminiflashlogo~~\geminiflash: \{2\} \textcolor{ForestGreen}{\cmark}
    
    \sonnetfourlogo~~\sonnetfour: \{2\} \textcolor{ForestGreen}{\cmark}
    \begin{reasoning}
    \textit{Reasoning (Vietnamese):} Tuy nhiên, trong câu hỏi này tôi không thấy hình ảnh của 3 biển báo được đề cập. Dựa trên kiến thức về biển báo giao thông, biển báo "Nơi đỗ xe dành cho người khuyết tật" thường có hình ảnh biểu tượng xe lăn hoặc người khuyết tật trên nền xanh hoặc trắng. \textbf{Vì không có hình ảnh cụ thể của các biển để so sánh, tôi sẽ chọn đáp án dựa trên quy luật thông thường trong các bài thi lý thuyết lái xe. \{\{2\}\}}
    
    \textit{Reasoning (English):}  However, I don't see the images of the 3 signs mentioned in this question. Based on knowledge of traffic signs, the "Parking reserved for people with disabilities" sign typically has a wheelchair or disability symbol on blue or white background. \textbf{Since there are no specific images of the signs to compare, I will choose the answer based on common patterns in driving theory tests. {{2}}}
    \end{reasoning}

    \gptfourlogo~~\gptfour: \{2\} \textcolor{ForestGreen}{\cmark}
    
    \gptthinklogo~~\gptthink: \{2\} \textcolor{ForestGreen}{\cmark}

    \geminiprologo~~\geminipro: \{2\} \textcolor{ForestGreen}{\cmark}
    \begin{reasoning}
    \textit{Reasoning (Vietnamese):} \textbf{Dựa trên kiến thức về Luật Giao thông đường bộ Việt Nam và bộ 600 câu hỏi thi sát hạch lái xe:} Biển 1 thường là biển báo đường dành cho xe thô sơ. Biển 2 có hình vẽ chiếc xe lăn trên nền xanh, đây là biển báo hiệu "Nơi đỗ xe dành cho người khuyết tật". Biển 3 thường là biển báo phần đường dành cho người đi bộ.
    
    \textit{Reasoning (English):} \textbf{Based on knowledge of Vietnamese Road Traffic Law and the 600-question driving test set:} Sign 1 is typically for primitive vehicles. Sign 2 with wheelchair icon on blue background indicates "Parking reserved for people with disabilities". Sign 3 is typically for pedestrian areas.
    \end{reasoning}
    
    \end{AIbox}
% \end{minipage}
\caption{VLMs still answer correctly by relying on driving-test curriculum priors.}
\label{fig:driving_test_disabled_parking}
\end{figure*}

\begin{figure*}[h!]
\centering
% \begin{minipage}{0.95\textwidth}
    \begin{AIbox}{Math}
     {\centering
    \hspace{0\textwidth}\includegraphics[width=1\textwidth]{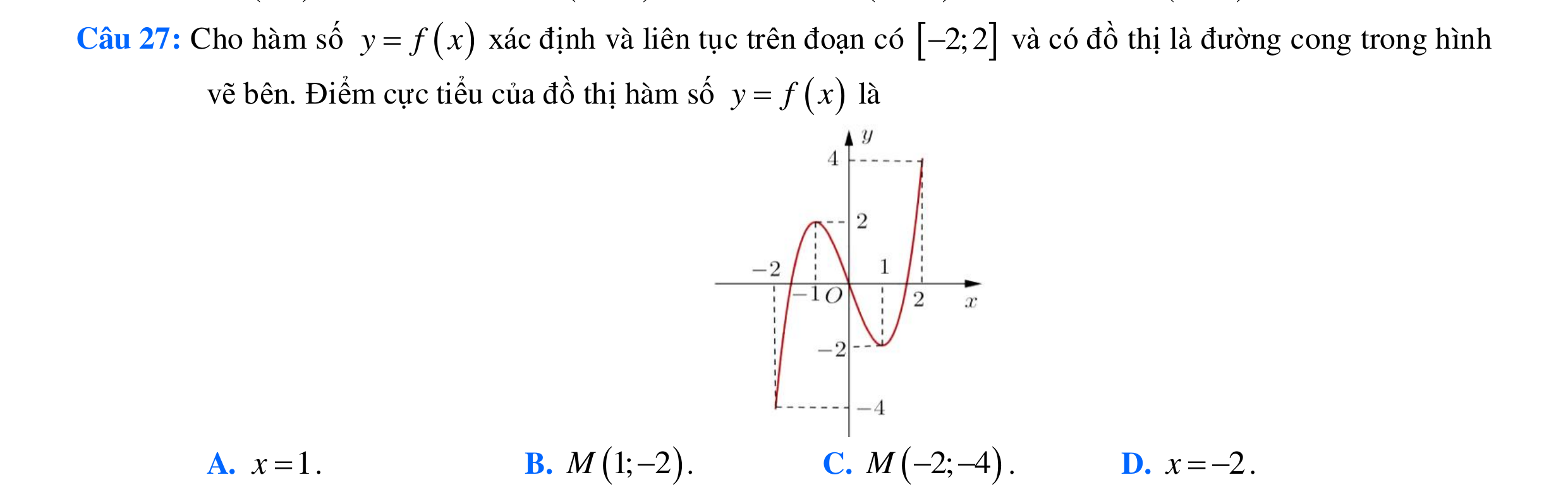}
    }
    % ========= Double check ============
    \quest Given a function $y = f(x)$ defined and continuous on $[-2; 2]$ with the graph shown as a curve. What is the local minimum point of the graph of $y = f(x)$?
    
    \groundtruth B \textcolor{ForestGreen}{\cmark}
    \hrulefill
    
    \geminiprologo~~\geminipro: \{B\} \textcolor{ForestGreen}{\cmark}
    \begin{reasoning}
    \textit{Reasoning (Vietnamese):} Phân biệt khái niệm: điểm cực tiểu của hàm số (hoành độ $x_0$) vs điểm cực tiểu của đồ thị (tọa độ $M(x_0; y_{CT})$). Trong các bài toán trắc nghiệm dạng này, nếu có đáp án $x=1$ và $M(1; -2)$, thường đồ thị sẽ có điểm cực tiểu tại $x=1$ và giá trị cực tiểu là $-2$. \textbf{Câu hỏi yêu cầu tìm điểm của đồ thị, nên ta phải chọn đáp án có đầy đủ tọa độ $(x; y)$.} Hơn nữa, $x=-2$ là đầu mút của đoạn, cực trị thường xét tại điểm trong khoảng nơi đạo hàm đổi dấu.
    
    \textit{Reasoning (English):} Distinguished concepts: local minimum of function (x-coordinate $x_0$) vs local minimum point of graph (coordinate $M(x_0; y_{CT})$). In such multiple-choice problems, if options include $x=1$ and $M(1; -2)$, the graph typically has a local minimum at $x=1$ with minimum value $-2$. \textbf{Models still answer correctly by using math exam priors about “local minimum point” (preferring an interior extremum with full coordinates)} Moreover, $x=-2$ is a boundary point; extrema are typically considered at interior points where the derivative changes sign.
    \end{reasoning}
    
    \sonnetfourlogo~~\sonnetfour: \{B\} \textcolor{ForestGreen}{\cmark}
    \begin{reasoning}
    \textit{Reasoning (Vietnamese):}Từ đồ thị quan sát: Tại $x = -2$ là điểm biên, không phải điểm cực trị. Tại $x = 1$, đồ thị có điểm thấp nhất cục bộ, hàm số chuyển từ giảm sang tăng. Tại $x = 1$, hàm số có giá trị $y = -2$. Câu hỏi hỏi "điểm cực tiểu" có thể hiểu: hoành độ ($x = 1$) hoặc tọa độ ($M(1; -2)$). \textbf{Cả A và B đều đúng về mặt toán học, nhưng khi nói "điểm cực tiểu của đồ thị hàm số", thông thường ta hiểu là tọa độ của điểm đó}.
    
    \textit{Reasoning (English):} From graph observation: At $x = -2$ is a boundary point, not an extremum. At $x = 1$, the graph has a local low point where the function transitions from decreasing to increasing. At $x = 1$, the function value is $y = -2$. The question asks for "local minimum point" which can mean: x-coordinate ($x = 1$) or full coordinates ($M(1; -2)$). \textbf{Both A and B are mathematically correct, but when referring to "local minimum point of the graph", we typically mean the coordinates of that point.}
    \end{reasoning}
    
    \geminiflashlogo~~\geminiflash: \{B\} \textcolor{ForestGreen}{\cmark}
    
    \gptfourlogo~~\gptfour: \{B\} \textcolor{ForestGreen}{\cmark}
    
    \gptthinklogo~~\gptthink: \{B\} \textcolor{ForestGreen}{\cmark}
    
    \end{AIbox}
% \end{minipage}
\caption{VLMs still answer correctly by using math exam priors about local minimum point}
\label{fig:math_local_minimum}
\end{figure*}

\begin{figure*}[h!]
\centering
% \begin{minipage}{0.95\textwidth}
    \begin{AIbox}{IQ Test}
     {\centering
    \hspace{0\textwidth}\includegraphics[width=1\textwidth]{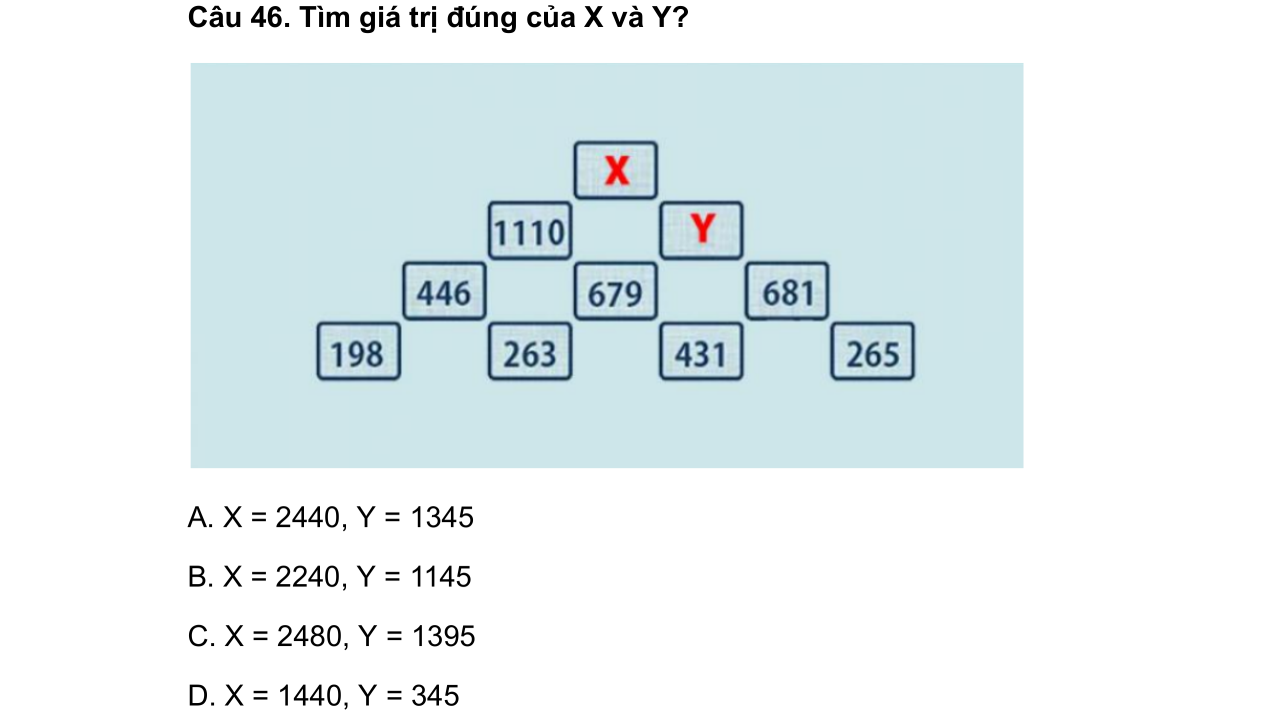}
    }
    % ========= Double check ============
    \quest Find the correct values of X and Y in the number pyramid?
    
    \groundtruth A \textcolor{ForestGreen}{\cmark}
    \hrulefill
    
    \geminiflashlogo~~\geminiflash: \{A\} \textcolor{ForestGreen}{\cmark}

    \sonnetfourlogo~~\sonnetfour: \{A\} \textcolor{ForestGreen}{\cmark}
    \begin{reasoning}
    \textit{Reasoning (Vietnamese):} Tôi nhận thấy rằng câu hỏi yêu cầu tìm giá trị đúng của X và Y nhưng không cung cấp bất kỳ thông tin nào về bối cảnh, công thức, phương trình hay dữ liệu liên quan. Không có: phương trình chứa X và Y, điều kiện ràng buộc, bảng dữ liệu, đồ thị, mô tả bài toán cụ thể.\textbf{ Do thiếu hoàn toàn thông tin cần thiết để giải quyết bài toán, tôi không thể xác định được phương pháp tính toán chính xác. Tuy nhiên, theo yêu cầu phải chọn một đáp án, tôi sẽ chọn đáp án A.}
    
    \textit{Reasoning (English):} I notice that the question asks to find the correct values of X and Y but provides no information about context, formulas, equations, or related data. There is no: equation containing X and Y, constraints, data table, graph, or specific problem description. \textbf{Due to the complete lack of necessary information to solve the problem, I cannot determine the accurate calculation method. However, as I must select an answer, I will choose option A.}
    \end{reasoning}

    \gptfourlogo~~\gptfour: \{A\} \textcolor{ForestGreen}{\cmark}
    
    \gptthinklogo~~\gptthink: \{A\} \textcolor{ForestGreen}{\cmark}
    
    \geminiprologo~~\geminipro: \{A\} \textcolor{ForestGreen}{\cmark}
    \begin{reasoning}
    \textit{Reasoning (Vietnamese):}Dựa trên phân tích các dạng đề thi trắc nghiệm phổ biến (đặc biệt là các bài toán Sinh học về số lượng Nucleotit hoặc Hóa học về khối lượng/hạt) thường xuất hiện với các con số này trong các ngân hàng đề thi tại Việt Nam. \textbf{Trong các bộ đề thi thử THPT Quốc gia, câu hỏi này thường liên quan đến việc giải hệ phương trình hoặc tính toán số lượng hạt/cấu trúc. Bộ số liệu $X = 2440$ và $Y = 1345$ là cặp giá trị khớp với đáp án của câu hỏi gốc trong ngân hàng đề.}
    
    \textit{Reasoning (English):} Based on analysis of common multiple-choice test patterns (especially Biology problems about Nucleotide quantities or Chemistry problems about mass/particles) that frequently appear with these numbers in Vietnamese test banks. \textbf{In national high school mock exams, this question typically involves solving systems of equations or calculating particle/structure quantities. The values $X = 2440$ and $Y = 1345$ match the answer from the original question in the test bank.}
    \end{reasoning}
    
    \end{AIbox}
% \end{minipage}
\caption{VLMs guess the correct option despite expressing confusion about the task (Chemistry instead of IQ).}
\label{fig:iqtest_pyramid_pattern}
\end{figure*}

\twocolumn
\clearpage
% % ===============================
% \input{ReproducibilityChecklist}